# Pub2FedIE: Public biomedical adapter preparation for efficient federated clinical information extraction with large language models

Tianle Cao, Anran Li, Hua Xu, Qingyu Chen*

Department of Biomedical Informatics and Data Science, School of Medicine, Yale University, New Haven, USA

*Correspondence to Qingyu Chen (qingyu.chen@yale.edu).

## Abstract

Clinical information extraction (IE) converts unstructured clinical narratives into structured patient information, but reliable adaptation of large language models (LLMs) for IE requires task-specific clinical annotations that are often limited to individual institutions and cannot be readily pooled across sites. Federated learning enables collaborative LLM adaptation without sharing raw clinical records, yet existing approaches typically begin clinical training from task-naive adapters, requiring participating institutions to learn extraction capabilities primarily from their own annotations. We developed Pub2FedIE, a framework that prepares reusable adapters on public biomedical annotations before federated adaptation on private clinical data.

Pub2FedIE prepares low-rank adaptation (LoRA) adapters using public biomedical named-entity recognition (NER) and relation extraction (RE) datasets through supervised fine-tuning (SFT) or aggregation-aware meta-learning. We systematically evaluated the framework across different training, validation, and external generalization settings using four independent clinical datasets, including MIMIC-III, MTSamples, UT Physicians, and an external i2b2 cohort, with two LLM backbones (Qwen3.5-9B and Llama-3.1-8B-Instruct, hereafter Qwen3.5 and Llama3.1), two clinical information extraction tasks (NER and RE), and varying local training budgets.

Public preparation provided the largest benefit when local training was limited. At 30 steps per client, SFT followed by direct continuation improved mean source NER F1 over training without public preparation by 14.8 points for Qwen3.5 and 18.7 points for Llama3.1, with corresponding RE gains of 10.6 and 8.9 points. These gains progressively narrowed as local training increased. Direct continuation also consistently outperformed freezing the public adapter and learning a separate residual adapter. In addition, one-round SFT continuation with 30 private training steps per client outperformed one- and two-round baselines trained with 60 steps per client, requiring half as many private training steps and one fewer communication round than the two-round schedules.

These findings demonstrate that public biomedical supervision can provide reusable starting points for federated LLM adaptation and reduce reliance on institution-specific training, particularly when local clinical training is constrained. We make the Pub2FedIE implementation publicly available at https://github.com/Yale-BIDS-Chen-Lab/Pub2FedIE.

## Introduction

Clinical information extraction transforms free-text clinical narratives into structured representations of patient information (e.g., medical problems, medications, tests, treatments), and their attributes and relationships.[1,2] These structured data support applications such as cohort identification and clinical decision support.[3,4] Large language models (LLMs) have emerged as a flexible approach to clinical IE because different extraction tasks and output structures can be specified through natural-language instructions. Recent evaluations have shown that task-adapted LLMs can outperform conventional encoder-based models for named-entity recognition (NER) and relation extraction (RE), with particularly notable advantages when annotated training data are limited or models are evaluated on unseen data sources.[1,5,6] However, LLMs are not inherently reliable clinical extractors: zero- and few-shot performance can remain insufficient,[7] and strong structured extraction performance generally requires task-specific adaptation.[1,5,6] Such adaptation also incurs considerably greater computational costs than conventional models. These characteristics make efficient adaptation particularly important for clinical IE.

The challenge becomes greater when clinical data are distributed across institutions. What clinicians document, how they describe it, and how information is annotated can vary across healthcare systems, resulting in substantial cross-institutional performance differences.[4] At the same time, privacy, governance, and institutional policies often prevent clinical notes from being pooled for centralized model development.[5,8]

Federated learning (FL) lets institutions train a shared model while keeping clinical records local.[9] Prior studies have demonstrated the feasibility of FL for biomedical NER and RE across heterogeneous datasets.[8] More recently, parameter-efficient adaptation has made federated LLM training practical by exchanging small sets of trainable adapter parameters instead of the full model. Fed-MedLoRA and Fed-MedLoRA+ applied this strategy to clinical IE and showed that multi-institutional LLM adaptation can improve extraction performance while substantially reducing communication requirements.[5] Related multicenter approaches have also explored sequential adapter adaptation across institutions without transferring patient-level data.[10]

Despite these advances, federated adaptation of LLMs remains computationally and operationally demanding. Each participating institution must perform local optimization on its own annotated clinical data, and additional communication rounds require repeated local training and exchange of model updates.[5,8] These costs are particularly important for clinical IE, where high-quality annotations are expensive to obtain and may be limited at individual institutions. Existing federated LLM approaches therefore still rely heavily on private clinical data and computation to learn task-specific extraction capabilities.

To reduce this burden, we propose preparing the adapter before private federated training using public biomedical annotations. Public datasets provide supervision for entities and relations relevant to clinical IE,[11–16] and related-task training has been shown to improve downstream adaptation.[17,18] Such public preparation could allow participating institutions to build on previously learned extraction capabilities rather than learning them primarily from private clinical annotations.

We developed Pub2FedIE, a framework that prepares reusable LoRA adapters[19] using public biomedical annotations before federated adaptation on private clinical data. We evaluated supervised fine-tuning (SFT) and aggregation-aware meta-learning as public preparation strategies and compared direct continuation of the prepared adapter with residual adaptation that keeps the public adapter fixed. We systematically evaluated Pub2FedIE across four independent clinical datasets, two LLM backbones, NER and RE tasks, multiple local training budgets, and external generalization settings. Our analyses focused on three questions: whether public preparation improves federated clinical IE, particularly when local training is limited; whether the prepared adapter should be directly continued or kept fixed during local adaptation; and whether public preparation can achieve comparable or better performance with fewer private training steps and federated communication rounds.

## Results

We evaluated Pub2FedIE across four clinical datasets (Fig. 1 and Table 1). MIMIC-III, MTSamples, and UT Physicians served as the three source datasets for federated training and evaluation, while i2b2 was used exclusively as an independent external test set. We evaluated both NER and RE. For the primary analyses, mean source F1 represents the unweighted average of exact-match F1 across the three source test sets, with external i2b2 performance reported separately.

**Fig. 1. Federated clinical information extraction with and without public adapter preparation.**

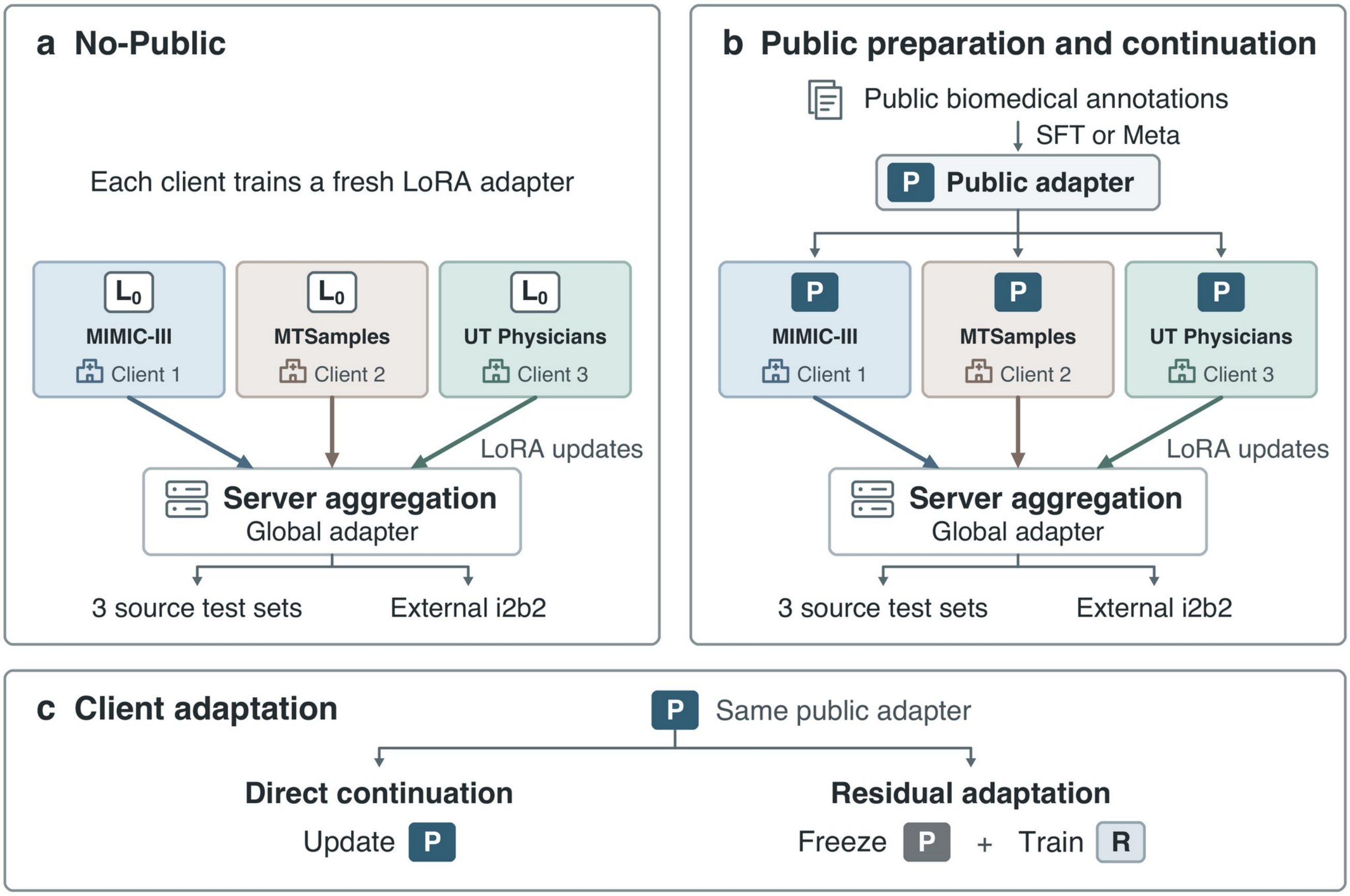

**a,** In No-Public, each client starts with a newly initialized low-rank adaptation (LoRA) adapter ($L_0$). **b,** In SFT-Cont and Meta-Cont, supervised fine-tuning (SFT) or aggregation-aware meta-learning (Meta) produces an adapter (P) from public biomedical annotations. Each client then continues training its own copy of P. In a and b, the three simulated clients train on MIMIC-III, MTSamples and UT Physicians, respectively, with the backbone frozen. The server aggregates their updates into a global adapter, which we evaluate on the held-out test set from each source and on the external i2b2 test set. **c,** Direct continuation updates the public adapter (blue P), whereas residual adaptation freezes it (gray P) and trains a separate adapter (R).

## Public biomedical preparation improves federated clinical IE, particularly with limited local training

Public biomedical preparation consistently improved clinical IE at matched local training budgets, with the largest gains observed when local training was limited. For NER, one-round SFT-Cont achieved higher mean source F1 than No-Public at all five training budgets on both LLM backbones (Fig. 2a–c, Table 2 and Supplementary Table S2). At 30 steps per client, SFT-Cont improved mean source F1 by 14.81 points for Qwen3.5 (95% CI, 11.40 to 19.20) and 18.68 points for Llama3.1 (95% CI, 15.29 to 24.19). As local training increased, these gains progressively narrowed, decreasing to 0.80 points for Qwen3.5 and 1.12 points for Llama3.1 at 480 steps per client.

SFT-Cont also consistently outperformed one-round Fed-MedLoRA+ at matched training budgets (Fig. 2e and Supplementary Table S3). At 30 steps per client, the gains were 15.75 points for Qwen3.5 (95% CI, 12.74 to 19.83) and 19.07 points for Llama3.1 (95% CI, 15.73 to 24.39). The differences again decreased with additional local training, reaching 1.54 points for Qwen3.5 and 2.33 points for Llama3.1 at 480 steps. Together, these results indicate that public preparation provides the greatest benefit when institutions have limited local training capacity, whereas its relative advantage diminishes as more private clinical optimization is performed.

**Fig. 2. NER performance and paired gains at matched training budgets.**

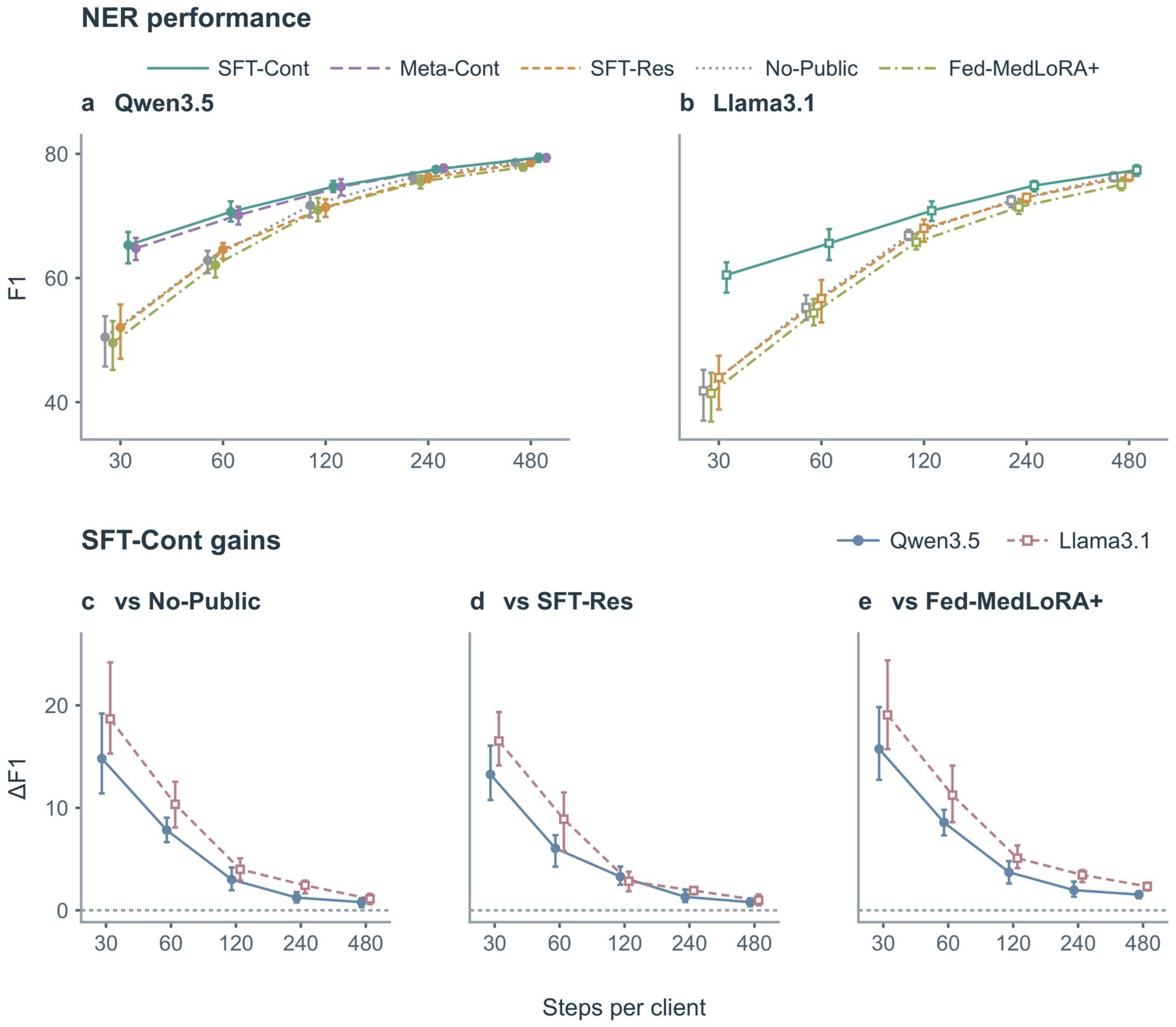

**a, b,** Mean source F1 for Qwen3.5-9B and Llama-3.1-8B-Instruct, respectively. **c–e,** SFT-Cont minus No-Public, SFT-Res, and Fed-MedLoRA+, respectively. Source F1 equally weights the three source test sets and is averaged over nine seed combinations. Error bars show pointwise 95% hierarchical bootstrap confidence intervals for absolute F1 and paired hierarchical bootstrap confidence intervals for differences. All runs use one communication round. Meta-Cont was evaluated on Qwen3.5. Positive differences favor SFT-Cont.

## Direct continuation outperforms residual adaptation

We next examined how the publicly prepared adapter should be incorporated into local clinical training. Direct continuation consistently outperformed residual adaptation, indicating that allowing clinical data to update the publicly learned adapter was more effective than keeping the public adapter fixed and learning a separate residual adapter.

For SFT preparation, SFT-Cont achieved higher mean source NER F1 than SFT-Res at all five training budgets on both LLM backbones (Fig. 2d and Table 2). At 30 steps per client, the gains were 13.26 points for Qwen3.5 (95% CI, 10.76 to 16.07) and 16.52 points for Llama3.1 (95% CI, 14.14 to 19.35). As local training increased, the differences narrowed to 0.80 and 0.99 points, respectively, at 480 steps per client.

The same pattern was observed with aggregation-aware meta-learning. Meta-Cont achieved higher mean source F1 than Meta-Res for both NER and RE across all evaluated training budgets and exceeded No-Public in nearly all comparisons (Supplementary Tables S3 and S5). Meta preparation produced higher NER F1 than SFT before clinical adaptation, but after direct continuation the two approaches achieved broadly similar performance, with Meta showing advantages in some RE settings. These results suggest that the main benefit of public preparation is best retained when the prepared adapter remains trainable during subsequent clinical adaptation.

### Public preparation improves relation extraction and external generalization

The benefits of public preparation extended to RE and to the external i2b2 test set. For RE, SFT-Cont achieved higher mean source F1 than No-Public when local training was limited. At 30 steps per client, the gains were 10.61 points for Qwen3.5 (95% CI, 3.75 to 19.17) and 8.92 points for Llama3.1 (95% CI, 1.04 to 17.07). SFT-Cont also outperformed SFT-Res by 8.76 points for Qwen3.5 and 8.70 points for Llama3.1 at the same training budget (Supplementary Fig. S2b). On Qwen3.5, both SFT-Cont (Supplementary Fig. S2c) and Meta-Cont achieved higher mean source RE F1 than one-round Fed-MedLoRA+ across all five matched training budgets (Supplementary Table S5). As with NER, the advantage over No-Public narrowed with additional local training (Supplementary Fig. S2a).

Public preparation also improved performance on the independent i2b2 test set. At 30 steps per client, SFT-Cont improved external NER F1 over No-Public by 14.80 points for Qwen3.5 and 15.85 points for Llama3.1 (Supplementary Figs. S1 and S4). The corresponding RE gains were 16.50 and 8.01 points, respectively (Supplementary Fig. S3 and Supplementary Tables S4 and S5). These results indicate that the benefits of public preparation were not limited to the source datasets used during federated adaptation.

### Public preparation can reduce private training and federated communication

We next examined whether public preparation could reduce the amount of private clinical training required to achieve comparable or better extraction performance. At the lowest training budget, one-round SFT-Cont using 30 steps per client achieved higher mean source NER F1 than No-Public and Fed-MedLoRA+ using 60 steps per client, whether the 60 steps were completed in one or two communication rounds (Fig. 3 and Table 3). Compared with one-round Fed-MedLoRA+, SFT-Cont improved mean source F1 by 3.22 points for Qwen3.5 and 6.14 points for Llama3.1. Compared with two-round Fed-MedLoRA+, the corresponding gains were 4.58 and 4.76 points. Thus, in this low-training setting, public preparation achieved better performance with half as many private optimization steps and, relative to the two-round schedules, one fewer federated communication round.

The efficiency advantage decreased as the amount of local training increased. With 60 steps per client, SFT-Cont remained within 1.4 F1 points of the corresponding baselines trained with 120 steps on both backbones. At 120 steps per client, however, baselines trained with 240 steps achieved higher mean source F1, although the gaps against one-round Fed-MedLoRA+ remained below one F1 point (Fig. 3 and Table 3). These results further indicate that public preparation is most useful when private training resources are limited.

The reduction in private optimization also translated to shorter clinical-stage training time. On a single NVIDIA H200 GPU, training all three clients with SFT-Cont for 30 steps per client required 11.40 minutes for Qwen3.5 and 2.06 minutes for Llama3.1, compared with 23.20 and 4.77 minutes, respectively, for two-round Fed-MedLoRA+ using 60 total steps per client (Supplementary Table S16). After initial distribution of the public adapter, the one-round schedule also exchanged half as much adapter data as the corresponding two-round schedules (Table 3).

**Fig. 3. NER performance with half as many private training steps.**

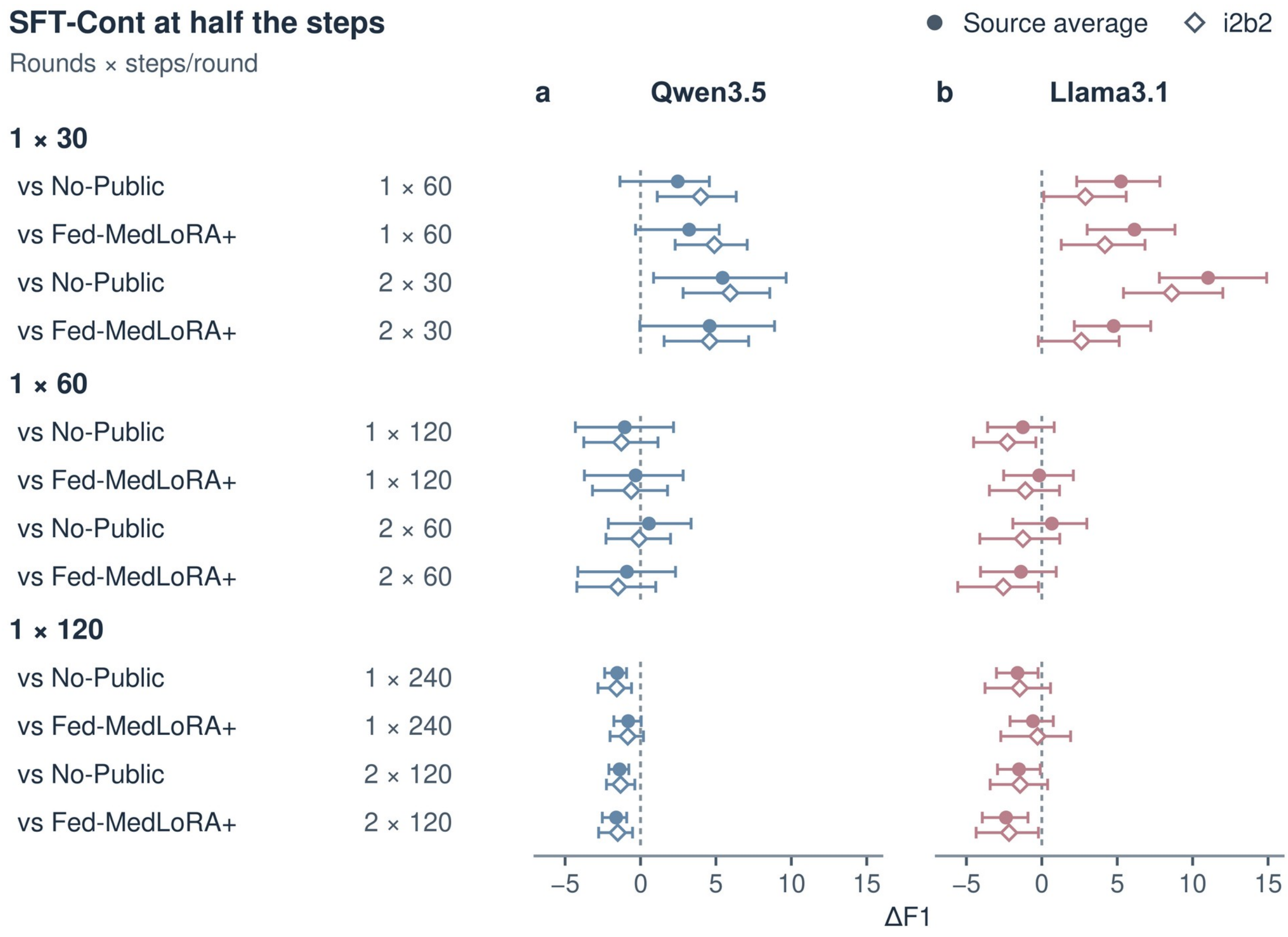

**a,** Qwen3.5. **b,** Llama3.1. Each group compares one round of SFT-Cont with one- and two-round No-Public and Fed-MedLoRA+ baselines trained for twice as many private steps per client. Schedules show rounds × steps per round. Points show SFT-Cont minus the named baseline, with filled circles for mean source NER F1 and open diamonds for i2b2 F1. Positive differences favor SFT-Cont. Error bars show pointwise 95% paired hierarchical bootstrap confidence intervals across nine paired settings.

## Federated aggregation provides additional gains when local training is limited

We next separated the contributions of public preparation and subsequent federated aggregation. Public preparation improved local NER performance before aggregation, particularly at low training budgets. At 30 steps per client, local SFT-Cont adapters achieved F1 scores 13.23 points higher for Qwen3.5 and 13.82 points higher for Llama3.1 than their corresponding local No-Public adapters (Fig. 4a).

Federated aggregation provided additional gains beyond public preparation when local training was limited. At 30 steps per client, the global SFT-Cont adapter outperformed the corresponding local adapters by 3.56 F1 points for Qwen3.5 (95% CI, 0.78 to 7.77) and 4.06 points for Llama3.1 (95% CI, 1.01 to 6.41). These gains diminished with additional local training; at 120 steps per client, the global-minus-local differences were 0.18 points for Qwen3.5 and −0.67 points for Llama3.1 (Fig. 4b and Supplementary Table S9).

The effect of aggregation also varied across source datasets. At 60 steps per client, for example, the global Llama3.1 adapter improved F1 on MIMIC-III by 10.96 points but decreased F1 on UT Physicians by 1.45 points relative to the corresponding local adapters (Fig. 4c). With the trained client adapters held fixed, alternative aggregation schemes produced similar mean source F1, differing by no more than 0.27 points (Supplementary Fig. S5 and Supplementary Tables S10 and S11).

Finally, increasing the number of communication rounds did not improve performance when the total local training budget was held fixed. With 120 total steps per client, two rounds of 60 steps produced slightly lower mean source F1 than one round of 120 steps for both Qwen3.5 and Llama3.1 (Fig. 4d and Table 3). Together, these results show that public preparation provides substantial gains before aggregation, while the additional benefit of federation is greatest under limited local training and can vary across institutions.

**Fig. 4. What public preparation and federation add to clinical NER.**

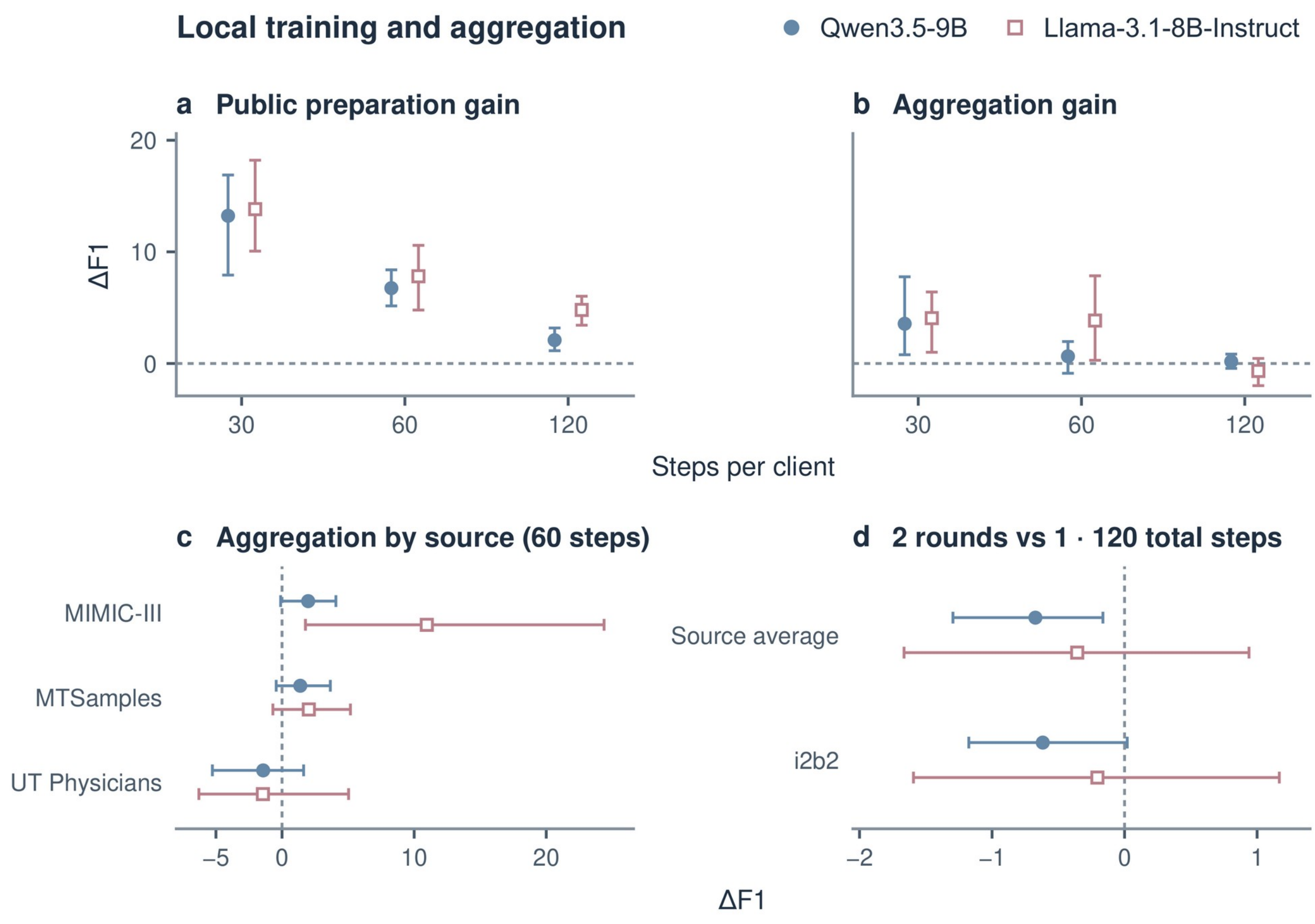

**a,** Public preparation gain in local models (SFT-Cont minus No-Public). **b,** Aggregation gain (global SFT-Cont minus the corresponding local adapters). **c,** The aggregation contrast in b by source at 60 steps per client. **d,** Two rounds of 60 steps minus one round of 120 steps for SFT-Cont, with the total fixed at 120 steps per client. In a and b, each local adapter is evaluated on its own source and the three source effects are weighted equally. Error bars show pointwise 95% paired hierarchical bootstrap confidence intervals across nine paired settings.

## Discussion

In this study, we investigated whether public biomedical supervision can provide a better starting point for federated LLM adaptation in clinical information extraction. Across two LLM backbones, NER and RE tasks, multiple local training budgets, and an independent external test set, public adapter preparation consistently provided the greatest benefit when local training was limited. Direct continuation of the prepared adapter outperformed keeping the public adapter fixed and learning a separate residual adapter. At low training budgets, public preparation also achieved comparable or better performance with fewer private optimization steps and, relative to multi-round schedules, fewer federated communication rounds. Collectively, these findings suggest that part of the task learning required for federated clinical IE can be moved from institution-specific training to a reusable public preparation stage.

The largest gains from public preparation occurred when local clinical training was limited, and these gains progressively narrowed as more private optimization was performed. Clinical notes differ from public biomedical text in wording and annotation conventions,[20] but public corpora contain related entity and relation supervision that can provide useful task knowledge before institutions begin training on their own data.[17,18] A task-naive adapter must learn these extraction capabilities primarily from institution-specific annotations, whereas a publicly prepared adapter can use local updates to adapt previously learned capabilities to clinical language and site-specific annotation schemes. As local training increases, adapters without public preparation have more opportunities to learn from the clinical examples, narrowing the relative advantage of public preparation. The improvements observed on the independent i2b2 test set further show that these benefits were not restricted to the source datasets used during federated adaptation.

Our findings complement prior work on parameter-efficient and multicenter LLM adaptation. Fed-MedLoRA established a parameter-efficient framework for federated LLM adaptation in clinical IE,[5] while FedDPA explored the reuse of global and local adapters for personalization.[21] More recently, multicenter methods such as MEDAL have adapted clinical LLMs by transferring lightweight adapters sequentially across institutions without sharing patient-level data.[10,22] These approaches primarily address how private institutional updates can be learned, exchanged, or retained. Pub2FedIE addresses a complementary question: whether task-relevant knowledge can be learned from public biomedical annotations before private clinical adaptation begins, thereby reducing the amount of institution-specific training subsequently required.

How the public adapter was incorporated into clinical training also mattered. Direct continuation consistently outperformed residual adaptation from the same public checkpoint, particularly at low training budgets. Direct continuation allows clinical examples to revise the parameters learned from public biomedical data, whereas residual adaptation preserves those parameters and learns a separate correction. The results suggest that public biomedical supervision provides a useful starting point that benefits from continued adaptation. This pattern was observed with both SFT and Meta preparation and extended to RE, despite differences between public and clinical relation schemas. Meta produced higher NER F1 before clinical training and advantages in some RE settings, but its advantage over SFT largely diminished after direct continuation. Meta also required approximately 5.6 times as many GPU-hours for public preparation as SFT (Supplementary Table S6). These findings make SFT a more economical default when preparing a new reusable adapter, while Meta may be useful when stronger performance is desired before institution-specific adaptation.

A central motivation for Pub2FedIE is to reduce reliance on repeated private clinical training. Public preparation is performed on shareable biomedical annotations before clinical federation, and the resulting adapter can subsequently be distributed to participating institutions. In the low-budget comparisons, one-round SFT continuation with 30 private steps per client outperformed No-Public and Fed-MedLoRA+ baselines trained with twice as many private steps. Compared with two-round schedules, it additionally required one fewer communication round and approximately half the clinical-stage fitting time in the targeted timing experiments. These benefits diminished as local training increased, and public preparation did not consistently outperform baselines using twice as many private steps at higher budgets. Public preparation should therefore be viewed as a strategy for improving the efficiency of institution-specific adaptation, particularly when local training resources are limited.

The local-versus-global analyses further separated the contributions of public preparation and federation. Clients benefited from public preparation before aggregation, and aggregation provided additional gains when local training was limited. However, these gains varied across sources. For example, aggregation substantially improved performance on MIMIC-III in one setting while slightly reducing performance on UT Physicians. The same number of optimization steps also represents different proportions of the available training data at each site, and differences in documentation and label distributions may further influence what individual institutions gain from federation.[4] The tested aggregation rules yielded similar mean source F1 when the underlying client adapters were held fixed, and splitting a fixed training budget across additional communication rounds did not improve overall performance. These findings suggest that increasing the number or complexity of federated exchanges is not necessarily beneficial and that institutions should evaluate both local and global adapters on their own validation data.

The error analyses provide additional context for how these improvements may translate to clinical IE applications. For NER, direct continuation improved both precision and recall compared with No-Public and residual adaptation and reduced false-positive extractions relative to residual adaptation. In applications such as cohort identification, higher recall could help retrieve more relevant patient records, while higher precision could reduce the proportion of spurious candidates requiring subsequent review. Some RE configurations recovered more annotated targets while also generating more false positives (Supplementary Tables S12 and S13), highlighting that improvements in overall F1 may reflect different precision-recall trade-offs across extraction tasks. NER requires identification of entity spans and categories, whereas the RE setting used here required extraction of a related target and assignment of its relation label for a supplied primary entity.

Several limitations and future directions should be considered. First, the current study represented three clinical data sources as federated clients rather than evaluating Pub2FedIE in a prospective multi-institutional deployment. Real-world collaborations may introduce additional heterogeneity in annotation practices, infrastructure, computational resources, and participation patterns. Second, we used six curated public biomedical datasets for adapter preparation. Future studies should systematically evaluate broader and larger-scale public biomedical resources, including automatically annotated resources[23,24], as well as alternative preparation strategies[25,26] and error-guided model refinement[27] to assess how the scale, annotation quality, and preparation algorithm affect subsequent federated clinical adaptation. Third, the LLM landscape is rapidly evolving. We evaluated two representative LLM backbones, and further evaluation with newer and larger model families will be important to establish the generalizability of the findings. To facilitate such evaluation and extension, we make the Pub2FedIE implementation publicly available. Finally, public preparation does not eliminate privacy risks during subsequent federated training; updates derived from clinical records may still expose information about local training data.[28,29] Future studies could explore Pub2FedIE with privacy-enhancing mechanisms.[30]

## Methods

### Study overview

Pub2FedIE consists of three stages: public biomedical adapter preparation, local adaptation on private clinical data, and federated aggregation (Fig. 1). We first prepared shared LoRA adapters using publicly available biomedical NER and RE annotations. The prepared adapters were then distributed to participating clients for task-specific adaptation on their local clinical data, followed by server aggregation. We evaluated supervised fine-tuning (SFT) as the primary public preparation strategy and aggregation-aware meta-learning (Meta) as an alternative preparation strategy. During clinical adaptation, we compared direct continuation of the prepared adapter with residual adaptation that kept the public adapter fixed and trained an additional LoRA adapter.

## Clinical datasets and information extraction tasks

We used the clinical information extraction annotations developed by Hu et al.[1] We simulated three clients, each assigned one dataset: MIMIC-III, MTSamples, or UT Physicians (UTP). Each client trained on records from its assigned dataset. We retained the supplied training and test partitions and evaluated the global adapter on each source's test set. A subset of the 2010 i2b2 corpus, reannotated by Hu et al. under the same clinical schema, provided an external test collection. The i2b2 records were excluded from training and aggregation-weight selection. Table 1 reports annotation-record counts; a clinical note can contribute multiple records.

For named-entity recognition (NER), we followed Hu et al.'s four categories: problem, test, drug, and treatment. For relation extraction (RE), we supplied the primary entity and its location in the text. The model extracted related target strings and assigned one of 16 modifier labels, including dosage, negation, and temporal information. We trained separate clinical adapters for NER and RE.

## Public biomedical datasets

Corpus selection drew on the biomedical dataset inventories in BigBio and BLURB.[31,32] We considered English-language corpora with accessible training annotations for entity recognition or relation extraction that could be represented through our extraction prompts. We selected six corpora with complementary coverage of diseases, drugs, molecular entities, biomedical relations, and clinical case narratives.[11–16]

BC5CDR supplied chemical and disease mentions together with chemical-induced disease relations. NCBI Disease added disease annotations from a separate resource. DrugProt contributed entity annotations and chemical–protein relations, while BioRED broadened coverage to document-level relations involving several biomedical entity types. DDIExtraction 2013 supplied drug–drug interactions. The English portion of E3C added clinical case narratives and test–result annotations. These corpora provided six NER and five RE task views, each retaining its original label schema (Supplementary Table S1).

## Public adapter preparation

We evaluated SFT on Qwen3.5-9B (Qwen3.5) and Llama-3.1-8B-Instruct (Llama3.1).[33,34] Meta was evaluated on Qwen3.5. Both preparation methods kept the backbone frozen and trained the LoRA factors. Model, training, and evaluation settings are summarized in Table 4.

For each public initialization and task view, we drew disjoint pools of 16 support and 16 query sampling units from the public training split. We kept these pools fixed throughout preparation. SFT used support examples to train the shared adapter. Meta used support examples for simulated client adaptation and query examples to update the shared adapter.

SFT trained the shared adapter directly on the public annotations. Training alternated between NER and RE. Each update sampled three distinct views from the current task and one support example from each view, then minimized the mean target-token cross-entropy. Prompt and padding tokens were masked from the loss.

Meta used the same public task views as simulated clients.[35,36] In each episode, we froze the shared adapter and fitted a fresh rank-16 residual adapter to support examples from each of three views. Successive episodes within each task alternated between two and five inner updates. We averaged the effective residual updates, projected the average to rank 16, and combined it with the shared adapter to evaluate held-out query examples. This strategy was designed to prepare the shared adapter for subsequent federated adaptation.

In each Meta episode, d indexes the three sampled public task views. Let $l_d^{post}$ denote the query target-token cross-entropy for view d after simulated adaptation and aggregation, and $l_d^{pre}$ the corresponding query loss of the public adapter alone before simulated adaptation. The Meta objective was

$$L_{\text{Meta}} = \frac{1}{3}\sum_{d=1}^{3} l_d^{\text{post}} + 0.5\ \max_{\text{d}} l_d^{\text{post}} + 0.1\left(\frac{1}{3}\sum_{d=1}^{3} l_d^{\text{pre}}\right) \quad (1)$$

The terms are the mean post-aggregation query loss, the query loss of the highest-loss sampled view, and the mean query loss before simulated adaptation. We selected the highest-loss view with dropout disabled. Outer gradients updated the shared adapter while treating the fitted residuals and their aggregate as fixed. Both preparation methods completed 2,000 shared-adapter updates, alternating between NER and RE. Clinical training used the final prepared adapter.

## Clinical adaptation and federated aggregation

With direct continuation, each client initialized its adapter from the shared public adapter and updated its LoRA factors on its own clinical records (SFT-Cont or Meta-Cont). With residual adaptation, each client froze the public adapter and trained a separate rank-16 adapter (SFT-Res or Meta-Res). No-Public used newly initialized adapter factors. In all three settings, clients kept the backbone frozen and minimized target-token cross-entropy on the clinical examples.

At the end of each round, the server averaged the three clients' effective LoRA updates with equal weights and projected the result to rank 16 using truncated singular-value decomposition. Let $W_0$ denote a frozen backbone weight matrix, P the prepared public update, $C_i$ the complete continued update from client i, and $R_i$ its residual update, with i indexing the three clinical clients. The effective global weight matrices of an adapted layer under continuation (Cont) and residual adaptation (Res) were

$$W_{\text{Cont}} = W_0 + \Pi_{16}\left[\frac{1}{3}\sum_{i=1}^{3} C_i\right] \quad (2)$$

$$W_{\text{Res}} = W_0 + P + \Pi_{16}\left[\frac{1}{3}\sum_{i=1}^{3} R_i\right] \quad (3)$$

where $\Pi_{16}$ denotes rank-16 projection. The continued updates already included the public initialization. Continuation therefore retained one rank-16 adapter, whereas residual adaptation retained the frozen public adapter and a rank-16 residual. No-Public used the same aggregation rule as continuation.

Averaging effective LoRA updates before low-rank reconstruction has also been used in FlexLoRA.[37] FedEx-LoRA addresses the discrepancy between factor-wise averaging and update averaging through exact error compensation.[38]

## Training budgets and experimental comparisons

We defined the clinical training budget as the total number of optimizer steps per client across all communication rounds. Because the client training sets differed in size, we matched the number of optimizer steps rather than epochs across clients. We evaluated 30, 60, 120, 240, and 480 steps per client, with eight examples per step. Within each paired setting, matched comparisons used the same ordered clinical examples. Public preparation was accounted for separately.

For NER and RE, we compared direct continuation and residual adaptation from the same public adapter, with No-Public as a common baseline. All three were evaluated after one round at each of the five training budgets. We also evaluated one-round Fed-MedLoRA+ across all five budgets for NER on both backbones and for RE on Qwen3.5.

For NER, we compared one-round SFT-Cont with one- and two-round No-Public and Fed-MedLoRA+ baselines that used twice as many private steps per client. For example, one round of 30 steps was compared with one round of 60 steps or two rounds of 30 steps (Table 3).

Fed-MedLoRA+ used newly initialized adapters and the same backbones, prompts, and scoring procedures as the other methods.[5] The server averaged LoRA factors separately and weighted each client in proportion to the exponential of its negative validation cross-entropy. Each paired setting used a 96-record validation panel, with 32 records from each clinical training partition. Panel texts were withheld from the corresponding training sequence.

All clients participated in every round and began each round with a fresh optimizer. In two-round runs, clients received the first-round global adapter and continued through their local example sequences.

To distinguish the benefits of public preparation and aggregation, we evaluated local SFT-Cont and No-Public adapters at 30, 60, and 120 NER steps on their own source test sets (Supplementary Table S8). We then evaluated the corresponding global adapters on the same test sets. Holding the trained client adapters fixed, we compared four aggregation rules that crossed equal or validation-loss-based weighting with effective-matrix or separate-factor averaging. To assess communication rounds at a fixed private budget, we compared one round of 120 steps with two rounds of 60 steps for SFT-Cont (Table 3).

## Evaluation and statistical analysis

For each clinical task, the model generated a JSON prediction using greedy decoding. Table 4 lists the decoding settings. The clinical prompts and output schemas are provided in Supplementary Section S1.

A NER prediction had to match the reference span and entity type exactly. The predicted occurrence index distinguished repeated strings in the same input. For RE, predicted target strings and modifier labels were matched exactly within each supplied primary-entity instance, with repeated targets counted separately. Invalid JSON or schema outputs counted as missed gold targets and one additional false positive.

We calculated precision, recall, and micro-F1 on each test collection. Within each paired setting, mean source F1 equally weighted the three source-specific micro-F1 scores. We then averaged over settings and reported i2b2 separately. F1 is on a 0–100 scale, and differences are expressed in F1 points. Supplementary Tables S14 and S15 report label-specific F1.

Three initialization blocks crossed with three private training seeds yielded nine paired settings. We used 20,000 hierarchical bootstrap draws to calculate percentile 95% confidence intervals. Each draw resampled initialization blocks and then private training runs within each selected block. We used the same resampled indices for methods and budgets in each paired contrast.

Intervals are pointwise and describe training variation on the fixed test collections. In a sensitivity analysis, we used the three initialization-block mean effects to calculate t-based intervals with two degrees of freedom (Supplementary Table S17).

## Computational resource accounting

We measured clinical computation and communication after public preparation and initial adapter distribution. We accounted for public preparation separately, once per adapter, and included Meta's inner updates and query evaluations.

We calculated communication volume from the FP32 adapter factors, counting three uploads and three downloads per round. For the clinical timing comparisons, the three clients trained sequentially on one NVIDIA H200 GPU. We timed client fitting, server validation, aggregation, and evaluation separately. Table 3 reports clinical training budgets and communication volume. Supplementary Section S6 reports the timing and public preparation costs.

## Data availability

Hu et al.[1] provide the annotated MTSamples data in the Kiwi-LLaMA repository. For MIMIC-III and the reannotated 2010 i2b2 subset, the repository provides annotation files with entity text and character offsets. Researchers can apply to PhysioNet for the original MIMIC-III notes and to the n2c2 Data Portal for the portion of the 2010 i2b2 corpus available for research, subject to each provider's data use agreement. UT Physicians records are not publicly available because of patient privacy restrictions. The public corpora used for adapter preparation are available through their cited resources.[11–16] Aggregate extraction scores, paired comparisons, and computational resource measurements are provided in this article and its Supplementary Information.

## Code availability

The Pub2FedIE code for public adapter preparation, federated adaptation, evaluation, and paired statistical analysis is publicly available at https://github.com/Yale-BIDS-Chen-Lab/Pub2FedIE.

## Acknowledgements

This study is supported by the National Institutes of Health grant 1R01LM014604.

## Author contributions

T.C. and Q.C. conceived the study. T.C. conducted the experiments and analyses. A.L. and Q.C. contributed to the evaluation and discussion. T.C., A.L., and Q.C. contributed to manuscript writing. H.X. contributed to clinical data resources and data curation. All authors approved the final manuscript.

## Competing interests

The authors declare no competing interests.

**Table 1. Clinical annotation records used for training and evaluation.**

| Collection | NER training | NER test | RE training | RE test |
|---|---|---|---|---|
| MIMIC-III | 4,260 | 4,049 | 6,117 | 6,446 |
| MTSamples | 5,211 | 2,885 | 6,919 | 3,882 |
| UT Physicians | 135,716 | 5,778 | 156,523 | 6,465 |
| i2b2 | — | 4,809 | — | 3,847 |

A clinical note may contribute multiple annotation records. i2b2 was used only for testing.

**Table 2. One-round NER performance and paired gains at matched training budgets.**

| Model | Steps/ client | No-Public | SFT-Res | SFT-Cont | Cont − No-Public [95% CI] | Cont − Res [95% CI] |
|---|---|---|---|---|---|---|
| Qwen3.5 | 30 | 50.50 | 52.05 | 65.31 | +14.81 [11.40, 19.20] | +13.26 [10.76, 16.07] |
| | 60 | 62.84 | 64.61 | 70.66 | +7.82 [6.65, 9.04] | +6.05 [4.26, 7.33] |
| | 120 | 71.71 | 71.42 | 74.71 | +3.00 [1.96, 4.18] | +3.29 [2.49, 4.27] |
| | 240 | 76.26 | 76.19 | 77.50 | +1.24 [0.77, 1.78] | +1.31 [0.80, 1.99] |
| | 480 | 78.60 | 78.60 | 79.40 | +0.80 [0.31, 1.19] | +0.80 [0.39, 1.17] |
| Llama3.1 | 30 | 41.82 | 43.98 | 60.50 | +18.68 [15.29, 24.19] | +16.52 [14.14, 19.35] |
| | 60 | 55.26 | 56.70 | 65.59 | +10.34 [8.08, 12.54] | +8.89 [5.78, 11.50] |
| | 120 | 66.85 | 68.01 | 70.85 | +4.00 [2.92, 5.07] | +2.84 [1.86, 3.77] |
| | 240 | 72.47 | 72.98 | 74.90 | +2.44 [1.65, 2.91] | +1.93 [1.56, 2.24] |
| | 480 | 76.27 | 76.40 | 77.39 | +1.12 [0.60, 1.61] | +0.99 [0.49, 1.55] |

F1 scores (0–100) equally weight the three sources and are averaged across nine paired settings. Brackets show pointwise 95% paired hierarchical bootstrap confidence intervals.

**Table 3. NER performance and clinical resources by training schedule.**

| Procedure | Total steps/client | Rounds × steps/round | Qwen3.5 source F1 | Llama3.1 source F1 | Clinical traffic (MiB) Qwen3.5 / Llama3.1 |
|---|---|---|---|---|---|
| **A. SFT-Cont 30 vs controls at 60 steps** | | | | | |
| SFT-Cont | 30 | 1 × 30 | 65.31 | 60.50 | 90 / 87.75 |
| No-Public | 60 | 1 × 60 | 62.84 | 55.26 | 90 / 87.75 |
| Fed-MedLoRA+ | 60 | 1 × 60 | 62.09 | 54.36 | 90 / 87.75 |
| No-Public | 60 | 2 × 30 | 59.87 | 49.48 | 180 / 175.5 |
| Fed-MedLoRA+ | 60 | 2 × 30 | 60.73 | 55.74 | 180 / 175.5 |
| **B. SFT-Cont 60 vs controls at 120 steps** | | | | | |
| SFT-Cont | 60 | 1 × 60 | 70.66 | 65.59 | 90 / 87.75 |
| No-Public | 120 | 1 × 120 | 71.71 | 66.85 | 90 / 87.75 |
| Fed-MedLoRA+ | 120 | 1 × 120 | 70.99 | 65.77 | 90 / 87.75 |
| No-Public | 120 | 2 × 60 | 70.10 | 64.93 | 180 / 175.5 |
| Fed-MedLoRA+ | 120 | 2 × 60 | 71.56 | 66.98 | 180 / 175.5 |
| **C. SFT-Cont 120 vs controls at 240 steps** | | | | | |
| SFT-Cont | 120 | 1 × 120 | 74.71 | 70.85 | 90 / 87.75 |
| No-Public | 240 | 1 × 240 | 76.26 | 72.47 | 90 / 87.75 |
| Fed-MedLoRA+ | 240 | 1 × 240 | 75.53 | 71.44 | 90 / 87.75 |
| No-Public | 240 | 2 × 120 | 76.10 | 72.37 | 180 / 175.5 |
| Fed-MedLoRA+ | 240 | 2 × 120 | 76.33 | 73.23 | 180 / 175.5 |
| **D. Fixed total of 120 steps: one vs two rounds** | | | | | |
| SFT-Cont | 120 | 1 × 120 | 74.71 | 70.85 | 90 / 87.75 |
| SFT-Cont | 120 | 2 × 60 | 74.04 | 70.50 | 180 / 175.5 |

Source F1 is averaged across nine paired settings, with sources weighted equally. Traffic counts three FP32 adapter uploads and three downloads per round and excludes initial public-adapter distribution. Paired confidence intervals for blocks A–C are reported in Supplementary Table S7.

**Table 4. Model, training, and evaluation settings.**

| Setting | Value |
|---|---|
| Backbone and adapter precision | Frozen BF16 backbone; FP32 LoRA factors |
| LoRA targets | Query, key, value, and output projections |
| Qwen3.5 adapted layers | 3, 7, 11, 15, 19, 23, 27, 31 |
| Llama3.1 adapted layers | 0, 4, 8, 12, 16, 19, 23, 27, 31 |
| LoRA rank / scaling parameter / dropout | 16 / 32 / 0.05 |
| Public outer and clinical optimizer | AdamW; weight decay 0.01; gradient clipping 1.0 |
| Learning rates and schedule | $10^{-5}$ for the shared public adapter and $10^{-4}$ for the Meta inner loop and clinical adaptation. Constant without warm-up. |
| Public sampling units | Documents for BC5CDR, NCBI Disease, DrugProt, and BioRED. Document-grouped sentences for DDIExtraction. Loader records for E3C NER and text identifiers for E3C RE. |
| SFT sampling | Preference alternated between nonempty and empty targets in four-update blocks. If the preferred type was unavailable, we sampled an available example from the selected unit. |
| Generation | Each backbone’s native chat template. Qwen3.5 in non-thinking mode. Maximum 4,096 input tokens and 2,560 new tokens. |
| Batch sizes and server validation | Clinical training microbatch size 1. Evaluation batch size 8. Server validation batch size 8 (dropout disabled). |

# Supplementary Information

## Pub2FedIE: Public biomedical adapter preparation for efficient federated clinical information extraction with large language models

Tianle Cao, Anran Li, Hua Xu, Qingyu Chen

Qwen3.5 and Llama3.1 refer to Qwen3.5-9B and Llama-3.1-8B-Instruct, respectively. Unless otherwise stated, means summarize nine paired settings from three initialization blocks crossed with three private training seeds. F1, precision, and recall use a 0–100 scale; differences are in F1 points. Source equally weights MIMIC-III, MTSamples, and UT Physicians (UTP); i2b2 is reported separately. Intervals are pointwise 95% confidence intervals (CIs). Hierarchical bootstrap intervals are paired for contrasts; block-t sensitivity intervals are defined in Section S7. Budgets are total optimizer steps per client across communication rounds.

### Contents

Corpus citations follow the main article reference numbering.

## S1 Datasets, tasks, and experimental coverage

The public corpora provide six task views for named-entity recognition (NER) and five for relation extraction (RE). Clinical extraction uses four entity labels and sixteen modifier labels.

### Supplementary Table S1. Public corpus selection and extraction views

| Corpus | Role | Text source | Selected task views or candidate annotations | Selection rationale |
|---|---|---|---|---|
| BC5CDR[11] | Selected | PubMed titles/abstracts | NER: chemical entities; disease entities. RE: chemical-induced disease. | Provides related entity and relation views in one document-based resource. |
| NCBI Disease[12] | Selected | PubMed abstracts | NER: disease mentions. | Adds disease annotations from a separate resource. The selected task uses mention extraction. |
| DrugProt[13] | Selected | English PubMed titles/abstracts | NER: native entities. RE: chemical–protein relations. | Adds molecular entities and typed chemical–protein relations. |
| BioRED[14] | Selected | PubMed abstracts | NER: native entities. RE: native document relations. | Broadens entity-pair and relation coverage beyond a single relation family. |
| DDIExtraction 2013 / DDI corpus[15] | Selected | DrugBank text and MEDLINE abstracts | RE: drug–drug interactions. | Adds pharmacological interaction supervision. The selected view is RE. |
| E3C 3.0, English portion[16] | Selected | Published clinical case narratives | NER: English clinical entities. RE: test-result relations. | Adds clinical narrative language and test–result supervision. |
| BC2GM[39] | Not included | Biomedical literature sentences | Gene mention spans | Gene-mention supervision overlaps the molecular-entity coverage in DrugProt and BioRED. It could add gene-name diversity rather than a new task family. |
| JNLPBA[40] | Not included | GENIA-derived MEDLINE abstracts | Protein, DNA, RNA, cell line, and cell type mentions | Additional cell and nucleic-acid categories were outside the selected views; the flattened spans also use different annotation conventions. |
| ChemProt, BioCreative VI[41] | Not included | PubMed abstracts | Chemical–protein interaction annotations | Chemical–protein relations are already represented by DrugProt. The two resources also share abstracts, so adding both would require overlap control. |
| GAD-derived gene–disease association corpus[42] | Not included | Biomedical literature sentences | Gene–disease association instances | Pair-classification annotations would need conversion to the target-span extraction format used here. |
| EBM-NLP / EBM PICO[43] | Not included | Clinical randomized-trial abstracts | Participant, intervention, comparator, and outcome spans | Broad PICO spans differ from the entity-mention granularity used here and would require a separate extraction view. |
| CHEMDNER / BC4CHEMD[44] | Not included | PubMed abstracts across chemical disciplines | Chemical and drug name mentions | Chemical and drug mentions are already covered by BC5CDR and DrugProt; this corpus adds no relation view. |
| AnatEM[45] | Not included | Biomedical literature documents | Anatomical entity mentions | Anatomy-focused annotation would add a separate entity family beyond the selected extraction views. |
| ADE benchmark corpus[46] | Not included | MEDLINE medical case reports | Drug and adverse-effect entities and relations | Drug–adverse-effect relations would add a separate task view; clinical case narratives are already represented by E3C. |

The inventory draws on BigBio, BLURB, and the cited corpus resources. The selected corpora provide six NER views and five RE views, shared by SFT and Meta.

### Evaluated configurations

One-round SFT-Cont, SFT-Res, and No-Public covered NER and RE at 30, 60, 120, 240, and 480 steps per client on both backbones. Meta-Cont and Meta-Res covered the same tasks and budgets on Qwen3.5. One-round Fed-MedLoRA+ covered NER on both backbones and RE on Qwen3.5 at all five budgets.

Two-round No-Public covered total budgets of 60, 120, and 240 steps per client for Llama3.1 NER and 60, 120, 240, and 480 steps for Qwen3.5 NER and RE. Two-round Fed-MedLoRA+ covered NER at 60, 120, and 240 total steps on both backbones. Two-round SFT-Cont used 120 total steps for NER on both backbones. Each two-round schedule split the total budget equally between rounds. All configurations used nine paired training settings.

### Clinical task prompts and output schemas

NER extracts exact spans and one-based occurrences from the space-joined token text. RE receives a primary entity, whose start is zero-based and whose end is exclusive, and extracts related targets with modifier labels. Repeated predictions retain their multiplicity in scoring.

### NER prompt

```
Extract every explicitly stated entity whose type belongs to the provided schema. Copy each span exactly from the text. To
distinguish repeated identical surface strings, occurrence is the one-based index of that exact, case-sensitive string when scanning
Text from left to right. Return only a valid top-level JSON array with exact item schema [{"text":"exact
span","type":"TYPE","occurrence":1}]. Return [] when none apply. Do not wrap the array in an object or Markdown.
Entity schema: {schema_json}
Text: {text}
```

The entity schema is:

```
["problem","test","drug","treatment"]
```

### RE prompt

```
Given the marked primary source and text, extract every related target and relation label from the provided schema. Source start/end
are zero-based character offsets in Text; end is exclusive. Source occurrence is the one-based exact-text occurrence in Text. These
identify the supplied source when identical mentions repeat. Copy target spans exactly. Return only valid JSON with exact schema
{"relations":[{"target":"exact span","relation":"TYPE"}]}. Return {"relations":[]} when none apply.
Relation schema: <compact JSON of the 16 labels below>
Source: {"text":<surface>,"type":<main entity type>,"start":<int>,"end":<int>,"occurrence":<one-based int>}
Text: <decoded original input annotation text>
```

Angle-bracket expressions denote substituted values. The relation schema is:

```
["bodyloc", "condition", "course", "dosage", "duration", "form", "frequency", "labvalue", "negation", "reference_range", "route",
"severity", "strength", "subject", "temporal", "uncertain"]
```

## S2 Extraction performance across training budgets

Absolute scores and paired contrasts are reported separately for NER and RE, including source-specific and external i2b2 results.

### Supplementary Table S2. NER performance across training budgets

| Model | Steps/client | Method | Source | MIMIC-III | MTSamples | UTP | i2b2 |
|---|---|---|---|---|---|---|---|
| Qwen3.5 | 30 | Fed-MedLoRA+ 1R | 49.56 [45.20, 53.07] | 56.30 [52.81, 58.88] | 51.93 [48.08, 54.84] | 40.46 [34.45, 45.69] | 48.72 [45.50, 51.30] |
| Qwen3.5 | 30 | Meta-Cont | 64.79 [62.91, 66.46] | 66.42 [65.16, 67.45] | 63.97 [62.70, 65.38] | 63.97 [60.51, 66.84] | 63.55 [61.85, 65.37] |
| Qwen3.5 | 30 | Meta-Res | 52.72 [47.28, 56.68] | 56.49 [54.38, 58.44] | 52.73 [48.37, 56.18] | 48.94 [38.05, 55.66] | 48.59 [44.99, 52.52] |
| Qwen3.5 | 30 | No-Public 1R | 50.50 [45.76, 53.85] | 57.00 [53.31, 59.30] | 52.73 [48.49, 55.47] | 41.77 [35.51, 46.96] | 49.40 [45.94, 51.94] |
| Qwen3.5 | 30 | SFT-Cont | 65.31 [62.37, 67.41] | 66.51 [64.96, 67.62] | 64.78 [62.54, 66.41] | 64.64 [59.54, 68.36] | 64.20 [61.75, 66.29] |
| Qwen3.5 | 30 | SFT-Res | 52.05 [47.00, 55.74] | 55.18 [50.50, 58.72] | 52.42 [48.21, 55.76] | 48.56 [41.93, 53.08] | 49.02 [46.01, 51.71] |
| Qwen3.5 | 60 | Fed-MedLoRA+ 1R | 62.09 [60.08, 63.52] | 63.88 [62.48, 64.89] | 61.45 [59.99, 62.42] | 60.94 [57.38, 63.84] | 59.31 [57.67, 60.53] |
| Qwen3.5 | 60 | Fed-MedLoRA+ 2R | 60.73 [56.28, 64.13] | 63.02 [59.16, 65.31] | 61.48 [59.42, 63.24] | 57.69 [49.48, 64.05] | 59.63 [57.25, 61.60] |
| Qwen3.5 | 60 | Meta-Cont | 70.14 [68.62, 71.50] | 69.70 [69.12, 70.41] | 70.05 [69.17, 70.97] | 70.67 [67.39, 73.72] | 69.25 [68.33, 70.20] |
| Qwen3.5 | 60 | Meta-Res | 63.59 [60.42, 66.20] | 64.27 [62.47, 65.66] | 62.77 [60.18, 65.33] | 63.74 [58.55, 67.71] | 60.44 [57.53, 62.98] |
| Qwen3.5 | 60 | No-Public 1R | 62.84 [60.80, 64.38] | 64.27 [62.87, 65.26] | 62.28 [60.69, 63.16] | 61.97 [58.39, 65.09] | 60.22 [58.49, 61.43] |
| Qwen3.5 | 60 | No-Public 2R | 59.87 [55.77, 63.01] | 62.46 [59.67, 64.53] | 60.27 [58.37, 61.85] | 56.88 [49.22, 62.71] | 58.26 [55.84, 60.18] |
| Qwen3.5 | 60 | SFT-Cont | 70.66 [69.10, 72.35] | 70.00 [69.22, 70.72] | 70.08 [69.24, 70.97] | 71.90 [68.66, 75.51] | 69.83 [68.79, 70.79] |
| Qwen3.5 | 60 | SFT-Res | 64.61 [63.12, 65.64] | 65.11 [63.97, 65.81] | 63.49 [62.48, 64.28] | 65.24 [62.94, 67.24] | 61.97 [60.61, 62.86] |
| Qwen3.5 | 120 | Fed-MedLoRA+ 1R | 70.99 [69.15, 72.91] | 70.01 [68.89, 70.95] | 71.80 [70.32, 73.34] | 71.16 [68.06, 74.40] | 70.45 [68.63, 72.20] |
| Qwen3.5 | 120 | Fed-MedLoRA+ 2R | 71.56 [69.47, 73.33] | 70.30 [68.57, 71.45] | 72.49 [70.84, 73.95] | 71.90 [68.60, 74.82] | 71.32 [69.21, 73.32] |
| Qwen3.5 | 120 | Meta-Cont | 74.71 [73.31, 75.93] | 72.93 [71.80, 73.96] | 75.17 [74.14, 76.16] | 76.03 [74.00, 77.73] | 74.61 [73.04, 76.01] |
| Qwen3.5 | 120 | Meta-Res | 71.27 [69.80, 72.41] | 70.09 [68.70, 71.34] | 71.51 [70.44, 72.35] | 72.20 [70.01, 74.00] | 70.62 [68.69, 72.56] |
| Qwen3.5 | 120 | No-Public 1R | 71.71 [69.76, 73.47] | 70.44 [69.20, 71.40] | 72.53 [71.22, 73.81] | 72.16 [68.76, 75.28] | 71.10 [69.22, 72.77] |
| Qwen3.5 | 120 | No-Public 2R | 70.10 [68.63, 71.33] | 69.60 [68.62, 70.37] | 70.86 [69.57, 71.97] | 69.84 [67.55, 71.76] | 69.94 [68.35, 71.50] |
| Qwen3.5 | 120 | SFT-Cont | 74.71 [73.85, 75.66] | 73.07 [72.19, 74.06] | 74.88 [74.07, 75.83] | 76.19 [74.94, 77.26] | 74.23 [72.88, 75.35] |
| Qwen3.5 | 120 | SFT-Cont 2R | 74.04 [73.13, 74.96] | 72.51 [71.61, 73.38] | 74.00 [73.07, 74.89] | 75.61 [74.57, 76.75] | 73.62 [72.31, 75.04] |
| Qwen3.5 | 120 | SFT-Res | 71.42 [69.85, 72.70] | 70.67 [69.36, 71.71] | 71.07 [69.40, 72.76] | 72.54 [70.63, 74.04] | 71.02 [68.72, 72.76] |
| Qwen3.5 | 240 | Fed-MedLoRA+ 1R | 75.53 [74.42, 76.50] | 73.67 [73.10, 74.22] | 76.12 [75.01, 76.98] | 76.81 [74.91, 78.36] | 75.08 [73.88, 75.93] |
| Qwen3.5 | 240 | Fed-MedLoRA+ 2R | 76.33 [75.66, 77.05] | 74.34 [74.04, 74.73] | 76.88 [76.09, 77.61] | 77.77 [76.62, 78.92] | 75.75 [75.03, 76.45] |
| Qwen3.5 | 240 | Meta-Cont | 77.71 [77.09, 78.25] | 75.35 [74.86, 75.95] | 78.00 [77.44, 78.66] | 79.77 [78.47, 80.70] | 77.84 [77.48, 78.17] |
| Qwen3.5 | 240 | Meta-Res | 76.21 [75.48, 76.89] | 74.18 [73.80, 74.66] | 76.53 [75.71, 77.33] | 77.92 [76.59, 78.97] | 75.84 [75.02, 76.71] |
| Qwen3.5 | 240 | No-Public 1R | 76.26 [75.51, 77.01] | 74.09 [73.65, 74.64] | 76.70 [75.74, 77.56] | 77.99 [76.87, 78.91] | 75.81 [74.88, 76.61] |

| Model | Steps/client | Method | Source | MIMIC-III | MTSamples | UTP | i2b2 |
|---|---|---|---|---|---|---|---|
| Qwen3.5 | 240 | No-Public 2R | 76.10 [75.58, 76.66] | 74.19 [73.87, 74.60] | 76.59 [76.15, 77.08] | 77.52 [76.54, 78.37] | 75.56 [74.92, 76.12] |
| Qwen3.5 | 240 | SFT-Cont | 77.50 [77.11, 77.90] | 75.39 [75.16, 75.72] | 77.89 [77.05, 78.48] | 79.23 [78.63, 79.78] | 77.43 [76.52, 78.22] |
| Qwen3.5 | 240 | SFT-Res | 76.19 [75.45, 76.84] | 74.30 [74.00, 74.64] | 76.16 [75.08, 77.13] | 78.12 [77.07, 79.03] | 75.87 [74.75, 76.67] |
| Qwen3.5 | 480 | Fed-MedLoRA+ 1R | 77.86 [77.32, 78.30] | 75.74 [75.41, 76.08] | 78.47 [78.19, 78.77] | 79.36 [77.76, 80.72] | 77.77 [77.22, 78.35] |
| Qwen3.5 | 480 | Meta-Cont | 79.40 [78.77, 79.90] | 76.89 [76.57, 77.26] | 80.07 [79.69, 80.42] | 81.23 [79.35, 82.70] | 79.57 [79.07, 80.15] |
| Qwen3.5 | 480 | Meta-Res | 79.10 [78.78, 79.42] | 76.66 [76.41, 76.92] | 79.73 [79.39, 80.11] | 80.90 [79.95, 82.08] | 79.12 [78.61, 79.63] |
| Qwen3.5 | 480 | No-Public 1R | 78.60 [78.31, 78.87] | 76.19 [75.93, 76.48] | 79.25 [78.89, 79.63] | 80.37 [79.37, 81.23] | 78.47 [77.98, 78.97] |
| Qwen3.5 | 480 | No-Public 2R | 78.97 [78.55, 79.30] | 76.51 [76.32, 76.68] | 79.54 [79.20, 79.85] | 80.85 [79.71, 81.64] | 78.53 [78.04, 79.03] |
| Qwen3.5 | 480 | SFT-Cont | 79.40 [78.70, 80.01] | 77.05 [76.68, 77.41] | 79.94 [79.52, 80.27] | 81.20 [79.28, 82.60] | 79.55 [78.95, 80.21] |
| Qwen3.5 | 480 | SFT-Res | 78.60 [78.07, 79.10] | 76.66 [76.28, 76.96] | 79.04 [78.73, 79.28] | 80.09 [78.57, 81.58] | 78.60 [78.04, 79.12] |
| Llama3.1 | 30 | Fed-MedLoRA+ 1R | 41.43 [36.88, 44.76] | 45.40 [40.69, 48.39] | 43.61 [39.67, 46.71] | 35.29 [30.04, 39.39] | 39.02 [34.89, 42.26] |
| Llama3.1 | 30 | No-Public 1R | 41.82 [37.03, 45.21] | 45.80 [40.59, 49.04] | 43.98 [40.29, 47.17] | 35.69 [30.13, 39.77] | 39.49 [35.26, 42.63] |
| Llama3.1 | 30 | SFT-Cont | 60.50 [57.64, 62.54] | 61.77 [58.79, 63.94] | 59.85 [57.66, 61.77] | 59.88 [56.40, 62.34] | 55.34 [52.72, 57.59] |
| Llama3.1 | 30 | SFT-Res | 43.98 [38.82, 47.49] | 42.81 [38.34, 47.02] | 47.73 [43.02, 50.94] | 41.42 [35.02, 45.91] | 39.41 [35.73, 42.15] |
| Llama3.1 | 60 | Fed-MedLoRA+ 1R | 54.36 [52.38, 56.62] | 57.88 [56.67, 59.15] | 55.09 [53.18, 57.19] | 50.11 [47.04, 53.60] | 51.15 [48.75, 53.86] |
| Llama3.1 | 60 | Fed-MedLoRA+ 2R | 55.74 [53.27, 57.94] | 58.79 [57.63, 59.96] | 56.25 [53.72, 58.37] | 52.18 [48.21, 55.81] | 52.71 [50.09, 55.30] |
| Llama3.1 | 60 | No-Public 1R | 55.26 [53.26, 57.25] | 58.45 [57.47, 59.55] | 55.77 [53.82, 57.70] | 51.54 [48.16, 54.86] | 52.45 [50.32, 54.77] |
| Llama3.1 | 60 | No-Public 2R | 49.48 [46.25, 52.22] | 53.07 [49.57, 55.71] | 50.66 [48.14, 52.96] | 44.70 [40.80, 48.29] | 46.73 [43.89, 49.24] |
| Llama3.1 | 60 | SFT-Cont | 65.59 [62.89, 67.88] | 66.15 [64.43, 67.63] | 65.49 [62.88, 67.53] | 65.14 [61.12, 68.66] | 63.37 [60.67, 65.57] |
| Llama3.1 | 60 | SFT-Res | 56.70 [52.85, 59.69] | 57.01 [52.38, 60.42] | 57.63 [55.42, 59.32] | 55.46 [50.56, 59.44] | 52.25 [49.99, 54.26] |
| Llama3.1 | 120 | Fed-MedLoRA+ 1R | 65.77 [64.61, 67.11] | 65.87 [64.89, 66.78] | 66.38 [65.38, 67.56] | 65.05 [63.19, 67.08] | 64.47 [63.20, 66.16] |
| Llama3.1 | 120 | Fed-MedLoRA+ 2R | 66.98 [65.85, 68.19] | 66.46 [65.37, 67.30] | 67.36 [66.31, 68.42] | 67.12 [65.42, 68.97] | 65.93 [64.76, 67.21] |
| Llama3.1 | 120 | No-Public 1R | 66.85 [66.13, 67.76] | 66.32 [65.48, 67.06] | 67.28 [66.62, 68.01] | 66.96 [65.77, 68.41] | 65.65 [64.57, 66.91] |
| Llama3.1 | 120 | No-Public 2R | 64.93 [63.28, 66.38] | 65.00 [64.01, 65.85] | 65.57 [64.03, 66.85] | 64.22 [61.63, 66.62] | 64.62 [63.23, 66.09] |
| Llama3.1 | 120 | SFT-Cont | 70.85 [69.20, 72.35] | 69.77 [68.31, 71.05] | 71.40 [70.31, 72.72] | 71.39 [68.86, 73.54] | 70.14 [68.22, 71.91] |
| Llama3.1 | 120 | SFT-Cont 2R | 70.50 [69.74, 71.46] | 69.35 [68.43, 70.24] | 70.66 [69.99, 71.66] | 71.48 [70.17, 72.70] | 69.94 [69.04, 71.00] |
| Llama3.1 | 120 | SFT-Res | 68.01 [65.86, 69.42] | 67.38 [64.70, 69.29] | 68.14 [66.66, 69.27] | 68.51 [66.04, 70.69] | 66.95 [64.36, 68.94] |
| Llama3.1 | 240 | Fed-MedLoRA+ 1R | 71.44 [70.33, 72.28] | 70.05 [69.55, 70.54] | 72.06 [71.13, 72.87] | 72.22 [70.05, 73.61] | 70.43 [69.07, 71.59] |
| Llama3.1 | 240 | Fed-MedLoRA+ 2R | 73.23 [72.68, 73.72] | 71.26 [70.90, 71.65] | 73.37 [72.78, 73.93] | 75.06 [73.87, 75.88] | 72.32 [71.45, 73.17] |
| Llama3.1 | 240 | No-Public 1R | 72.47 [71.42, 73.22] | 70.78 [70.24, 71.35] | 72.93 [72.04, 73.64] | 73.69 [71.66, 74.95] | 71.61 [70.49, 72.62] |
| Llama3.1 | 240 | No-Public 2R | 72.37 [71.88, 72.85] | 70.42 [70.00, 70.82] | 72.41 [71.94, 72.87] | 74.28 [73.20, 75.07] | 71.58 [70.85, 72.29] |
| Llama3.1 | 240 | SFT-Cont | 74.90 [73.95, 75.65] | 72.91 [72.11, 73.59] | 75.44 [74.54, 76.27] | 76.37 [74.78, 77.60] | 74.21 [73.15, 75.06] |

| Model | Steps/client | Method | Source | MIMIC-III | MTSamples | UTP | i2b2 |
|---|---|---|---|---|---|---|---|
| Llama3.1 | 240 | SFT-Res | 72.98 [72.17, 73.59] | 71.71 [70.86, 72.47] | 73.48 [72.70, 74.15] | 73.75 [72.50, 74.61] | 72.42 [71.56, 73.19] |
| Llama3.1 | 480 | Fed-MedLoRA+ 1R | 75.06 [74.15, 75.88] | 73.02 [72.40, 73.67] | 75.94 [75.27, 76.46] | 76.21 [74.47, 77.79] | 74.48 [73.63, 75.29] |
| Llama3.1 | 480 | No-Public 1R | 76.27 [75.57, 77.01] | 74.14 [73.53, 74.78] | 77.42 [76.93, 77.86] | 77.25 [75.89, 78.65] | 75.81 [75.16, 76.37] |
| Llama3.1 | 480 | SFT-Cont | 77.39 [76.45, 78.18] | 74.96 [74.27, 75.56] | 78.72 [78.05, 79.30] | 78.48 [76.51, 80.21] | 76.98 [76.26, 77.51] |
| Llama3.1 | 480 | SFT-Res | 76.40 [75.64, 77.09] | 74.76 [74.15, 75.21] | 77.53 [76.69, 78.28] | 76.90 [75.43, 78.09] | 76.13 [75.49, 76.69] |

Values are exact-match F1 [95% CI]. Rounds are included in method labels.

## Supplementary Table S3. Paired NER preparation, adaptation, and baseline comparisons

| Model | Steps/client | Contrast | Source | MIMIC-III | MTSamples | UTP | i2b2 |
|---|---|---|---|---|---|---|---|
| Llama3.1 | 30 | SFT-Cont − Fed-MedLoRA+ 1R | +19.07 [15.73, 24.39] | +16.37 [11.95, 21.90] | +16.24 [13.59, 20.38] | +24.59 [20.64, 31.06] | +16.31 [13.59, 20.59] |
| Llama3.1 | 30 | SFT-Cont − No-Public 1R | +18.68 [15.29, 24.19] | +15.97 [11.47, 22.16] | +15.87 [13.24, 19.77] | +24.19 [20.05, 30.91] | +15.85 [12.76, 20.34] |
| Llama3.1 | 30 | SFT-Cont − SFT-Res | +16.52 [14.14, 19.35] | +18.97 [15.85, 22.29] | +12.12 [9.84, 14.76] | +18.46 [15.84, 21.89] | +15.93 [14.30, 17.71] |
| Llama3.1 | 30 | SFT-Res − No-Public 1R | +2.16 [−2.92, 8.43] | −2.99 [−8.11, 2.48] | +3.74 [−0.14, 8.23] | +5.73 [−0.87, 14.22] | −0.08 [−3.41, 4.45] |
| Llama3.1 | 60 | SFT-Cont − Fed-MedLoRA+ 1R | +11.24 [8.60, 14.12] | +8.27 [6.60, 10.19] | +10.41 [7.74, 13.00] | +15.03 [10.94, 19.38] | +12.22 [9.50, 14.94] |
| Llama3.1 | 60 | SFT-Cont − No-Public 1R | +10.34 [8.08, 12.54] | +7.69 [6.07, 9.60] | +9.72 [7.47, 11.68] | +13.60 [10.11, 16.95] | +10.93 [8.63, 13.06] |
| Llama3.1 | 60 | SFT-Cont − SFT-Res | +8.89 [5.78, 11.50] | +9.13 [6.28, 12.59] | +7.86 [4.96, 9.83] | +9.68 [5.80, 13.00] | +11.12 [7.99, 13.44] |
| Llama3.1 | 60 | SFT-Res − No-Public 1R | +1.45 [−2.12, 4.01] | −1.44 [−5.88, 1.91] | +1.85 [−0.43, 3.54] | +3.92 [−0.24, 7.26] | −0.19 [−2.84, 1.78] |
| Llama3.1 | 120 | SFT-Cont − Fed-MedLoRA+ 1R | +5.09 [4.11, 6.33] | +3.90 [3.06, 4.73] | +5.02 [4.28, 5.77] | +6.34 [4.58, 8.57] | +5.67 [4.62, 6.77] |
| Llama3.1 | 120 | SFT-Cont − No-Public 1R | +4.00 [2.92, 5.07] | +3.46 [2.63, 4.21] | +4.12 [3.31, 5.03] | +4.43 [2.13, 6.37] | +4.49 [3.39, 5.53] |
| Llama3.1 | 120 | SFT-Cont − SFT-Res | +2.84 [1.86, 3.77] | +2.40 [1.14, 3.83] | +3.26 [2.44, 3.97] | +2.88 [1.16, 4.66] | +3.19 [1.77, 4.36] |
| Llama3.1 | 120 | SFT-Res − No-Public 1R | +1.16 [−0.42, 2.45] | +1.06 [−0.96, 2.49] | +0.86 [−0.15, 1.75] | +1.56 [−0.56, 3.87] | +1.30 [−0.44, 2.67] |
| Llama3.1 | 240 | SFT-Cont − Fed-MedLoRA+ 1R | +3.46 [2.75, 3.92] | +2.86 [2.26, 3.52] | +3.38 [2.71, 4.02] | +4.14 [2.97, 5.14] | +3.78 [3.02, 4.64] |
| Llama3.1 | 240 | SFT-Cont − No-Public 1R | +2.44 [1.65, 2.91] | +2.13 [1.33, 2.86] | +2.51 [1.81, 3.10] | +2.68 [1.60, 3.66] | +2.60 [1.82, 3.41] |
| Llama3.1 | 240 | SFT-Cont − SFT-Res | +1.93 [1.56, 2.24] | +1.20 [0.79, 1.55] | +1.96 [1.52, 2.42] | +2.62 [1.92, 3.27] | +1.79 [1.09, 2.47] |
| Llama3.1 | 240 | SFT-Res − No-Public 1R | +0.51 [−0.08, 1.09] | +0.93 [0.10, 1.74] | +0.55 [0.01, 0.98] | +0.06 [−0.70, 1.41] | +0.81 [0.09, 1.37] |
| Llama3.1 | 480 | SFT-Cont − Fed-MedLoRA+ 1R | +2.33 [1.94, 2.70] | +1.94 [1.61, 2.20] | +2.78 [2.51, 3.10] | +2.27 [1.48, 3.07] | +2.50 [1.75, 3.07] |
| Llama3.1 | 480 | SFT-Cont − No-Public 1R | +1.12 [0.60, 1.61] | +0.81 [0.31, 1.22] | +1.30 [0.88, 1.66] | +1.23 [0.43, 2.38] | +1.17 [0.61, 1.71] |
| Llama3.1 | 480 | SFT-Cont − SFT-Res | +0.99 [0.49, 1.55] | +0.20 [−0.46, 0.75] | +1.20 [0.56, 1.92] | +1.58 [0.90, 2.40] | +0.85 [0.31, 1.40] |
| Llama3.1 | 480 | SFT-Res − No-Public 1R | +0.12 [−0.12, 0.37] | +0.61 [0.25, 1.11] | +0.11 [−0.33, 0.57] | −0.34 [−0.84, 0.15] | +0.32 [−0.12, 0.68] |
| Qwen3.5 | 30 | Adaptation main effect: Cont − Res | +12.66 [10.58, 14.62] | +10.63 [9.14, 12.75] | +11.80 [10.13, 13.35] | +15.56 [11.98, 20.07] | +15.08 [12.60, 16.93] |
| Qwen3.5 | 30 | Meta-Cont − Meta-Res | +12.07 [7.23, 17.82] | +9.93 [7.48, 12.28] | +11.24 [7.38, 15.89] | +15.03 [6.40, 25.97] | +14.96 [10.71, 18.69] |
| Qwen3.5 | 30 | Meta-Cont − No-Public 1R | +14.28 [10.09, 19.01] | +9.41 [6.83, 13.19] | +11.24 [8.24, 14.81] | +22.20 [14.86, 29.38] | +14.15 [11.04, 17.46] |

| Model | Steps/client | Contrast | Source | MIMIC-III | MTSamples | UTP | i2b2 |
|---|---|---|---|---|---|---|---|
| Qwen3.5 | 30 | Meta-Cont − SFT-Cont | −0.52 [−2.80, 1.66] | −0.09 [−1.04, 0.70] | −0.81 [−2.21, 0.98] | −0.67 [−5.58, 3.58] | −0.65 [−1.98, 0.75] |
| Qwen3.5 | 30 | Meta-Res − No-Public 1R | +2.22 [−4.52, 8.64] | −0.52 [−3.31, 2.55] | +0.00 [−4.99, 4.95] | +7.17 [−5.56, 18.44] | −0.81 [−4.28, 2.49] |
| Qwen3.5 | 30 | Meta-Res − SFT-Res | +0.67 [−8.29, 8.76] | +1.31 [−3.48, 6.40] | +0.31 [−6.90, 6.91] | +0.38 [−14.10, 13.10] | −0.43 [−5.45, 4.41] |
| Qwen3.5 | 30 | Preparation main effect: Meta − SFT | +0.07 [−4.61, 4.87] | +0.61 [−1.97, 3.33] | −0.25 [−4.05, 3.84] | −0.14 [−7.64, 7.48] | −0.54 [−3.29, 2.40] |
| Qwen3.5 | 30 | Preparation × adaptation | −1.19 [−8.06, 6.83] | −1.40 [−6.16, 3.38] | −1.12 [−6.64, 5.37] | −1.05 [−11.74, 12.05] | −0.22 [−4.40, 4.89] |
| Qwen3.5 | 30 | SFT-Cont − Fed-MedLoRA+ 1R | +15.75 [12.74, 19.83] | +10.22 [8.39, 13.75] | +12.85 [9.93, 16.51] | +24.18 [19.30, 29.68] | +15.48 [12.67, 18.91] |
| Qwen3.5 | 30 | SFT-Cont − No-Public 1R | +14.81 [11.40, 19.20] | +9.51 [7.39, 13.25] | +12.05 [8.92, 16.13] | +22.86 [17.74, 28.60] | +14.80 [11.68, 18.38] |
| Qwen3.5 | 30 | SFT-Cont − SFT-Res | +13.26 [10.76, 16.07] | +11.34 [8.37, 15.10] | +12.36 [10.09, 14.87] | +16.08 [13.46, 18.68] | +15.19 [13.24, 17.80] |
| Qwen3.5 | 30 | SFT-Res − No-Public 1R | +1.55 [−3.37, 5.00] | −1.83 [−5.72, 0.58] | −0.31 [−4.56, 2.74] | +6.78 [−0.32, 12.73] | −0.39 [−2.92, 1.90] |
| Qwen3.5 | 60 | Meta-Cont − Meta-Res | +6.55 [4.68, 8.69] | +5.43 [3.91, 6.95] | +7.28 [5.27, 9.29] | +6.93 [3.83, 10.12] | +8.81 [6.70, 10.99] |
| Qwen3.5 | 60 | Meta-Cont − No-Public 1R | +7.30 [6.30, 8.31] | +5.43 [4.43, 6.46] | +7.77 [6.75, 8.92] | +8.70 [6.69, 10.15] | +9.03 [7.87, 10.14] |
| Qwen3.5 | 60 | Meta-Cont − SFT-Cont | −0.52 [−1.25, 0.39] | −0.30 [−1.03, 0.29] | −0.03 [−0.47, 0.47] | −1.22 [−2.89, 0.67] | −0.58 [−1.14, −0.01] |
| Qwen3.5 | 60 | Meta-Res − No-Public 1R | +0.75 [−0.86, 1.93] | −0.01 [−0.66, 0.60] | +0.49 [−1.63, 2.39] | +1.77 [−0.68, 3.38] | +0.22 [−1.21, 1.87] |
| Qwen3.5 | 60 | Meta-Res − SFT-Res | −1.02 [−3.83, 0.70] | −0.84 [−2.35, 0.08] | −0.72 [−3.31, 1.66] | −1.50 [−5.97, 0.97] | −1.54 [−3.49, 0.44] |
| Qwen3.5 | 60 | Preparation × adaptation | +0.50 [−1.74, 3.49] | +0.54 [−0.91, 2.29] | +0.69 [−1.95, 3.26] | +0.28 [−3.06, 5.27] | +0.95 [−1.21, 3.03] |
| Qwen3.5 | 60 | SFT-Cont − Fed-MedLoRA+ 1R | +8.57 [7.31, 9.80] | +6.12 [5.25, 6.93] | +8.64 [7.55, 9.74] | +10.95 [8.93, 12.93] | +10.52 [8.93, 12.04] |
| Qwen3.5 | 60 | SFT-Cont − No-Public 1R | +7.82 [6.65, 9.04] | +5.73 [4.89, 6.52] | +7.80 [6.80, 9.04] | +9.93 [8.07, 11.86] | +9.61 [8.07, 11.18] |
| Qwen3.5 | 60 | SFT-Cont − SFT-Res | +6.05 [4.26, 7.33] | +4.89 [3.88, 5.94] | +6.59 [5.20, 7.54] | +6.65 [3.83, 9.18] | +7.86 [6.51, 9.19] |
| Qwen3.5 | 60 | SFT-Res − No-Public 1R | +1.77 [0.73, 3.09] | +0.84 [0.03, 2.04] | +1.21 [0.41, 2.12] | +3.27 [1.39, 5.54] | +1.76 [1.06, 2.64] |
| Qwen3.5 | 120 | Meta-Cont − Meta-Res | +3.44 [2.66, 4.26] | +2.83 [2.33, 3.35] | +3.66 [2.77, 4.41] | +3.82 [2.35, 5.35] | +3.99 [2.97, 5.14] |
| Qwen3.5 | 120 | Meta-Cont − No-Public 1R | +3.00 [2.42, 3.86] | +2.48 [1.80, 3.29] | +2.64 [2.16, 3.08] | +3.87 [2.43, 5.53] | +3.50 [3.06, 3.96] |
| Qwen3.5 | 120 | Meta-Cont − SFT-Cont | +0.00 [−0.92, 0.68] | −0.15 [−0.61, 0.19] | +0.29 [−0.09, 0.78] | −0.16 [−2.10, 1.34] | +0.37 [−0.38, 0.98] |
| Qwen3.5 | 120 | Meta-Res − No-Public 1R | −0.44 [−1.24, 0.24] | −0.35 [−0.98, 0.28] | −1.03 [−1.82, −0.06] | +0.04 [−1.97, 1.80] | −0.49 [−1.37, 0.33] |
| Qwen3.5 | 120 | Meta-Res − SFT-Res | −0.16 [−0.57, 0.30] | −0.57 [−1.24, −0.06] | +0.44 [−0.60, 1.42] | −0.33 [−1.32, 0.54] | −0.40 [−1.30, 0.45] |
| Qwen3.5 | 120 | Preparation × adaptation | +0.15 [−0.96, 1.07] | +0.43 [−0.11, 1.07] | −0.15 [−1.45, 1.08] | +0.18 [−1.96, 2.06] | +0.77 [−0.48, 2.00] |
| Qwen3.5 | 120 | SFT-Cont − Fed-MedLoRA+ 1R | +3.72 [2.62, 4.80] | +3.06 [2.42, 3.71] | +3.08 [2.34, 3.78] | +5.02 [2.40, 7.42] | +3.78 [3.06, 4.61] |
| Qwen3.5 | 120 | SFT-Cont − No-Public 1R | +3.00 [1.96, 4.18] | +2.63 [1.98, 3.44] | +2.35 [1.73, 2.94] | +4.02 [1.54, 6.71] | +3.13 [2.40, 4.06] |
| Qwen3.5 | 120 | SFT-Cont − SFT-Res | +3.29 [2.49, 4.27] | +2.41 [1.72, 3.23] | +3.81 [2.86, 4.77] | +3.65 [2.31, 4.93] | +3.22 [2.14, 4.47] |
| Qwen3.5 | 120 | SFT-Res − No-Public 1R | −0.29 [−1.10, 0.50] | +0.22 [−0.05, 0.58] | −1.46 [−2.01, −0.91] | +0.38 [−2.03, 2.71] | −0.08 [−1.06, 0.88] |
| Qwen3.5 | 240 | Meta-Cont − Meta-Res | +1.50 [0.72, 2.29] | +1.16 [0.83, 1.53] | +1.47 [0.98, 2.04] | +1.86 [0.09, 4.01] | +1.99 [1.30, 2.73] |
| Qwen3.5 | 240 | Meta-Cont − No-Public 1R | +1.45 [0.73, 2.25] | +1.25 [0.93, 1.64] | +1.31 [0.90, 1.79] | +1.78 [0.20, 3.66] | +2.02 [1.49, 2.78] |
| Qwen3.5 | 240 | Meta-Cont − SFT-Cont | +0.20 [−0.33, 0.72] | −0.04 [−0.46, 0.33] | +0.11 [−0.32, 0.58] | +0.54 [−0.52, 1.62] | +0.41 [−0.14, 1.10] |

| Model | Steps/client | Contrast | Source | MIMIC-III | MTSamples | UTP | i2b2 |
|---|---|---|---|---|---|---|---|
| Qwen3.5 | 240 | Meta-Res − No-Public 1R | −0.05 [−0.34, 0.26] | +0.09 [−0.18, 0.35] | −0.17 [−0.56, 0.19] | −0.07 [−0.55, 0.40] | +0.03 [−0.55, 0.32] |
| Qwen3.5 | 240 | Meta-Res − SFT-Res | +0.02 [−0.43, 0.54] | −0.12 [−0.50, 0.27] | +0.38 [−0.24, 0.90] | −0.20 [−1.34, 0.75] | −0.02 [−0.74, 0.64] |
| Qwen3.5 | 240 | Preparation × adaptation | +0.19 [−0.51, 0.93] | +0.08 [−0.50, 0.55] | −0.26 [−1.02, 0.55] | +0.75 [−0.70, 2.77] | +0.43 [−0.34, 1.26] |
| Qwen3.5 | 240 | SFT-Cont − Fed-MedLoRA+ 1R | +1.97 [1.33, 2.80] | +1.71 [1.38, 2.11] | +1.78 [1.43, 2.13] | +2.43 [1.05, 4.29] | +2.35 [2.00, 2.77] |
| Qwen3.5 | 240 | SFT-Cont − No-Public 1R | +1.24 [0.77, 1.78] | +1.29 [1.00, 1.59] | +1.20 [0.82, 1.61] | +1.24 [0.24, 2.49] | +1.62 [1.31, 1.93] |
| Qwen3.5 | 240 | SFT-Cont − SFT-Res | +1.31 [0.80, 1.99] | +1.08 [0.80, 1.42] | +1.74 [1.08, 2.43] | +1.11 [0.24, 2.37] | +1.56 [0.66, 2.22] |
| Qwen3.5 | 240 | SFT-Res − No-Public 1R | −0.07 [−0.55, 0.42] | +0.21 [−0.14, 0.64] | −0.54 [−0.97, −0.03] | +0.13 [−0.74, 1.00] | +0.05 [−0.44, 0.73] |
| Qwen3.5 | 480 | Meta-Cont − Meta-Res | +0.30 [−0.10, 0.67] | +0.23 [0.00, 0.45] | +0.35 [0.01, 0.69] | +0.32 [−0.70, 1.15] | +0.45 [0.15, 0.81] |
| Qwen3.5 | 480 | Meta-Cont − No-Public 1R | +0.80 [0.41, 1.09] | +0.70 [0.46, 0.95] | +0.83 [0.67, 1.01] | +0.86 [−0.12, 1.63] | +1.10 [0.67, 1.48] |
| Qwen3.5 | 480 | Meta-Cont − SFT-Cont | +0.00 [−0.27, 0.28] | −0.17 [−0.39, 0.00] | +0.13 [−0.02, 0.28] | +0.02 [−0.62, 0.77] | +0.02 [−0.34, 0.39] |
| Qwen3.5 | 480 | Meta-Res − No-Public 1R | +0.50 [0.24, 0.79] | +0.47 [0.19, 0.74] | +0.48 [0.25, 0.71] | +0.54 [0.01, 1.16] | +0.65 [0.34, 0.94] |
| Qwen3.5 | 480 | Meta-Res − SFT-Res | +0.50 [0.12, 0.83] | −0.01 [−0.31, 0.32] | +0.69 [0.39, 0.94] | +0.81 [0.03, 1.69] | +0.52 [0.00, 1.04] |
| Qwen3.5 | 480 | Preparation × adaptation | −0.50 [−0.84, −0.14] | −0.16 [−0.44, 0.14] | −0.55 [−0.84, −0.21] | −0.79 [−1.85, −0.10] | −0.51 [−1.24, 0.26] |
| Qwen3.5 | 480 | SFT-Cont − Fed-MedLoRA+ 1R | +1.54 [1.18, 1.85] | +1.31 [0.98, 1.68] | +1.47 [1.20, 1.77] | +1.84 [1.03, 2.56] | +1.78 [1.26, 2.25] |
| Qwen3.5 | 480 | SFT-Cont − No-Public 1R | +0.80 [0.31, 1.19] | +0.86 [0.58, 1.18] | +0.70 [0.47, 0.92] | +0.84 [−0.25, 1.70] | +1.09 [0.45, 1.58] |
| Qwen3.5 | 480 | SFT-Cont − SFT-Res | +0.80 [0.39, 1.17] | +0.39 [0.15, 0.62] | +0.90 [0.57, 1.18] | +1.11 [0.10, 1.92] | +0.96 [0.18, 1.61] |
| Qwen3.5 | 480 | SFT-Res − No-Public 1R | +0.00 [−0.38, 0.28] | +0.48 [0.27, 0.63] | −0.20 [−0.50, 0.12] | −0.27 [−1.29, 0.42] | +0.13 [−0.13, 0.46] |
| Qwen3.5 | 30 | Fed-MedLoRA+ 1R − No-Public 1R | −0.94 [−1.64, −0.39] | −0.71 [−1.59, −0.17] | −0.80 [−1.36, −0.26] | −1.32 [−2.08, −0.57] | −0.68 [−1.21, −0.37] |
| Qwen3.5 | 30 | Meta-Cont − Fed-MedLoRA+ 1R | +15.23 [10.98, 19.64] | +10.12 [7.46, 13.69] | +12.04 [9.02, 15.35] | +23.52 [16.25, 30.51] | +14.83 [11.87, 17.95] |
| Qwen3.5 | 60 | Fed-MedLoRA+ 1R − No-Public 1R | −0.75 [−1.07, −0.51] | −0.40 [−0.50, −0.31] | −0.84 [−1.27, −0.44] | −1.03 [−1.50, −0.67] | −0.91 [−1.27, −0.60] |
| Qwen3.5 | 60 | Meta-Cont − Fed-MedLoRA+ 1R | +8.05 [6.92, 9.10] | +5.82 [4.85, 6.84] | +8.61 [7.41, 9.82] | +9.73 [7.52, 11.31] | +9.93 [8.68, 11.04] |
| Qwen3.5 | 120 | Fed-MedLoRA+ 1R − No-Public 1R | −0.72 [−1.08, −0.45] | −0.43 [−0.64, −0.25] | −0.73 [−1.04, −0.42] | −1.00 [−1.66, −0.47] | −0.65 [−0.96, −0.37] |
| Qwen3.5 | 120 | Meta-Cont − Fed-MedLoRA+ 1R | +3.72 [2.96, 4.58] | +2.92 [2.23, 3.64] | +3.37 [2.72, 3.90] | +4.87 [3.14, 6.55] | +4.16 [3.74, 4.53] |
| Qwen3.5 | 240 | Fed-MedLoRA+ 1R − No-Public 1R | −0.73 [−1.14, −0.51] | −0.42 [−0.66, −0.22] | −0.58 [−0.78, −0.41] | −1.18 [−2.07, −0.52] | −0.73 [−1.06, −0.49] |
| Qwen3.5 | 240 | Meta-Cont − Fed-MedLoRA+ 1R | +2.18 [1.35, 3.35] | +1.68 [1.29, 2.26] | +1.89 [1.42, 2.51] | +2.97 [1.05, 5.51] | +2.75 [2.15, 3.81] |
| Qwen3.5 | 480 | Fed-MedLoRA+ 1R − No-Public 1R | −0.74 [−1.04, −0.48] | −0.45 [−0.69, −0.26] | −0.77 [−1.00, −0.59] | −1.00 [−1.70, −0.44] | −0.69 [−0.97, −0.42] |
| Qwen3.5 | 480 | Meta-Cont − Fed-MedLoRA+ 1R | +1.54 [1.10, 1.85] | +1.15 [0.91, 1.50] | +1.60 [1.40, 1.79] | +1.86 [0.77, 2.68] | +1.80 [1.39, 2.17] |

Values are F1-point differences [paired 95% CI]; positive differences favor the first method. Preparation × adaptation is (Meta-Cont − Meta-Res) − (SFT-Cont − SFT-Res).

## Supplementary Table S4. RE performance across training budgets

| Model | Steps/client | Method | Source | MIMIC-III | MTSamples | UTP | i2b2 |
|---|---|---|---|---|---|---|---|
| Qwen3.5 | 30 | No-Public 1R | 39.94 [30.56, 48.19] | 33.00 [21.77, 43.12] | 42.94 [35.79, 48.98] | 43.89 [34.12, 52.58] | 33.94 [23.28, 43.38] |
| Qwen3.5 | 30 | SFT-Res | 41.79 [31.36, 50.65] | 38.03 [27.64, 46.38] | 44.32 [35.46, 52.24] | 43.01 [31.27, 53.53] | 39.19 [26.08, 49.17] |
| Qwen3.5 | 30 | SFT-Cont | 50.55 [41.65, 55.81] | 44.89 [33.84, 51.57] | 53.45 [46.19, 57.90] | 53.31 [44.32, 58.45] | 50.44 [37.15, 57.81] |
| Qwen3.5 | 30 | Meta-Res | 41.85 [36.28, 46.40] | 36.68 [28.78, 42.36] | 45.05 [41.02, 48.09] | 43.81 [38.32, 49.28] | 37.29 [30.25, 44.23] |
| Qwen3.5 | 30 | Meta-Cont | 52.40 [47.23, 56.20] | 47.45 [40.51, 52.06] | 54.56 [50.68, 57.53] | 55.18 [49.75, 59.59] | 50.63 [43.35, 56.19] |
| Qwen3.5 | 30 | Fed-MedLoRA+ 1R | 40.69 [32.01, 48.44] | 33.81 [23.27, 43.30] | 43.51 [37.04, 49.09] | 44.76 [35.30, 52.93] | 34.81 [24.74, 43.68] |
| Qwen3.5 | 60 | No-Public 1R | 54.41 [50.23, 56.93] | 51.15 [45.62, 54.35] | 55.57 [52.36, 57.59] | 56.53 [52.28, 59.39] | 53.97 [48.38, 57.19] |
| Qwen3.5 | 60 | SFT-Res | 53.72 [50.82, 55.79] | 50.25 [46.68, 52.71] | 55.86 [53.32, 57.53] | 55.05 [52.41, 57.40] | 54.27 [50.43, 56.95] |
| Qwen3.5 | 60 | SFT-Cont | 58.25 [56.87, 59.29] | 53.64 [51.83, 55.18] | 59.47 [58.64, 60.32] | 61.63 [59.78, 63.09] | 58.41 [55.86, 59.67] |
| Qwen3.5 | 60 | Meta-Res | 53.30 [51.24, 55.51] | 48.11 [45.77, 50.37] | 55.48 [53.67, 57.09] | 56.31 [52.27, 59.80] | 52.25 [48.39, 55.94] |
| Qwen3.5 | 60 | Meta-Cont | 58.40 [57.33, 59.49] | 53.63 [51.31, 55.93] | 59.37 [58.66, 60.20] | 62.21 [60.57, 63.88] | 58.08 [56.14, 59.70] |
| Qwen3.5 | 60 | Fed-MedLoRA+ 1R | 54.15 [50.81, 56.39] | 50.35 [46.41, 52.92] | 55.70 [53.20, 57.42] | 56.40 [52.58, 58.98] | 54.53 [50.56, 56.96] |
| Qwen3.5 | 60 | No-Public 2R | 51.88 [49.63, 53.87] | 48.35 [45.79, 50.65] | 53.44 [51.32, 55.24] | 53.85 [51.50, 56.02] | 50.17 [46.61, 53.12] |
| Qwen3.5 | 120 | No-Public 1R | 61.36 [60.43, 62.45] | 59.14 [57.61, 61.00] | 60.89 [59.77, 61.80] | 64.05 [62.68, 65.29] | 61.57 [60.69, 62.27] |
| Qwen3.5 | 120 | SFT-Res | 60.76 [59.76, 61.65] | 56.69 [55.39, 57.91] | 60.91 [60.09, 61.60] | 64.68 [62.81, 66.31] | 60.57 [59.32, 61.33] |
| Qwen3.5 | 120 | SFT-Cont | 63.05 [62.47, 63.67] | 58.50 [57.07, 59.92] | 62.85 [62.37, 63.32] | 67.81 [66.62, 68.87] | 62.85 [61.98, 63.45] |
| Qwen3.5 | 120 | Meta-Res | 61.10 [60.40, 61.83] | 56.44 [54.45, 58.07] | 61.47 [60.66, 62.32] | 65.40 [64.49, 66.29] | 60.02 [58.54, 61.52] |
| Qwen3.5 | 120 | Meta-Cont | 62.91 [61.53, 64.22] | 58.53 [56.15, 61.00] | 62.46 [61.48, 63.27] | 67.74 [66.30, 69.18] | 61.12 [58.24, 63.15] |
| Qwen3.5 | 120 | Fed-MedLoRA+ 1R | 61.19 [60.46, 62.00] | 58.75 [57.74, 60.12] | 60.96 [59.92, 61.68] | 63.85 [62.71, 64.84] | 61.68 [60.64, 62.29] |
| Qwen3.5 | 120 | No-Public 2R | 61.71 [60.69, 62.63] | 62.72 [60.91, 64.49] | 59.31 [58.48, 60.23] | 63.10 [61.39, 64.58] | 59.11 [57.65, 60.42] |
| Qwen3.5 | 240 | No-Public 1R | 66.30 [65.00, 67.56] | 64.03 [62.62, 65.39] | 64.12 [63.07, 65.39] | 70.75 [68.71, 72.62] | 64.96 [64.00, 65.99] |
| Qwen3.5 | 240 | SFT-Res | 64.81 [63.90, 65.71] | 60.74 [59.32, 62.08] | 63.43 [62.47, 64.53] | 70.27 [69.38, 71.26] | 63.25 [62.16, 64.25] |
| Qwen3.5 | 240 | SFT-Cont | 66.72 [65.91, 67.42] | 62.50 [60.79, 63.91] | 65.78 [65.16, 66.70] | 71.87 [71.13, 72.42] | 64.97 [64.39, 65.74] |
| Qwen3.5 | 240 | Meta-Res | 64.64 [63.89, 65.36] | 60.32 [59.07, 61.40] | 63.58 [62.92, 64.40] | 70.02 [69.17, 70.79] | 63.04 [61.75, 64.02] |
| Qwen3.5 | 240 | Meta-Cont | 67.25 [66.23, 68.16] | 63.45 [61.35, 64.99] | 66.09 [65.11, 66.96] | 72.20 [71.34, 73.19] | 65.16 [63.95, 66.28] |
| Qwen3.5 | 240 | Fed-MedLoRA+ 1R | 65.44 [64.36, 66.58] | 62.71 [61.28, 63.97] | 64.24 [63.18, 65.35] | 69.38 [67.76, 70.71] | 64.73 [63.89, 65.68] |
| Qwen3.5 | 240 | No-Public 2R | 69.98 [68.88, 71.05] | 70.41 [69.50, 71.33] | 64.96 [63.42, 66.30] | 74.58 [73.49, 75.95] | 65.15 [64.50, 65.94] |
| Qwen3.5 | 480 | No-Public 1R | 73.33 [71.68, 74.73] | 71.65 [69.55, 73.40] | 70.23 [69.21, 70.98] | 78.12 [75.18, 80.75] | 69.00 [68.27, 69.82] |
| Qwen3.5 | 480 | SFT-Res | 72.21 [70.66, 73.28] | 68.24 [66.31, 70.47] | 70.29 [69.31, 71.18] | 78.12 [75.79, 80.23] | 68.20 [67.04, 69.49] |
| Qwen3.5 | 480 | SFT-Cont | 73.00 [71.18, 74.45] | 69.45 [66.72, 72.14] | 71.36 [70.50, 72.17] | 78.20 [75.88, 80.39] | 69.21 [67.49, 70.88] |
| Qwen3.5 | 480 | Meta-Res | 71.80 [70.53, 72.99] | 67.98 [66.04, 70.03] | 70.22 [69.28, 71.04] | 77.20 [75.34, 78.81] | 67.33 [66.14, 68.56] |

| Model | Steps/client | Method | Source | MIMIC-III | MTSamples | UTP | i2b2 |
|---|---|---|---|---|---|---|---|
| Qwen3.5 | 480 | Meta-Cont | 73.26 [71.90, 74.46] | 70.19 [68.17, 72.14] | 71.71 [71.09, 72.41] | 77.88 [76.06, 79.55] | 68.54 [67.66, 69.52] |
| Qwen3.5 | 480 | Fed-MedLoRA+ 1R | 71.41 [69.45, 72.99] | 69.34 [67.28, 71.40] | 69.62 [68.64, 70.49] | 75.26 [71.87, 78.13] | 68.43 [67.56, 69.10] |
| Qwen3.5 | 480 | No-Public 2R | 76.56 [75.80, 77.21] | 75.42 [74.35, 76.35] | 71.25 [70.70, 71.79] | 83.00 [82.00, 83.81] | 71.12 [70.56, 71.76] |
| Llama3.1 | 30 | No-Public 1R | 6.67 [2.95, 13.62] | 5.43 [2.14, 11.82] | 8.08 [4.04, 14.33] | 6.49 [2.39, 14.58] | 3.06 [0.77, 7.64] |
| Llama3.1 | 30 | SFT-Res | 6.89 [1.88, 11.89] | 6.93 [1.24, 12.31] | 8.21 [2.14, 14.16] | 5.54 [1.88, 10.51] | 3.66 [0.52, 6.83] |
| Llama3.1 | 30 | SFT-Cont | 15.59 [7.88, 23.04] | 14.31 [6.97, 21.10] | 17.21 [9.09, 24.62] | 15.25 [6.97, 24.45] | 11.08 [5.44, 17.59] |
| Llama3.1 | 60 | No-Public 1R | 32.28 [25.61, 38.50] | 25.20 [17.44, 32.95] | 33.01 [26.95, 38.35] | 38.64 [32.67, 44.28] | 22.03 [14.45, 30.42] |
| Llama3.1 | 60 | SFT-Res | 35.87 [25.32, 44.13] | 33.85 [23.70, 41.52] | 36.02 [25.31, 44.16] | 37.76 [26.47, 46.72] | 31.29 [17.24, 41.53] |
| Llama3.1 | 60 | SFT-Cont | 42.94 [33.25, 50.21] | 39.05 [27.32, 47.67] | 43.15 [34.03, 50.04] | 46.62 [38.44, 53.47] | 39.56 [26.16, 49.22] |
| Llama3.1 | 120 | No-Public 1R | 51.55 [48.64, 54.09] | 47.92 [43.70, 51.50] | 51.56 [48.93, 53.93] | 55.17 [52.86, 57.09] | 49.10 [44.99, 52.71] |
| Llama3.1 | 120 | SFT-Res | 53.62 [52.33, 55.02] | 50.81 [48.64, 52.70] | 53.77 [52.32, 55.22] | 56.28 [54.99, 57.57] | 53.28 [51.35, 55.09] |
| Llama3.1 | 120 | SFT-Cont | 57.11 [55.66, 57.84] | 54.71 [52.97, 55.55] | 57.03 [56.08, 57.84] | 59.60 [57.58, 60.57] | 57.43 [56.11, 58.47] |
| Llama3.1 | 240 | No-Public 1R | 57.64 [55.41, 59.32] | 54.15 [51.77, 55.79] | 57.98 [55.95, 59.67] | 60.80 [58.35, 62.80] | 57.61 [55.82, 58.85] |
| Llama3.1 | 240 | SFT-Res | 58.19 [54.42, 60.37] | 55.40 [51.42, 57.98] | 58.31 [54.37, 60.43] | 60.85 [57.42, 63.26] | 58.42 [55.26, 60.14] |
| Llama3.1 | 240 | SFT-Cont | 59.15 [56.81, 60.94] | 55.63 [53.67, 57.34] | 59.55 [56.90, 61.68] | 62.28 [59.61, 64.55] | 59.64 [57.56, 61.24] |
| Llama3.1 | 480 | No-Public 1R | 65.69 [64.64, 66.89] | 60.36 [59.38, 61.48] | 67.13 [66.10, 68.09] | 69.60 [68.02, 71.45] | 63.96 [62.85, 65.14] |
| Llama3.1 | 480 | SFT-Res | 65.70 [64.34, 67.46] | 60.89 [59.39, 63.22] | 66.69 [65.82, 67.70] | 69.53 [67.36, 71.80] | 63.91 [62.97, 64.91] |
| Llama3.1 | 480 | SFT-Cont | 67.36 [66.58, 68.23] | 61.62 [60.45, 63.25] | 68.39 [67.74, 69.17] | 72.08 [71.16, 73.11] | 65.45 [64.73, 66.20] |

Values are exact-match F1 [95% CI]. Rounds are included in method labels.

## Supplementary Table S5. Paired RE preparation, adaptation, and baseline comparisons

| Model | Steps/client | Contrast | Source | MIMIC-III | MTSamples | UTP | i2b2 |
|---|---|---|---|---|---|---|---|
| Llama3.1 | 30 | SFT-Cont − No-Public 1R | +8.92 [1.04, 17.07] | +8.87 [1.95, 15.92] | +9.12 [2.01, 16.59] | +8.76 [−1.52, 19.80] | +8.01 [2.36, 14.59] |
| Llama3.1 | 30 | SFT-Cont − SFT-Res | +8.70 [3.32, 14.59] | +7.38 [1.97, 12.78] | +8.99 [3.42, 15.34] | +9.71 [3.92, 16.88] | +7.42 [2.98, 11.72] |
| Llama3.1 | 30 | SFT-Res − No-Public 1R | +0.22 [−4.70, 5.05] | +1.50 [−2.71, 5.82] | +0.13 [−4.15, 5.28] | −0.96 [−8.13, 5.07] | +0.59 [−2.49, 3.62] |
| Llama3.1 | 60 | SFT-Cont − No-Public 1R | +10.66 [4.51, 15.56] | +13.85 [7.11, 20.19] | +10.14 [3.75, 14.72] | +7.98 [2.76, 12.49] | +17.53 [8.92, 24.33] |
| Llama3.1 | 60 | SFT-Cont − SFT-Res | +7.06 [−7.04, 20.13] | +5.20 [−9.48, 17.83] | +7.13 [−6.98, 20.06] | +8.86 [−4.50, 22.14] | +8.26 [−10.63, 24.71] |
| Llama3.1 | 60 | SFT-Res − No-Public 1R | +3.59 [−5.01, 11.79] | +8.65 [1.49, 16.58] | +3.01 [−5.91, 10.89] | −0.88 [−10.56, 8.33] | +9.27 [−1.55, 20.19] |
| Llama3.1 | 120 | SFT-Cont − No-Public 1R | +5.56 [3.47, 8.32] | +6.79 [3.90, 10.45] | +5.47 [2.91, 8.34] | +4.43 [3.21, 6.37] | +8.33 [5.14, 12.72] |
| Llama3.1 | 120 | SFT-Cont − SFT-Res | +3.49 [1.94, 5.05] | +3.89 [1.96, 6.31] | +3.26 [1.92, 4.72] | +3.33 [1.53, 4.92] | +4.15 [1.84, 6.49] |
| Llama3.1 | 120 | SFT-Res − No-Public 1R | +2.07 [−0.20, 4.78] | +2.90 [−0.13, 6.56] | +2.21 [0.16, 4.74] | +1.11 [−0.91, 3.21] | +4.18 [0.71, 8.22] |
| Llama3.1 | 240 | SFT-Cont − No-Public 1R | +1.51 [0.25, 2.51] | +1.48 [0.02, 2.87] | +1.58 [0.08, 2.73] | +1.48 [0.26, 2.54] | +2.04 [0.89, 2.83] |

| Model | Steps/client | Contrast | Source | MIMIC-III | MTSamples | UTP | i2b2 |
|---|---|---|---|---|---|---|---|
| Llama3.1 | 240 | SFT-Cont − SFT-Res | +0.97 [−0.57, 2.65] | +0.23 [−1.51, 2.61] | +1.24 [−0.51, 2.91] | +1.42 [−0.21, 2.98] | +1.22 [−0.32, 2.71] |
| Llama3.1 | 240 | SFT-Res − No-Public 1R | +0.55 [−1.14, 1.69] | +1.25 [−0.55, 2.66] | +0.33 [−1.88, 1.86] | +0.05 [−1.15, 1.35] | +0.81 [−0.70, 1.78] |
| Llama3.1 | 480 | SFT-Cont − No-Public 1R | +1.67 [0.62, 2.71] | +1.26 [0.22, 2.49] | +1.27 [0.18, 2.55] | +2.48 [1.29, 3.62] | +1.48 [0.23, 2.78] |
| Llama3.1 | 480 | SFT-Cont − SFT-Res | +1.66 [0.61, 2.39] | +0.73 [−0.27, 1.62] | +1.71 [1.13, 2.26] | +2.55 [0.47, 3.95] | +1.54 [0.89, 2.36] |
| Llama3.1 | 480 | SFT-Res − No-Public 1R | +0.01 [−1.42, 1.54] | +0.53 [−0.95, 2.10] | −0.44 [−1.56, 0.76] | −0.06 [−2.16, 2.23] | −0.06 [−1.55, 1.27] |
| Qwen3.5 | 30 | Adaptation main effect: Cont − Res | +9.66 [6.96, 13.08] | +8.82 [6.10, 12.63] | +9.32 [7.04, 12.17] | +10.83 [7.57, 14.95] | +12.30 [8.37, 15.99] |
| Qwen3.5 | 30 | Meta-Cont − Meta-Res | +10.55 [7.41, 13.32] | +10.77 [7.70, 13.62] | +9.51 [7.42, 11.46] | +11.37 [6.76, 15.50] | +13.34 [8.23, 19.39] |
| Qwen3.5 | 30 | Meta-Cont − No-Public 1R | +12.45 [7.03, 18.30] | +14.45 [7.37, 21.57] | +11.62 [7.62, 16.02] | +11.29 [5.86, 17.49] | +16.70 [10.30, 23.04] |
| Qwen3.5 | 30 | Meta-Cont − SFT-Cont | +1.85 [−2.73, 9.84] | +2.56 [−3.13, 11.74] | +1.11 [−1.98, 7.54] | +1.88 [−3.49, 10.28] | +0.20 [−7.55, 12.58] |
| Qwen3.5 | 30 | Meta-Res − No-Public 1R | +1.90 [−5.97, 7.90] | +3.68 [−5.55, 11.22] | +2.12 [−3.60, 7.01] | −0.08 [−8.28, 6.37] | +3.36 [−8.22, 12.27] |
| Qwen3.5 | 30 | Meta-Res − SFT-Res | +0.06 [−8.49, 9.33] | −1.35 [−9.06, 8.38] | +0.73 [−6.88, 8.44] | +0.80 [−9.50, 11.54] | −1.90 [−14.70, 12.91] |
| Qwen3.5 | 30 | Preparation main effect: Meta − SFT | +0.95 [−4.28, 8.82] | +0.60 [−4.86, 9.18] | +0.92 [−3.97, 7.38] | +1.34 [−4.55, 10.08] | −0.85 [−9.70, 12.01] |
| Qwen3.5 | 30 | Preparation × adaptation | +1.79 [−4.78, 8.57] | +3.90 [−3.46, 10.84] | +0.38 [−5.90, 5.92] | +1.08 [−5.99, 9.94] | +2.10 [−7.24, 11.18] |
| Qwen3.5 | 30 | SFT-Cont − No-Public 1R | +10.61 [3.75, 19.17] | +11.89 [4.51, 21.64] | +10.51 [3.95, 17.02] | +9.41 [2.42, 18.76] | +16.50 [6.44, 27.92] |
| Qwen3.5 | 30 | SFT-Cont − SFT-Res | +8.76 [4.02, 15.19] | +6.87 [2.08, 14.22] | +9.12 [4.85, 14.94] | +10.29 [4.24, 17.17] | +11.25 [6.39, 18.87] |
| Qwen3.5 | 30 | SFT-Res − No-Public 1R | +1.84 [−2.66, 7.02] | +5.03 [0.49, 11.05] | +1.39 [−3.54, 6.21] | −0.88 [−6.13, 4.41] | +5.25 [−1.93, 13.02] |
| Qwen3.5 | 60 | Meta-Cont − Meta-Res | +5.10 [2.70, 7.26] | +5.52 [2.53, 8.36] | +3.90 [2.47, 5.37] | +5.90 [2.79, 8.87] | +5.83 [3.02, 9.47] |
| Qwen3.5 | 60 | Meta-Cont − No-Public 1R | +3.99 [1.38, 8.55] | +2.49 [−1.02, 8.00] | +3.80 [2.08, 6.69] | +5.68 [2.26, 10.96] | +4.11 [0.90, 9.69] |
| Qwen3.5 | 60 | Meta-Cont − SFT-Cont | +0.16 [−0.99, 1.43] | −0.01 [−1.77, 1.99] | −0.10 [−1.18, 0.84] | +0.58 [−1.36, 2.85] | −0.33 [−1.73, 0.96] |
| Qwen3.5 | 60 | Meta-Res − No-Public 1R | −1.12 [−4.34, 2.48] | −3.03 [−5.71, 0.37] | −0.09 [−2.13, 2.60] | −0.22 [−5.64, 4.95] | −1.72 [−4.81, 1.22] |
| Qwen3.5 | 60 | Meta-Res − SFT-Res | −0.42 [−2.61, 1.98] | −2.13 [−5.55, 1.61] | −0.38 [−1.59, 1.06] | +1.26 [−2.14, 4.46] | −2.02 [−7.43, 2.90] |
| Qwen3.5 | 60 | Preparation × adaptation | +0.57 [−2.61, 3.54] | +2.12 [−2.30, 6.17] | +0.28 [−1.45, 1.95] | −0.68 [−4.90, 3.03] | +1.69 [−3.91, 7.22] |
| Qwen3.5 | 60 | SFT-Cont − No-Public 1R | +3.83 [1.67, 8.38] | +2.50 [−0.62, 8.39] | +3.90 [1.80, 7.37] | +5.10 [2.58, 9.27] | +4.44 [1.03, 10.54] |
| Qwen3.5 | 60 | SFT-Cont − SFT-Res | +4.53 [2.41, 7.59] | +3.40 [0.59, 7.60] | +3.62 [2.01, 6.12] | +6.58 [3.97, 9.63] | +4.14 [0.40, 8.54] |
| Qwen3.5 | 60 | SFT-Res − No-Public 1R | −0.70 [−4.88, 4.53] | −0.90 [−6.86, 5.37] | +0.29 [−2.65, 3.90] | −1.48 [−5.65, 4.32] | +0.30 [−5.78, 7.47] |
| Qwen3.5 | 120 | Meta-Cont − Meta-Res | +1.81 [0.55, 3.11] | +2.09 [0.07, 3.98] | +0.98 [0.01, 1.95] | +2.34 [0.50, 3.78] | +1.10 [−1.09, 3.18] |
| Qwen3.5 | 120 | Meta-Cont − No-Public 1R | +1.55 [0.69, 2.37] | −0.61 [−2.08, 0.80] | +1.57 [1.21, 1.92] | +3.69 [2.64, 4.90] | −0.45 [−2.63, 1.33] |
| Qwen3.5 | 120 | Meta-Cont − SFT-Cont | −0.14 [−1.12, 0.82] | +0.03 [−1.95, 2.03] | −0.40 [−1.73, 0.53] | −0.06 [−0.77, 0.89] | −1.73 [−4.01, 0.01] |
| Qwen3.5 | 120 | Meta-Res − No-Public 1R | −0.26 [−1.16, 0.49] | −2.70 [−4.19, −1.25] | +0.58 [−0.20, 1.56] | +1.35 [−0.41, 2.73] | −1.55 [−2.72, −0.39] |
| Qwen3.5 | 120 | Meta-Res − SFT-Res | +0.35 [−0.75, 1.25] | −0.25 [−1.82, 1.18] | +0.56 [−0.53, 1.93] | +0.72 [−0.59, 2.37] | −0.55 [−2.33, 0.83] |
| Qwen3.5 | 120 | Preparation × adaptation | −0.49 [−1.76, 1.20] | +0.28 [−2.57, 2.99] | −0.96 [−2.05, 0.23] | −0.79 [−3.12, 1.13] | −1.18 [−3.81, 1.16] |
| Qwen3.5 | 120 | SFT-Cont − No-Public 1R | +1.69 [1.16, 2.23] | −0.65 [−1.73, 0.45] | +1.96 [1.09, 3.40] | +3.76 [2.69, 5.40] | +1.28 [0.71, 1.84] |

| Model | Steps/client | Contrast | Source | MIMIC-III | MTSamples | UTP | i2b2 |
|---|---|---|---|---|---|---|---|
| Qwen3.5 | 120 | SFT-Cont − SFT-Res | +2.29 [1.32, 3.24] | +1.81 [0.33, 2.97] | +1.94 [1.32, 2.51] | +3.13 [1.12, 5.52] | +2.28 [1.14, 3.54] |
| Qwen3.5 | 120 | SFT-Res − No-Public 1R | −0.60 [−1.54, 0.44] | −2.46 [−3.84, −1.34] | +0.02 [−1.21, 1.52] | +0.63 [−0.29, 1.74] | −1.00 [−2.43, 0.08] |
| Qwen3.5 | 240 | Meta-Cont − Meta-Res | +2.61 [1.66, 3.93] | +3.13 [1.45, 5.13] | +2.51 [1.30, 3.66] | +2.19 [0.97, 3.48] | +2.12 [1.22, 3.03] |
| Qwen3.5 | 240 | Meta-Cont − No-Public 1R | +0.94 [−0.88, 2.67] | −0.58 [−3.39, 2.05] | +1.97 [1.06, 3.11] | +1.45 [−1.27, 3.69] | +0.20 [−0.94, 1.58] |
| Qwen3.5 | 240 | Meta-Cont − SFT-Cont | +0.53 [−0.62, 1.63] | +0.95 [−1.04, 2.85] | +0.31 [−0.68, 1.41] | +0.33 [−0.84, 1.79] | +0.18 [−1.00, 1.52] |
| Qwen3.5 | 240 | Meta-Res − No-Public 1R | −1.66 [−2.94, −0.45] | −3.72 [−5.39, −1.96] | −0.54 [−1.44, 0.25] | −0.74 [−2.67, 0.92] | −1.92 [−3.34, −0.66] |
| Qwen3.5 | 240 | Meta-Res − SFT-Res | −0.17 [−1.19, 0.75] | −0.42 [−2.02, 0.75] | +0.15 [−0.70, 1.39] | −0.25 [−1.33, 0.73] | −0.21 [−1.10, 0.52] |
| Qwen3.5 | 240 | Preparation × adaptation | +0.71 [−0.73, 2.28] | +1.37 [−0.89, 3.48] | +0.16 [−1.56, 1.78] | +0.58 [−1.28, 2.71] | +0.40 [−1.11, 1.94] |
| Qwen3.5 | 240 | SFT-Cont − No-Public 1R | +0.41 [−1.28, 2.03] | −1.53 [−4.44, 1.02] | +1.65 [0.83, 2.49] | +1.12 [−0.77, 3.17] | +0.02 [−1.03, 0.98] |
| Qwen3.5 | 240 | SFT-Cont − SFT-Res | +1.90 [0.95, 2.85] | +1.76 [−0.23, 3.31] | +2.35 [1.43, 3.27] | +1.60 [0.45, 2.76] | +1.72 [0.68, 2.71] |
| Qwen3.5 | 240 | SFT-Res − No-Public 1R | −1.49 [−2.70, −0.27] | −3.29 [−5.34, −1.51] | −0.69 [−1.49, 0.05] | −0.49 [−2.34, 1.15] | −1.71 [−2.76, −0.69] |
| Qwen3.5 | 480 | Meta-Cont − Meta-Res | +1.46 [0.81, 1.98] | +2.20 [0.66, 3.33] | +1.49 [0.77, 2.09] | +0.69 [−0.12, 1.37] | +1.22 [0.39, 2.00] |
| Qwen3.5 | 480 | Meta-Cont − No-Public 1R | −0.07 [−1.15, 1.15] | −1.46 [−3.12, 0.25] | +1.48 [0.75, 2.38] | −0.23 [−1.91, 1.66] | −0.46 [−1.20, 0.50] |
| Qwen3.5 | 480 | Meta-Cont − SFT-Cont | +0.25 [−0.25, 0.87] | +0.74 [−0.30, 1.61] | +0.34 [−0.36, 1.13] | −0.32 [−1.15, 0.55] | −0.67 [−1.96, 0.56] |
| Qwen3.5 | 480 | Meta-Res − No-Public 1R | −1.53 [−2.66, −0.32] | −3.67 [−5.29, −2.19] | −0.01 [−0.92, 1.08] | −0.92 [−2.27, 0.46] | −1.67 [−2.98, −0.32] |
| Qwen3.5 | 480 | Meta-Res − SFT-Res | −0.42 [−1.39, 0.55] | −0.25 [−1.40, 1.20] | −0.07 [−1.37, 1.02] | −0.92 [−1.98, 0.28] | −0.87 [−1.92, 0.05] |
| Qwen3.5 | 480 | Preparation × adaptation | +0.67 [−0.11, 1.71] | +0.99 [−0.10, 2.09] | +0.41 [−0.83, 2.16] | +0.61 [−0.52, 1.58] | +0.20 [−1.14, 1.41] |
| Qwen3.5 | 480 | SFT-Cont − No-Public 1R | −0.33 [−1.45, 1.02] | −2.20 [−4.40, −0.10] | +1.13 [0.19, 2.22] | +0.08 [−1.43, 2.09] | +0.21 [−1.55, 1.86] |
| Qwen3.5 | 480 | SFT-Cont − SFT-Res | +0.79 [−0.13, 1.57] | +1.21 [−0.20, 2.70] | +1.08 [−0.35, 2.06] | +0.08 [−0.76, 0.92] | +1.02 [0.10, 2.17] |
| Qwen3.5 | 480 | SFT-Res − No-Public 1R | −1.12 [−2.09, 0.12] | −3.41 [−5.05, −1.52] | +0.06 [−0.97, 1.02] | +0.00 [−1.28, 1.39] | −0.80 [−1.82, 0.40] |
| Qwen3.5 | 30 | Fed-MedLoRA+ 1R − No-Public 1R | +0.75 [0.21, 1.50] | +0.81 [0.20, 1.60] | +0.57 [0.09, 1.33] | +0.87 [0.28, 1.67] | +0.88 [0.26, 1.53] |
| Qwen3.5 | 30 | Meta-Cont − Fed-MedLoRA+ 1R | +11.70 [6.79, 17.08] | +13.64 [7.13, 20.05] | +11.05 [7.47, 15.09] | +10.42 [5.44, 16.17] | +15.82 [9.91, 21.78] |
| Qwen3.5 | 30 | SFT-Cont − Fed-MedLoRA+ 1R | +9.86 [3.00, 18.07] | +11.08 [3.55, 20.60] | +9.94 [3.33, 16.16] | +8.54 [1.59, 17.58] | +15.62 [5.57, 26.76] |
| Qwen3.5 | 60 | Fed-MedLoRA+ 1R − No-Public 1R | −0.26 [−0.87, 0.66] | −0.79 [−2.11, 0.75] | +0.13 [−0.35, 0.88] | −0.13 [−0.65, 0.49] | +0.56 [−0.43, 2.33] |
| Qwen3.5 | 60 | Meta-Cont − Fed-MedLoRA+ 1R | +4.25 [1.76, 7.89] | +3.28 [−0.12, 7.37] | +3.67 [2.36, 5.85] | +5.81 [2.58, 10.66] | +3.56 [1.15, 7.33] |
| Qwen3.5 | 60 | SFT-Cont − Fed-MedLoRA+ 1R | +4.10 [2.23, 7.72] | +3.29 [0.70, 7.75] | +3.77 [2.02, 6.60] | +5.23 [2.98, 8.73] | +3.88 [0.88, 8.31] |
| Qwen3.5 | 120 | Fed-MedLoRA+ 1R − No-Public 1R | −0.18 [−0.54, 0.18] | −0.39 [−0.97, 0.21] | +0.07 [−0.30, 0.36] | −0.20 [−0.66, 0.22] | +0.11 [−0.15, 0.46] |
| Qwen3.5 | 120 | Meta-Cont − Fed-MedLoRA+ 1R | +1.72 [0.76, 2.63] | −0.22 [−1.88, 1.35] | +1.50 [0.95, 1.95] | +3.90 [2.72, 5.15] | −0.56 [−2.84, 1.21] |
| Qwen3.5 | 120 | SFT-Cont − Fed-MedLoRA+ 1R | +1.87 [1.59, 2.21] | −0.26 [−1.26, 0.78] | +1.90 [1.04, 3.20] | +3.96 [2.84, 5.57] | +1.17 [0.70, 1.70] |
| Qwen3.5 | 240 | Fed-MedLoRA+ 1R − No-Public 1R | −0.86 [−1.26, −0.49] | −1.32 [−2.28, −0.39] | +0.12 [−0.22, 0.46] | −1.37 [−2.13, −0.71] | −0.22 [−0.59, 0.14] |
| Qwen3.5 | 240 | Meta-Cont − Fed-MedLoRA+ 1R | +1.80 [0.09, 3.43] | +0.74 [−2.06, 3.02] | +1.85 [1.08, 2.96] | +2.82 [0.64, 4.87] | +0.43 [−0.54, 1.55] |
| Qwen3.5 | 240 | SFT-Cont − Fed-MedLoRA+ 1R | +1.27 [−0.33, 2.74] | −0.21 [−3.11, 2.38] | +1.54 [0.58, 2.49] | +2.49 [0.95, 4.19] | +0.24 [−0.78, 1.18] |

| Model | Steps/client | Contrast | Source | MIMIC-III | MTSamples | UTP | i2b2 |
|---|---|---|---|---|---|---|---|
| Qwen3.5 | 480 | Fed-MedLoRA+ 1R − No-Public 1R | −1.93 [−2.43, −1.38] | −2.31 [−3.25, −1.43] | −0.61 [−1.10, −0.09] | −2.86 [−3.60, −2.05] | −0.57 [−1.34, 0.23] |
| Qwen3.5 | 480 | Meta-Cont − Fed-MedLoRA+ 1R | +1.85 [0.74, 3.22] | +0.85 [−0.52, 2.28] | +2.08 [1.27, 2.80] | +2.63 [0.71, 5.13] | +0.11 [−0.91, 1.18] |
| Qwen3.5 | 480 | SFT-Cont − Fed-MedLoRA+ 1R | +1.60 [0.54, 3.07] | +0.11 [−1.64, 1.91] | +1.74 [0.54, 2.67] | +2.94 [1.24, 5.44] | +0.79 [−1.15, 2.35] |

Values are F1-point differences [paired 95% CI]; positive differences favor the first method. Preparation × adaptation is (Meta-Cont − Meta-Res) − (SFT-Cont − SFT-Res).

## Supplementary Table S6. Public preparation performance and computational cost

| Model | Preparation | Source NER F1 | MIMIC-III | MTSamples | UT Physicians | Preparation (GPU-hours) |
|---|---|---|---|---|---|---|
| Qwen3.5 | Backbone only | 4.66 | 6.42 | 4.42 | 3.14 | — |
| Qwen3.5 | SFT | 26.17 (23.04–28.47) | 28.63 (22.13–34.02) | 29.48 (25.65–31.51) | 20.40 (19.76–21.33) | 2.51 (2.502–2.532) |
| Qwen3.5 | Meta | 37.90 (35.59–42.13) | 38.51 (34.96–41.49) | 41.56 (39.35–44.53) | 33.63 (26.89–40.37) | 14.09 (13.962–14.276) |
| Llama3.1 | SFT | — | — | — | — | 0.34 (0.336–0.347) |

NER F1 is measured before clinical training. SFT and Meta entries are means across three public initializers, with initializer ranges in parentheses; GPU-hours were measured on one H200. Backbone only uses no adapter. A dash denotes an unreported score or inapplicable preparation cost.

**Supplementary Fig. S1. i2b2 NER performance across training budgets**

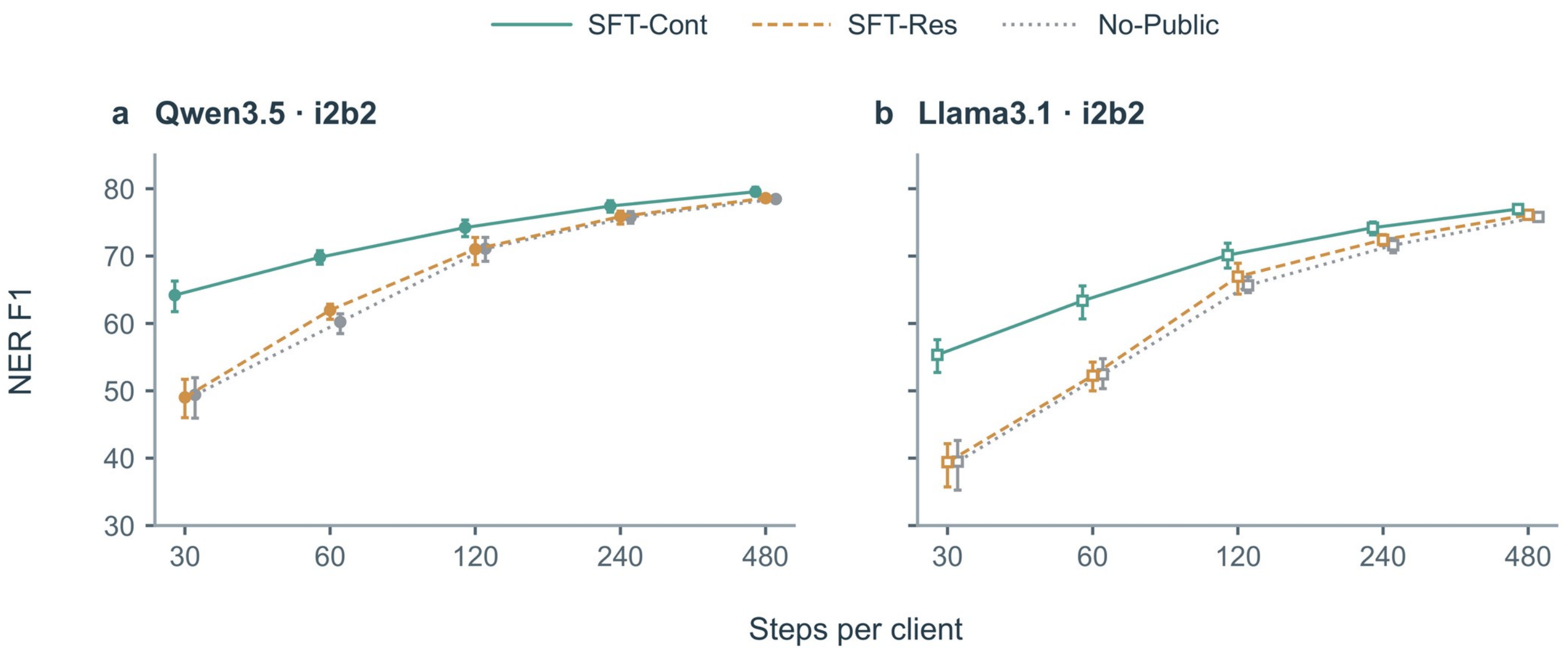

**a,** Qwen3.5. **b,** Llama3.1. Mean exact-match NER F1 on i2b2 across nine paired settings. Error bars show pointwise 95% hierarchical bootstrap CIs. All methods use one round. Source NER performance is shown in main Fig. 2; numerical results are in Table S2.

**Supplementary Fig. S2. Matched budget RE gains from public preparation and continuation**

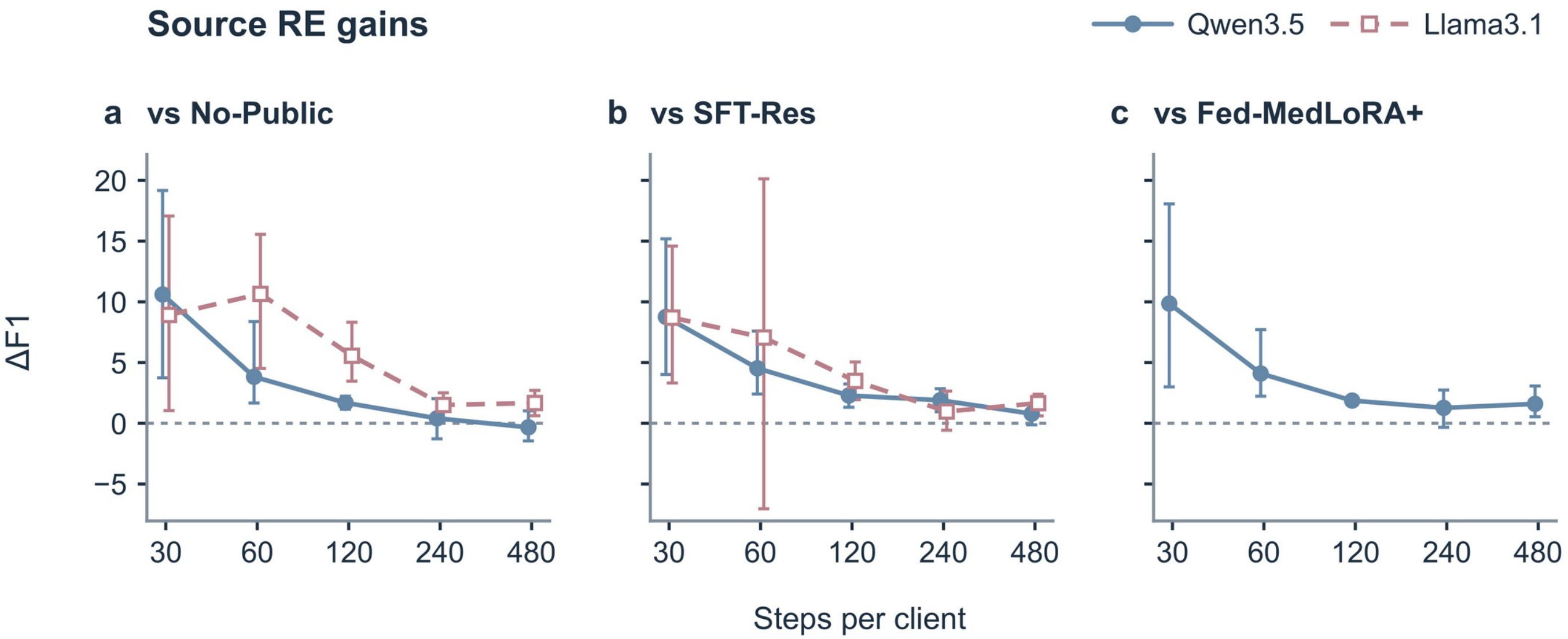

**a,** SFT-Cont minus No-Public. **b,** SFT-Cont minus SFT-Res. **c,** SFT-Cont minus Fed-MedLoRA+. Points show mean source RE F1 differences across nine paired settings; error bars show pointwise 95% paired hierarchical bootstrap CIs. Panel c shows Qwen3.5. All methods use one round. Positive differences favor SFT-Cont.

**Supplementary Fig. S3. RE performance across training budgets**

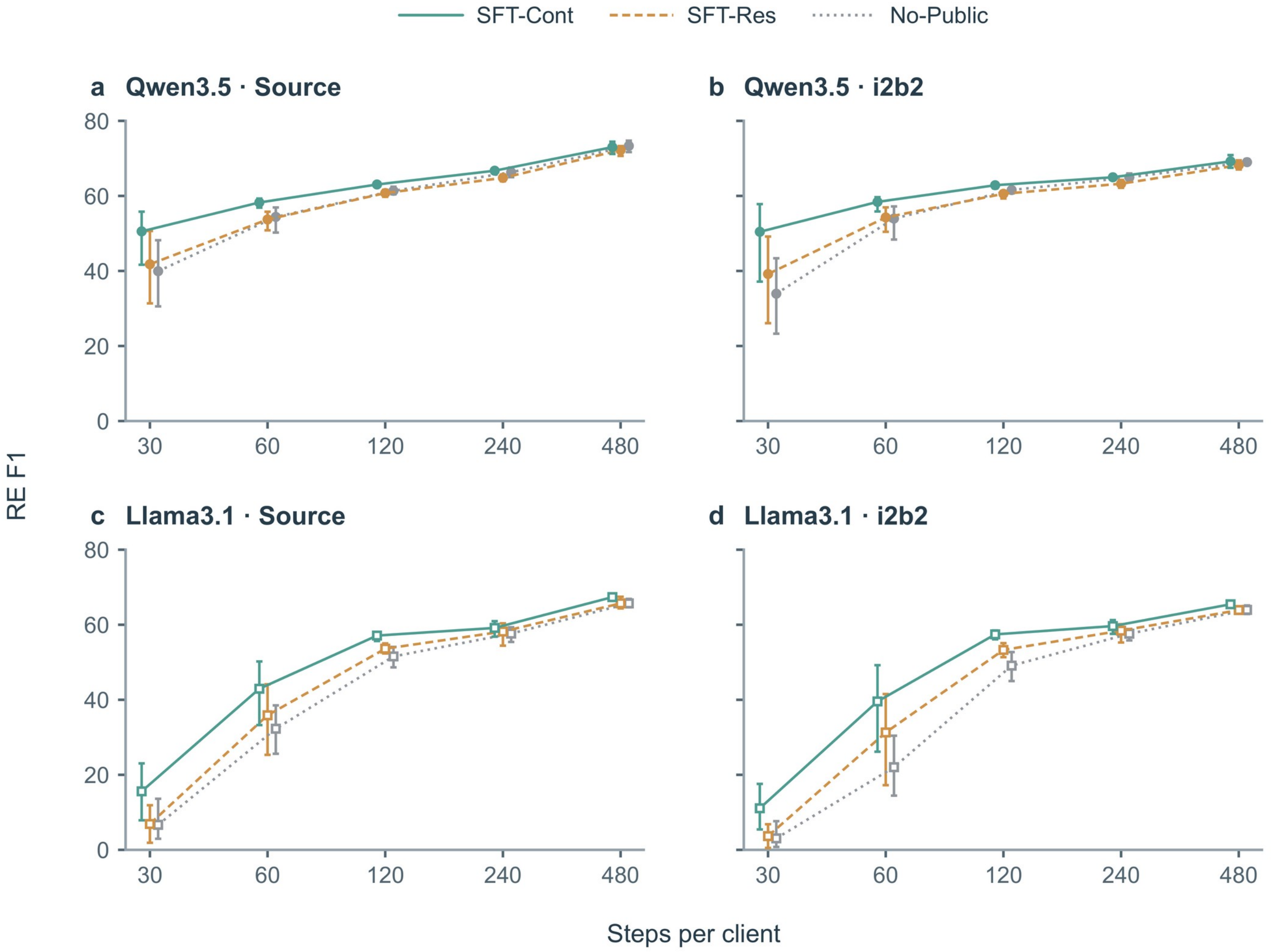

**a,** Qwen3.5 source mean. **b,** Qwen3.5 i2b2. **c,** Llama3.1 source mean. **d,** Llama3.1 i2b2. Points show mean exact-match RE F1 across nine paired settings; error bars show pointwise 95% hierarchical bootstrap CIs. All methods use one round. Numerical results are in Table S4.

**Supplementary Fig. S4. Matched budget NER gains on the external i2b2 test set**

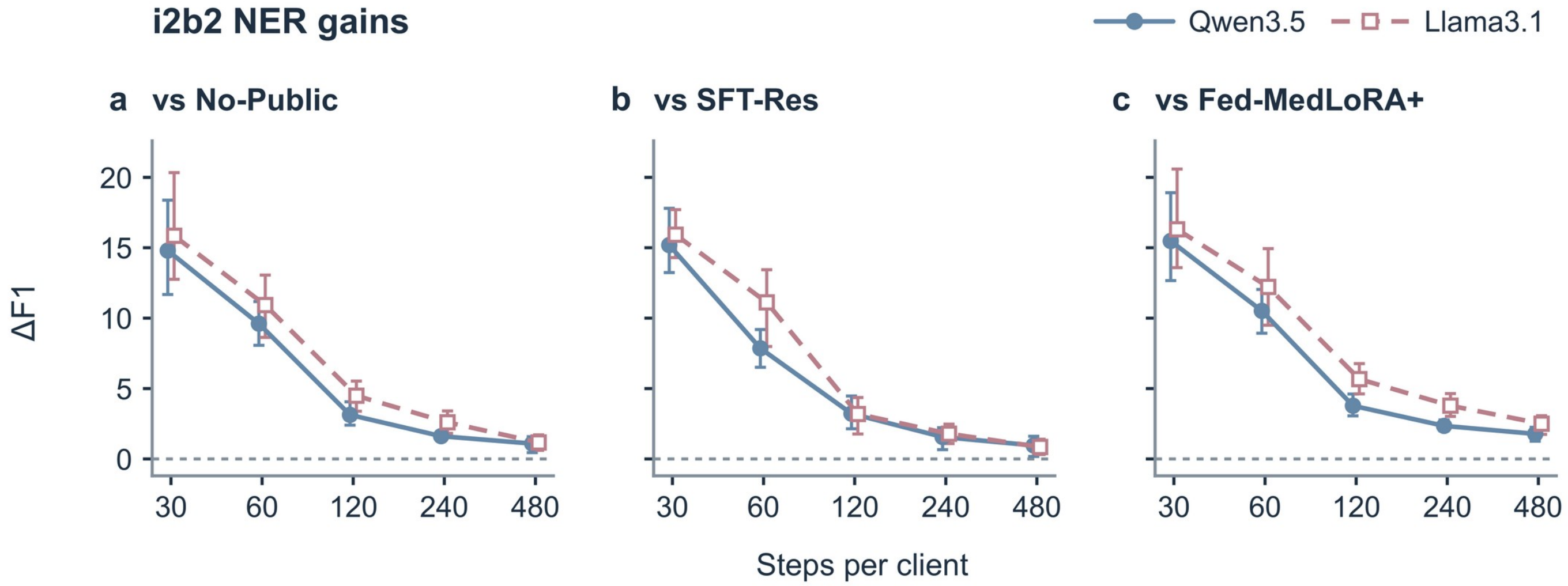

**a,** SFT-Cont minus No-Public. **b,** SFT-Cont minus SFT-Res. **c,** SFT-Cont minus Fed-MedLoRA+. Points show i2b2 NER F1 differences across nine paired settings; error bars show pointwise 95% paired hierarchical bootstrap CIs. All methods use one round. Positive differences favor SFT-Cont.

## S3 Private training budgets and communication rounds

Half-budget comparisons assess training and communication efficiency. The fixed-total comparison separates round count from the amount of private training.

### Supplementary Table S7. Private training budget and communication-round comparisons

| Task | Model | Steps/client | Contrast | Source | MIMIC-III | MTSamples | UTP | i2b2 |
|---|---|---|---|---|---|---|---|---|
| **A Half the private training budget** | | | | | | | | |
| NER | Llama3.1 | 30/60 | SFT-Cont (30 steps) − Fed-MedLoRA+ 1R (60 steps) | +6.14 [3.01, 8.83] | +3.89 [1.23, 6.21] | +4.76 [2.18, 6.99] | +9.77 [5.36, 13.73] | +4.18 [1.29, 6.84] |
| NER | Llama3.1 | 30/60 | SFT-Cont (30 steps) − Fed-MedLoRA+ 2R (60 steps) | +4.76 [2.15, 7.22] | +2.98 [0.46, 5.15] | +3.59 [0.89, 6.24] | +7.71 [4.27, 11.06] | +2.62 [−0.23, 5.13] |
| NER | Llama3.1 | 30/60 | SFT-Cont (30 steps) − No-Public 1R (60 steps) | +5.24 [2.31, 7.83] | +3.32 [0.58, 5.68] | +4.07 [1.68, 6.44] | +8.34 [4.30, 11.92] | +2.89 [0.13, 5.59] |
| NER | Llama3.1 | 30/60 | SFT-Cont (30 steps) − No-Public 2R (60 steps) | +11.02 [7.79, 14.91] | +8.70 [5.06, 13.18] | +9.19 [6.85, 12.26] | +15.18 [11.25, 19.55] | +8.61 [5.41, 12.00] |
| NER | Llama3.1 | 60/120 | SFT-Cont (60 steps) − Fed-MedLoRA+ 1R (120 steps) | −0.17 [−2.53, 2.09] | +0.28 [−1.04, 1.54] | −0.89 [−3.37, 1.36] | +0.10 [−3.49, 3.65] | −1.09 [−3.48, 1.17] |
| NER | Llama3.1 | 60/120 | SFT-Cont (60 steps) − Fed-MedLoRA+ 2R (120 steps) | −1.39 [−4.07, 0.95] | −0.31 [−1.95, 0.99] | −1.87 [−4.59, 0.20] | −1.98 [−6.03, 1.96] | −2.56 [−5.58, −0.22] |
| NER | Llama3.1 | 60/120 | SFT-Cont (60 steps) − No-Public 1R (120 steps) | −1.26 [−3.59, 0.82] | −0.17 [−1.49, 0.95] | −1.79 [−4.19, 0.16] | −1.81 [−5.40, 1.66] | −2.28 [−4.53, −0.40] |
| NER | Llama3.1 | 60/120 | SFT-Cont (60 steps) − No-Public 2R (120 steps) | +0.66 [−1.92, 2.98] | +1.14 [−0.34, 2.50] | −0.08 [−2.83, 2.31] | +0.92 [−2.84, 4.35] | −1.25 [−4.11, 1.19] |
| NER | Llama3.1 | 120/240 | SFT-Cont (120 steps) − Fed-MedLoRA+ 1R (240 steps) | −0.59 [−2.11, 0.76] | −0.27 [−1.69, 0.93] | −0.66 [−1.96, 0.61] | −0.83 [−2.85, 1.07] | −0.28 [−2.72, 1.91] |
| NER | Llama3.1 | 120/240 | SFT-Cont (120 steps) − Fed-MedLoRA+ 2R (240 steps) | −2.37 [−3.95, −0.92] | −1.49 [−2.90, −0.21] | −1.97 [−3.01, −0.78] | −3.67 [−6.02, −1.49] | −2.17 [−4.36, −0.22] |
| NER | Llama3.1 | 120/240 | SFT-Cont (120 steps) − No-Public 1R (240 steps) | −1.61 [−3.01, −0.25] | −1.00 [−2.42, 0.26] | −1.53 [−2.71, −0.26] | −2.30 [−4.20, −0.42] | −1.47 [−3.77, 0.58] |
| NER | Llama3.1 | 120/240 | SFT-Cont (120 steps) − No-Public 2R (240 steps) | −1.51 [−2.94, −0.10] | −0.64 [−2.06, 0.67] | −1.01 [−1.86, 0.26] | −2.89 [−4.97, −0.83] | −1.44 [−3.43, 0.39] |
| NER | Qwen3.5 | 30/60 | SFT-Cont (30 steps) − Fed-MedLoRA+ 1R (60 steps) | +3.22 [−0.33, 5.21] | +2.63 [0.71, 4.08] | +3.33 [1.13, 4.81] | +3.70 [−2.99, 7.40] | +4.89 [2.29, 7.07] |
| NER | Qwen3.5 | 30/60 | SFT-Cont (30 steps) − Fed-MedLoRA+ 2R (60 steps) | +4.58 [−0.05, 8.89] | +3.49 [0.71, 7.10] | +3.30 [0.55, 5.34] | +6.95 [−1.60, 14.88] | +4.58 [1.56, 7.17] |
| NER | Qwen3.5 | 30/60 | SFT-Cont (30 steps) − No-Public 1R (60 steps) | +2.47 [−1.37, 4.56] | +2.24 [0.24, 3.71] | +2.50 [−0.03, 4.23] | +2.67 [−4.40, 6.40] | +3.99 [1.11, 6.34] |
| NER | Qwen3.5 | 30/60 | SFT-Cont (30 steps) − No-Public 2R (60 steps) | +5.44 [0.86, 9.66] | +4.06 [1.31, 6.80] | +4.51 [1.79, 6.60] | +7.76 [−0.44, 15.76] | +5.94 [2.82, 8.57] |

| Task | Model | Steps/client | Contrast | Source | MIMIC-III | MTSamples | UTP | i2b2 |
|---|---|---|---|---|---|---|---|---|
| NER | Qwen3.5 | 60/120 | SFT-Cont (60 steps) − Fed-MedLoRA+ 1R (120 steps) | −0.33 [−3.72, 2.82] | −0.01 [−1.67, 1.74] | −1.71 [−4.07, 0.53] | +0.74 [−5.45, 6.52] | −0.62 [−3.19, 1.79] |
| NER | Qwen3.5 | 60/120 | SFT-Cont (60 steps) − Fed-MedLoRA+ 2R (120 steps) | −0.90 [−4.16, 2.32] | −0.30 [−2.10, 1.97] | −2.41 [−4.56, −0.20] | +0.00 [−5.77, 5.87] | −1.49 [−4.22, 1.01] |
| NER | Qwen3.5 | 60/120 | SFT-Cont (60 steps) − No-Public 1R (120 steps) | −1.05 [−4.32, 2.19] | −0.44 [−2.12, 1.42] | −2.45 [−4.53, −0.36] | −0.26 [−6.38, 5.78] | −1.27 [−3.76, 1.16] |
| NER | Qwen3.5 | 60/120 | SFT-Cont (60 steps) − No-Public 2R (120 steps) | +0.56 [−2.14, 3.35] | +0.40 [−1.09, 1.98] | −0.78 [−2.68, 1.16] | +2.05 [−2.83, 7.06] | −0.11 [−2.29, 1.99] |
| NER | Qwen3.5 | 120/240 | SFT-Cont (120 steps) − Fed-MedLoRA+ 1R (240 steps) | −0.82 [−1.77, 0.04] | −0.60 [−1.44, 0.14] | −1.24 [−2.37, −0.21] | −0.62 [−1.85, 0.69] | −0.85 [−2.02, 0.19] |
| NER | Qwen3.5 | 120/240 | SFT-Cont (120 steps) − Fed-MedLoRA+ 2R (240 steps) | −1.62 [−2.53, −0.92] | −1.27 [−2.16, −0.45] | −2.00 [−3.10, −0.97] | −1.58 [−2.77, −0.26] | −1.51 [−2.78, −0.53] |
| NER | Qwen3.5 | 120/240 | SFT-Cont (120 steps) − No-Public 1R (240 steps) | −1.55 [−2.38, −0.93] | −1.02 [−1.89, −0.29] | −1.82 [−2.91, −0.86] | −1.81 [−2.65, −0.95] | −1.58 [−2.82, −0.60] |
| NER | Qwen3.5 | 120/240 | SFT-Cont (120 steps) − No-Public 2R (240 steps) | −1.39 [−2.10, −0.78] | −1.12 [−1.75, −0.46] | −1.72 [−2.52, −1.01] | −1.34 [−2.49, −0.19] | −1.33 [−2.26, −0.38] |
| **B Matched-budget comparisons with two-round No-Public** | | | | | | | | |
| NER | Qwen3.5 | 60 | Meta-Cont − No-Public 2R | +10.27 [7.70, 13.87] | +7.25 [5.07, 9.96] | +9.79 [8.29, 11.75] | +13.79 [9.52, 20.17] | +10.98 [9.13, 13.19] |
| NER | Qwen3.5 | 60 | SFT-Cont − No-Public 2R | +10.79 [8.38, 13.87] | +7.55 [5.71, 10.12] | +9.82 [8.62, 11.31] | +15.02 [10.66, 20.48] | +11.57 [9.51, 13.72] |
| NER | Qwen3.5 | 120 | Meta-Cont − No-Public 2R | +4.61 [3.91, 5.34] | +3.33 [2.76, 3.90] | +4.31 [3.60, 5.04] | +6.18 [5.10, 7.23] | +4.66 [3.95, 5.38] |
| NER | Qwen3.5 | 120 | SFT-Cont − No-Public 2R | +4.61 [3.61, 5.61] | +3.47 [2.84, 4.02] | +4.02 [3.19, 4.87] | +6.34 [4.25, 8.45] | +4.29 [3.31, 5.10] |
| NER | Qwen3.5 | 240 | Meta-Cont − No-Public 2R | +1.61 [0.97, 2.20] | +1.15 [0.80, 1.45] | +1.41 [1.13, 1.73] | +2.25 [0.82, 3.88] | +2.27 [1.46, 3.09] |
| NER | Qwen3.5 | 240 | SFT-Cont − No-Public 2R | +1.40 [0.93, 1.92] | +1.19 [0.87, 1.54] | +1.30 [0.73, 1.72] | +1.71 [0.73, 2.77] | +1.87 [0.84, 2.77] |
| NER | Qwen3.5 | 480 | Meta-Cont − No-Public 2R | +0.43 [0.18, 0.66] | +0.38 [0.09, 0.68] | +0.54 [0.28, 0.74] | +0.38 [−0.49, 1.24] | +1.04 [0.62, 1.46] |
| NER | Qwen3.5 | 480 | SFT-Cont − No-Public 2R | +0.43 [0.12, 0.71] | +0.54 [0.27, 0.81] | +0.40 [0.17, 0.61] | +0.35 [−0.51, 1.06] | +1.03 [0.68, 1.32] |
| RE | Qwen3.5 | 60 | Meta-Cont − No-Public 2R | +6.52 [4.14, 8.87] | +5.28 [1.85, 8.76] | +5.93 [4.53, 7.61] | +8.36 [5.54, 11.41] | +7.92 [5.79, 10.65] |
| RE | Qwen3.5 | 60 | SFT-Cont − No-Public 2R | +6.37 [4.37, 9.02] | +5.29 [2.74, 8.63] | +6.03 [4.21, 8.40] | +7.78 [5.77, 10.14] | +8.24 [5.17, 12.23] |
| RE | Qwen3.5 | 120 | Meta-Cont − No-Public 2R | +1.20 [−0.12, 2.24] | −4.19 [−6.50, −2.47] | +3.14 [2.29, 3.91] | +4.64 [3.29, 6.26] | +2.01 [−0.33, 4.12] |
| RE | Qwen3.5 | 120 | SFT-Cont − No-Public 2R | +1.34 [0.64, 2.27] | −4.22 [−5.94, −2.38] | +3.54 [2.60, 4.53] | +4.71 [3.26, 6.74] | +3.74 [2.34, 5.34] |
| RE | Qwen3.5 | 240 | Meta-Cont − No-Public 2R | −2.74 [−4.58, −0.94] | −6.96 [−9.43, −4.94] | +1.13 [−0.58, 3.19] | −2.37 [−4.63, −0.47] | +0.01 [−0.84, 0.91] |
| RE | Qwen3.5 | 240 | SFT-Cont − No-Public 2R | −3.27 [−4.88, −1.76] | −7.91 [−10.46, −5.63] | +0.82 [−0.46, 2.10] | −2.71 [−4.14, −1.47] | −0.18 [−1.13, 0.84] |
| RE | Qwen3.5 | 480 | Meta-Cont − No-Public 2R | −3.30 [−4.10, −2.46] | −5.23 [−6.67, −3.76] | +0.45 [0.00, 0.96] | −5.12 [−6.37, −3.94] | −2.58 [−3.27, −1.72] |
| RE | Qwen3.5 | 480 | SFT-Cont − No-Public 2R | −3.55 [−4.72, −2.46] | −5.97 [−8.12, −3.84] | +0.11 [−0.67, 0.86] | −4.80 [−6.32, −3.20] | −1.91 [−3.61, −0.54] |

| Task | Model | Steps/client | Contrast | Source | MIMIC-III | MTSamples | UTP | i2b2 |
|---|---|---|---|---|---|---|---|---|
| **C One versus two rounds of SFT-Cont** | | | | | | | | |
| NER | Llama3.1 | 120/60 | SFT-Cont 2R (120 steps) − SFT-Cont (60 steps) | +4.90 [2.79, 7.08] | +3.20 [2.07, 4.50] | +5.17 [3.17, 7.34] | +6.34 [3.12, 9.98] | +6.56 [4.59, 8.84] |
| NER | Llama3.1 | 120 | SFT-Cont 2R − SFT-Cont | −0.36 [−1.66, 0.94] | −0.43 [−1.24, 0.31] | −0.74 [−1.71, 0.22] | +0.09 [−2.09, 2.79] | −0.20 [−1.59, 1.17] |
| NER | Qwen3.5 | 120/60 | SFT-Cont 2R (120 steps) − SFT-Cont (60 steps) | +3.38 [0.88, 5.77] | +2.51 [0.97, 4.08] | +3.91 [2.17, 5.61] | +3.71 [−0.69, 7.70] | +3.79 [1.87, 6.05] |
| NER | Qwen3.5 | 120 | SFT-Cont 2R − SFT-Cont | −0.67 [−1.30, −0.16] | −0.56 [−1.04, −0.16] | −0.88 [−1.49, −0.16] | −0.57 [−1.87, 0.16] | −0.62 [−1.18, 0.02] |

Values are F1-point differences [paired 95% CI]; positive differences favor the first method. A: SFT-Cont uses half the baseline steps, with one- or two-round baselines as labeled. B: all methods use the same total steps. C: 120 steps compares one round of 120 with two rounds of 60; 120/60 changes both total steps and round count. Budget pairs list the first method followed by the second.

# S4 Local training and federated aggregation

Local evaluations distinguish the benefit of public preparation from the additional effect of aggregation. Aggregation-rule comparisons use the same client adapters within each configuration.

### Supplementary Table S8. NER performance of local adapters

| Model | Steps/client | Method | Source | MIMIC-III | MTSamples | UTP |
|---|---|---|---|---|---|---|
| Llama3.1 | 30 | No-Public 1R | 42.61 [37.47, 45.27] | 45.27 [36.56, 50.47] | 43.28 [39.35, 46.53] | 39.29 [33.38, 43.39] |
| Llama3.1 | 30 | SFT-Cont | 56.44 [54.56, 58.40] | 57.46 [51.19, 61.91] | 55.32 [50.74, 60.05] | 56.54 [49.18, 62.81] |
| Llama3.1 | 60 | No-Public 1R | 53.93 [48.53, 57.68] | 50.35 [33.39, 59.19] | 55.60 [52.96, 58.49] | 55.85 [49.85, 60.29] |
| Llama3.1 | 60 | SFT-Cont | 61.75 [57.09, 66.17] | 55.19 [41.62, 64.73] | 63.46 [58.12, 67.21] | 66.59 [60.89, 71.44] |
| Llama3.1 | 120 | No-Public 1R | 66.73 [65.59, 67.64] | 67.08 [66.39, 67.95] | 64.24 [61.76, 66.45] | 68.85 [64.54, 71.59] |
| Llama3.1 | 120 | SFT-Cont | 71.52 [70.57, 72.38] | 69.24 [66.81, 71.32] | 69.85 [68.11, 71.95] | 75.48 [73.52, 77.08] |
| Qwen3.5 | 30 | No-Public 1R | 48.52 [41.33, 55.21] | 54.74 [47.00, 59.26] | 49.18 [39.42, 54.34] | 41.63 [25.69, 56.07] |
| Qwen3.5 | 30 | SFT-Cont | 61.75 [57.02, 65.81] | 63.55 [61.28, 65.74] | 59.43 [41.00, 66.75] | 62.27 [52.73, 70.17] |
| Qwen3.5 | 60 | No-Public 1R | 63.26 [61.31, 65.21] | 60.97 [56.82, 64.93] | 61.95 [57.33, 65.50] | 66.86 [62.93, 69.95] |
| Qwen3.5 | 60 | SFT-Cont | 70.02 [68.65, 70.92] | 68.02 [65.70, 69.77] | 68.70 [65.93, 70.86] | 73.32 [71.16, 74.74] |
| Qwen3.5 | 120 | No-Public 1R | 72.44 [70.85, 73.65] | 69.77 [68.17, 71.22] | 71.41 [69.62, 73.06] | 76.12 [73.78, 78.04] |
| Qwen3.5 | 120 | SFT-Cont | 74.53 [73.77, 75.27] | 72.53 [71.50, 73.56] | 72.98 [70.84, 75.36] | 78.07 [76.72, 79.79] |

Values are own-source F1 [95% CI]. Each local adapter is evaluated on its corresponding source test set; Source averages these three own-source scores. Global F1 is in Table S2 and paired aggregation effects are in Table S9.

### Supplementary Table S9. Paired local preparation and aggregation effects

| Model | Steps/client | Contrast | Source | MIMIC-III | MTSamples | UTP |
|---|---|---|---|---|---|---|
| Llama3.1 | 30 | (Global − local): SFT-Cont − No-Public | +4.85 [2.39, 7.18] | +3.78 [−0.13, 8.78] | +3.83 [0.51, 8.67] | +6.94 [1.63, 12.90] |
| Llama3.1 | 30 | No-Public: global − own-local | −0.79 [−2.33, 0.70] | +0.53 [−3.18, 4.75] | +0.70 [−2.17, 3.51] | −3.60 [−7.95, 0.23] |
| Llama3.1 | 30 | SFT-Cont − No-Public in local models | +13.82 [10.06, 18.21] | +12.19 [5.19, 18.39] | +12.04 [9.22, 14.59] | +17.25 [9.13, 26.55] |
| Llama3.1 | 30 | SFT-Cont: global − own-local | +4.06 [1.01, 6.41] | +4.32 [−1.54, 11.39] | +4.53 [0.87, 8.41] | +3.34 [−3.28, 11.01] |
| Llama3.1 | 60 | (Global − local): SFT-Cont − No-Public | +2.52 [−0.01, 6.67] | +2.86 [−2.19, 9.07] | +1.86 [−0.59, 4.92] | +2.86 [−3.26, 10.53] |
| Llama3.1 | 60 | No-Public 1R: global − own-local | +1.32 [−1.16, 5.26] | +8.10 [−0.18, 24.46] | +0.17 [−2.55, 2.58] | −4.30 [−7.95, −0.17] |
| Llama3.1 | 60 | SFT-Cont local − No-Public 1R local | +7.81 [4.79, 10.58] | +4.83 [−0.80, 10.30] | +7.86 [3.05, 10.66] | +10.75 [4.91, 15.61] |
| Llama3.1 | 60 | SFT-Cont: global − own-local | +3.85 [0.29, 7.85] | +10.96 [1.77, 24.38] | +2.03 [−0.69, 5.18] | −1.45 [−6.29, 5.04] |
| Llama3.1 | 120 | (Global − local): SFT-Cont − No-Public | −0.79 [−1.87, 0.40] | +1.30 [−0.37, 3.48] | −1.49 [−3.09, −0.16] | −2.19 [−4.62, −0.49] |
| Llama3.1 | 120 | No-Public: global − own-local | +0.13 [−1.09, 1.25] | −0.76 [−1.66, 0.39] | +3.04 [0.82, 5.26] | −1.90 [−4.95, 2.07] |
| Llama3.1 | 120 | SFT-Cont − No-Public in local models | +4.80 [3.43, 6.03] | +2.16 [0.03, 3.83] | +5.60 [4.17, 6.98] | +6.63 [4.49, 9.52] |
| Llama3.1 | 120 | SFT-Cont: global − own-local | −0.67 [−1.99, 0.46] | +0.53 [−1.36, 3.00] | +1.55 [0.23, 2.83] | −4.09 [−7.34, −1.56] |
| Qwen3.5 | 30 | (Global − local): SFT-Cont − No-Public | +1.58 [−1.70, 5.27] | +0.70 [−5.98, 7.04] | +1.80 [−4.82, 14.16] | +2.23 [−9.40, 14.76] |
| Qwen3.5 | 30 | No-Public: global − own-local | +1.98 [−2.39, 5.92] | +2.26 [−4.36, 11.06] | +3.55 [−0.40, 9.33] | +0.14 [−11.61, 11.68] |
| Qwen3.5 | 30 | SFT-Cont − No-Public in local models | +13.23 [7.92, 16.88] | +8.80 [5.07, 14.93] | +10.25 [1.87, 14.49] | +20.63 [2.87, 37.70] |
| Qwen3.5 | 30 | SFT-Cont: global − own-local | +3.56 [0.78, 7.77] | +2.97 [0.89, 5.72] | +5.34 [−2.12, 23.50] | +2.37 [−3.81, 9.01] |
| Qwen3.5 | 60 | (Global − local): SFT-Cont − No-Public | +1.07 [−0.64, 3.15] | −1.32 [−5.60, 2.93] | +1.05 [−4.55, 5.28] | +3.46 [−0.29, 6.96] |
| Qwen3.5 | 60 | No-Public 1R: global − own-local | −0.42 [−1.93, 1.08] | +3.30 [−0.18, 7.12] | +0.33 [−2.81, 5.22] | −4.89 [−7.61, −1.74] |
| Qwen3.5 | 60 | SFT-Cont local − No-Public 1R local | +6.75 [5.16, 8.39] | +7.05 [3.18, 11.37] | +6.75 [1.91, 12.44] | +6.46 [3.94, 9.82] |
| Qwen3.5 | 60 | SFT-Cont: global − own-local | +0.65 [−0.87, 1.97] | +1.98 [−0.10, 4.08] | +1.38 [−0.44, 3.66] | −1.43 [−5.27, 1.64] |
| Qwen3.5 | 120 | (Global − local): SFT-Cont − No-Public | +0.91 [0.28, 1.48] | −0.13 [−1.14, 1.02] | +0.78 [−0.53, 2.07] | +2.08 [0.50, 3.85] |
| Qwen3.5 | 120 | No-Public: global − own-local | −0.72 [−1.37, −0.11] | +0.67 [−0.31, 1.75] | +1.12 [−0.12, 2.28] | −3.96 [−5.68, −2.06] |
| Qwen3.5 | 120 | SFT-Cont − No-Public in local models | +2.09 [1.15, 3.18] | +2.76 [1.71, 3.61] | +1.57 [−0.01, 2.89] | +1.95 [−0.49, 4.23] |
| Qwen3.5 | 120 | SFT-Cont: global − own-local | +0.18 [−0.44, 0.84] | +0.54 [−0.60, 1.59] | +1.89 [0.25, 3.76] | −1.89 [−3.74, 0.24] |
| Llama3.1 | 120 − 30 | (Global − local): SFT-Cont − No-Public | −5.65 [−7.36, −3.69] | −2.49 [−7.14, 2.12] | −5.32 [−10.32, −1.19] | −9.14 [−14.40, −5.16] |
| Llama3.1 | 120 − 30 | SFT-Cont: global − own-local | −4.73 [−6.85, −1.99] | −3.79 [−10.88, 2.51] | −2.98 [−6.57, −0.06] | −7.43 [−15.07, −0.83] |
| Qwen3.5 | 120 − 30 | (Global − local): SFT-Cont − No-Public | −0.67 [−4.55, 2.46] | −0.84 [−8.01, 6.71] | −1.02 [−12.97, 6.06] | −0.16 [−12.87, 10.55] |
| Qwen3.5 | 120 − 30 | SFT-Cont: global − own-local | −3.38 [−8.09, −0.57] | −2.43 [−5.15, −0.23] | −3.45 [−21.48, 4.94] | −4.26 [−10.60, 1.40] |

Values are F1-point differences [paired 95% CI]; positive differences favor the first method. Global − local compares each global adapter with its corresponding own-source local adapter. The 120 − 30 rows show changes in the aggregation benefit across budgets.

## Supplementary Table S10. F1 under four aggregation rules

| Model | Task | Steps/client | Evaluation | A | B | C | D |
|---|---|---|---|---|---|---|---|
| Qwen3.5 | NER | 60 | Source | 70.66 | 70.66 | 70.46 | 70.45 |
| Qwen3.5 | NER | 60 | MIMIC-III | 70.00 | 70.00 | 69.94 | 69.91 |
| Qwen3.5 | NER | 60 | MTSamples | 70.08 | 70.11 | 70.01 | 70.03 |
| Qwen3.5 | NER | 60 | UTP | 71.90 | 71.87 | 71.42 | 71.40 |
| Qwen3.5 | NER | 60 | i2b2 | 69.83 | 69.85 | 69.80 | 69.83 |
| Llama3.1 | NER | 60 | Source | 65.59 | 65.62 | 65.56 | 65.67 |
| Llama3.1 | NER | 60 | MIMIC-III | 66.15 | 66.13 | 65.96 | 66.12 |
| Llama3.1 | NER | 60 | MTSamples | 65.49 | 65.56 | 65.47 | 65.54 |
| Llama3.1 | NER | 60 | UTP | 65.14 | 65.17 | 65.26 | 65.36 |
| Llama3.1 | NER | 60 | i2b2 | 63.37 | 63.64 | 63.40 | 63.34 |
| Qwen3.5 | NER | 120 | Source | 74.71 | — | — | 74.55 |
| Qwen3.5 | NER | 120 | MIMIC-III | 73.07 | — | — | 72.94 |
| Qwen3.5 | NER | 120 | MTSamples | 74.88 | — | — | 74.72 |
| Qwen3.5 | NER | 120 | UTP | 76.19 | — | — | 75.99 |
| Qwen3.5 | NER | 120 | i2b2 | 74.23 | — | — | 74.14 |
| Qwen3.5 | NER | 480 | Source | 79.40 | 79.37 | 79.17 | 79.12 |
| Qwen3.5 | NER | 480 | i2b2 | 79.55 | 79.59 | 79.36 | 79.45 |
| Qwen3.5 | RE | 60 | Source | 57.87 | 57.96 | 57.88 | 57.89 |
| Qwen3.5 | RE | 60 | i2b2 | 58.19 | 58.20 | 58.20 | 58.23 |

A: equal weights/effective-matrix averaging; B: validation-loss weights/effective-matrix averaging; C: equal weights/separate-factor averaging; D: validation-loss weights/separate-factor averaging. A dash denotes an unevaluated configuration. For Qwen3.5 RE with SFT-Cont at 60 steps, Tables S10 and S11 used client adapters from dedicated 60-step runs, and Table S4 used checkpoints saved at step 60 of 480-step runs.

## Supplementary Table S11. Paired aggregation-rule effects

| Model | Task | Steps/client | Effect | Source | MIMIC-III | MTSamples | UTP | i2b2 |
|---|---|---|---|---|---|---|---|---|
| Llama3.1 | NER | 60 | B − A | +0.03 [−0.09, 0.12] | −0.02 [−0.15, 0.10] | +0.07 [−0.07, 0.21] | +0.03 [−0.28, 0.27] | +0.26 [0.06, 0.46] |
| Llama3.1 | NER | 60 | C − A | −0.03 [−0.12, 0.07] | −0.19 [−0.46, 0.02] | −0.02 [−0.12, 0.08] | +0.11 [−0.12, 0.38] | +0.03 [−0.15, 0.20] |
| Llama3.1 | NER | 60 | D − A | +0.08 [−0.02, 0.19] | −0.03 [−0.28, 0.27] | +0.05 [−0.05, 0.17] | +0.22 [−0.05, 0.48] | −0.03 [−0.22, 0.17] |
| Llama3.1 | NER | 60 | Factor − effective | +0.01 [−0.08, 0.13] | −0.10 [−0.31, 0.11] | −0.02 [−0.10, 0.07] | +0.15 [−0.07, 0.47] | −0.13 [−0.24, −0.04] |
| Llama3.1 | NER | 60 | Weighting | +0.07 [−0.04, 0.16] | +0.07 [−0.07, 0.19] | +0.07 [0.00, 0.15] | +0.07 [−0.12, 0.25] | +0.10 [−0.05, 0.26] |
| Llama3.1 | NER | 60 | Interaction | +0.09 [−0.01, 0.16] | +0.18 [−0.05, 0.35] | +0.00 [−0.16, 0.17] | +0.08 [−0.09, 0.30] | −0.32 [−0.55, −0.16] |

| Model | Task | Steps/client | Effect | Source | MIMIC-III | MTSamples | UTP | i2b2 |
|---|---|---|---|---|---|---|---|---|
| Qwen3.5 | NER | 60 | B − A | +0.00 [−0.09, 0.11] | +0.00 [−0.07, 0.07] | +0.03 [−0.15, 0.23] | −0.02 [−0.13, 0.10] | +0.02 [−0.08, 0.14] |
| Qwen3.5 | NER | 60 | C − A | −0.20 [−0.40, −0.01] | −0.06 [−0.13, 0.02] | −0.08 [−0.27, 0.05] | −0.47 [−0.87, −0.05] | −0.03 [−0.26, 0.19] |
| Qwen3.5 | NER | 60 | D − A | −0.21 [−0.43, 0.00] | −0.09 [−0.25, 0.05] | −0.05 [−0.25, 0.14] | −0.50 [−0.85, −0.12] | +0.00 [−0.22, 0.23] |
| Qwen3.5 | NER | 60 | Factor − effective | −0.21 [−0.42, 0.03] | −0.07 [−0.18, 0.03] | −0.08 [−0.27, 0.13] | −0.47 [−0.85, −0.03] | −0.03 [−0.20, 0.14] |
| Qwen3.5 | NER | 60 | Weighting | −0.01 [−0.05, 0.04] | −0.02 [−0.07, 0.04] | +0.02 [−0.06, 0.10] | −0.02 [−0.10, 0.06] | +0.03 [−0.07, 0.11] |
| Qwen3.5 | NER | 60 | Interaction | −0.01 [−0.19, 0.12] | −0.03 [−0.16, 0.08] | −0.01 [−0.33, 0.27] | +0.00 [−0.23, 0.23] | +0.00 [−0.12, 0.12] |
| Qwen3.5 | NER | 120 | D − A | −0.16 [−0.23, −0.06] | −0.13 [−0.24, −0.02] | −0.16 [−0.26, −0.05] | −0.19 [−0.31, −0.07] | −0.10 [−0.31, 0.10] |
| Qwen3.5 | NER | 480 | B − A | −0.02 [−0.05, 0.00] | — | — | — | +0.04 [−0.04, 0.12] |
| Qwen3.5 | NER | 480 | C − A | −0.22 [−0.30, −0.14] | — | — | — | −0.19 [−0.33, −0.04] |
| Qwen3.5 | NER | 480 | D − A | −0.27 [−0.38, −0.18] | — | — | — | −0.10 [−0.30, 0.09] |
| Qwen3.5 | NER | 480 | Weighting | −0.04 [−0.09, 0.00] | — | — | — | +0.07 [−0.01, 0.13] |
| Qwen3.5 | NER | 480 | Factor − effective | −0.24 [−0.32, −0.16] | — | — | — | −0.16 [−0.30, −0.03] |
| Qwen3.5 | NER | 480 | Interaction | −0.03 [−0.09, 0.03] | — | — | — | +0.05 [−0.07, 0.23] |
| Qwen3.5 | RE | 60 | B − A | +0.09 [−0.02, 0.19] | — | — | — | +0.01 [−0.18, 0.20] |
| Qwen3.5 | RE | 60 | C − A | +0.01 [−0.40, 0.34] | — | — | — | +0.01 [−0.44, 0.42] |
| Qwen3.5 | RE | 60 | D − A | +0.02 [−0.37, 0.33] | — | — | — | +0.05 [−0.33, 0.43] |
| Qwen3.5 | RE | 60 | Weighting | +0.05 [−0.03, 0.11] | — | — | — | +0.02 [−0.16, 0.18] |
| Qwen3.5 | RE | 60 | Factor − effective | −0.03 [−0.46, 0.32] | — | — | — | +0.02 [−0.46, 0.46] |
| Qwen3.5 | RE | 60 | Interaction | −0.08 [−0.20, 0.02] | — | — | — | +0.02 [−0.08, 0.12] |

Values are F1-point differences [paired 95% CI]; positive differences favor the first method. Rules A–D are defined in Table S10. Weighting = [(B−A)+(D−C)]/2; factor − effective = [(C−A)+(D−B)]/2; interaction = (D−C)−(B−A).

**Supplementary Fig. S5. Aggregation effects on identical continued client adapters**

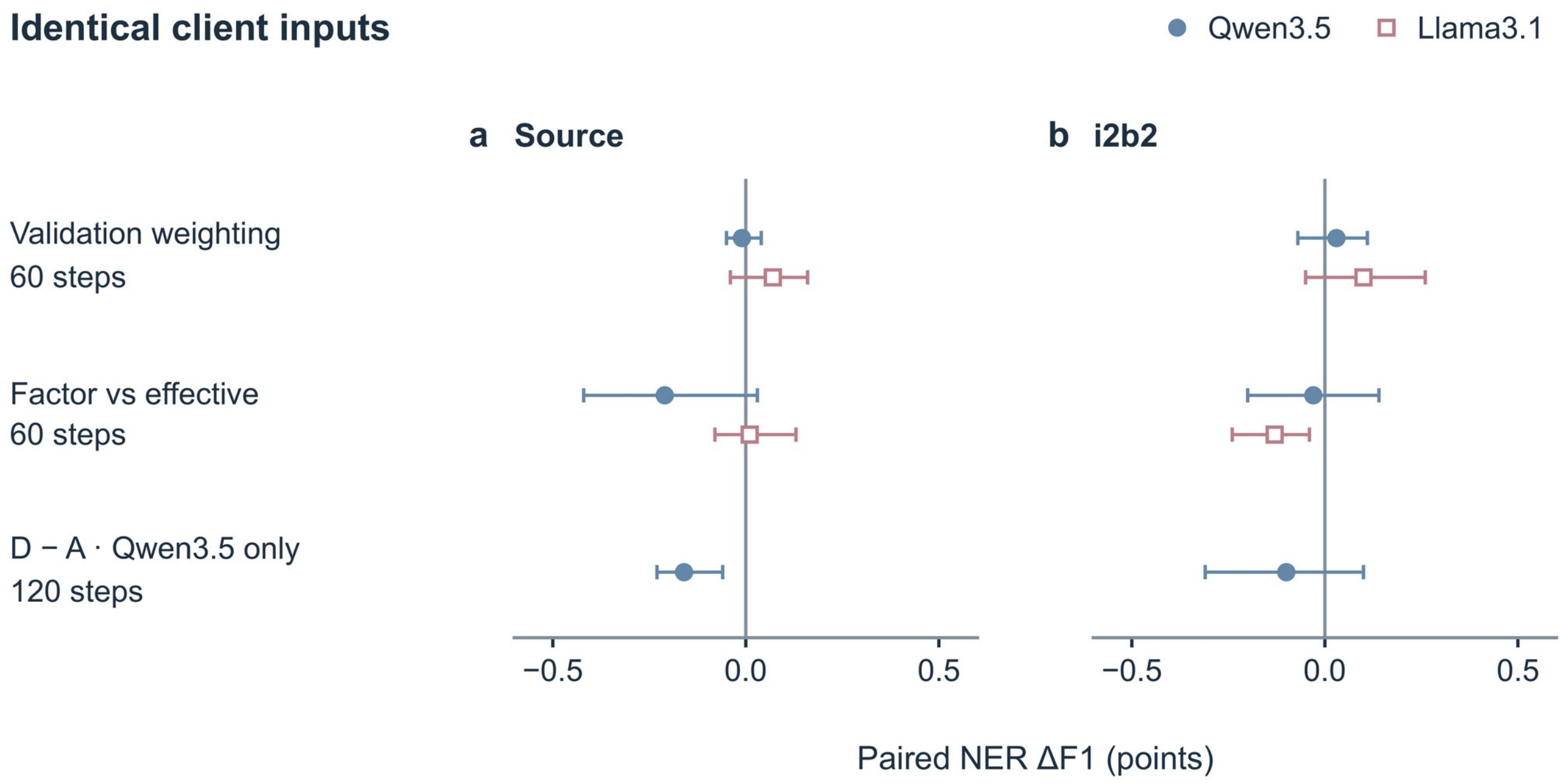

**a,** Mean source NER F1. **b,** i2b2 NER F1. Points show paired aggregation effects on the same continued client adapters across nine settings; error bars show pointwise 95% paired hierarchical bootstrap CIs. The 60-step rows compare weighting and factor representation; the 120-step row compares D with A on Qwen3.5. Rules and effect definitions are in Tables S10 and S11.

## S5 Source and label-specific error profiles

Overall error counts explain the precision–recall changes. Label-specific F1 profiles show how continuation compares with residual adaptation across budgets. Zero-denominator metrics are set to zero; label-specific scores exclude the untyped false-positive penalty for invalid outputs.

### Supplementary Table S12. NER precision, recall, and extraction counts

| Model | Steps/ client | Method | Domain | P (%) | R (%) | TP | FP | FN |
|---|---|---|---|---|---|---|---|---|
| Llama3.1 | 30 | No-Public 1R | Source | 54.77 | 34.57 | 1927.15 | 1640.52 | 3685.19 |
| Llama3.1 | 30 | No-Public 1R | i2b2 | 52.68 | 32.30 | 1243.33 | 1138.89 | 2605.67 |
| Llama3.1 | 30 | SFT-Cont | Source | 65.66 | 56.53 | 3172.67 | 1670.48 | 2439.67 |
| Llama3.1 | 30 | SFT-Cont | i2b2 | 60.40 | 51.35 | 1976.33 | 1305.44 | 1872.67 |
| Llama3.1 | 30 | SFT-Res | Source | 52.77 | 38.14 | 2105.67 | 1925.41 | 3506.67 |
| Llama3.1 | 30 | SFT-Res | i2b2 | 48.07 | 33.69 | 1296.78 | 1413.67 | 2552.22 |
| Llama3.1 | 60 | Fed-MedLoRA+ 1R | Source | 65.79 | 46.80 | 2617.89 | 1372.85 | 2994.44 |
| Llama3.1 | 60 | Fed-MedLoRA+ 1R | i2b2 | 61.75 | 44.02 | 1694.33 | 1054.56 | 2154.67 |
| Llama3.1 | 60 | No-Public 1R | Source | 66.48 | 47.80 | 2675.11 | 1362.48 | 2937.22 |
| Llama3.1 | 60 | No-Public 1R | i2b2 | 62.76 | 45.41 | 1747.78 | 1045.11 | 2101.22 |
| Llama3.1 | 60 | No-Public 2R | Source | 65.65 | 40.33 | 2250.63 | 1189.56 | 3361.70 |
| Llama3.1 | 60 | No-Public 2R | i2b2 | 61.23 | 38.21 | 1470.56 | 941.89 | 2378.44 |
| Llama3.1 | 60 | SFT-Cont | Source | 72.63 | 60.18 | 3373.30 | 1271.74 | 2239.04 |
| Llama3.1 | 60 | SFT-Cont | i2b2 | 70.01 | 58.15 | 2238.33 | 961.67 | 1610.67 |
| Llama3.1 | 60 | SFT-Res | Source | 67.27 | 49.32 | 2749.96 | 1330.78 | 2862.37 |
| Llama3.1 | 60 | SFT-Res | i2b2 | 62.48 | 45.05 | 1734.11 | 1047.89 | 2114.89 |
| Llama3.1 | 120 | Fed-MedLoRA+ 2R | Source | 73.83 | 61.46 | 3440.15 | 1221.41 | 2172.19 |
| Llama3.1 | 120 | Fed-MedLoRA+ 2R | i2b2 | 72.34 | 60.69 | 2335.89 | 896.89 | 1513.11 |
| Llama3.1 | 120 | No-Public 1R | Source | 73.16 | 61.71 | 3453.37 | 1270.07 | 2158.96 |
| Llama3.1 | 120 | No-Public 1R | i2b2 | 71.54 | 60.79 | 2339.67 | 934.67 | 1509.33 |
| Llama3.1 | 120 | No-Public 2R | Source | 73.78 | 58.24 | 3256.19 | 1163.22 | 2356.15 |
| Llama3.1 | 120 | No-Public 2R | i2b2 | 72.78 | 58.27 | 2242.67 | 843.67 | 1606.33 |
| Llama3.1 | 120 | SFT-Cont | Source | 75.88 | 66.65 | 3725.93 | 1182.33 | 1886.41 |
| Llama3.1 | 120 | SFT-Cont | i2b2 | 75.01 | 66.00 | 2540.44 | 849.11 | 1308.56 |
| Llama3.1 | 120 | SFT-Cont 2R | Source | 75.56 | 66.17 | 3703.37 | 1190.22 | 1908.96 |
| Llama3.1 | 120 | SFT-Cont 2R | i2b2 | 74.89 | 65.66 | 2527.11 | 848.33 | 1321.89 |

| Model | Steps/ client | Method | Domain | P (%) | R (%) | TP | FP | FN |
|---|---|---|---|---|---|---|---|---|
| Llama3.1 | 120 | SFT-Res | Source | 73.15 | 63.78 | 3570.04 | 1307.22 | 2042.30 |
| Llama3.1 | 120 | SFT-Res | i2b2 | 71.82 | 62.86 | 2419.33 | 950.11 | 1429.67 |
| Llama3.1 | 240 | No-Public 1R | Source | 76.61 | 68.81 | 3852.30 | 1174.44 | 1760.04 |
| Llama3.1 | 240 | No-Public 1R | i2b2 | 75.33 | 68.29 | 2628.44 | 861.44 | 1220.56 |
| Llama3.1 | 240 | No-Public 2R | Source | 77.54 | 67.87 | 3801.96 | 1092.78 | 1810.37 |
| Llama3.1 | 240 | No-Public 2R | i2b2 | 76.29 | 67.43 | 2595.33 | 806.44 | 1253.67 |
| Llama3.1 | 240 | SFT-Cont | Source | 77.44 | 72.64 | 4063.70 | 1181.48 | 1548.63 |
| Llama3.1 | 240 | SFT-Cont | i2b2 | 76.52 | 72.11 | 2775.33 | 854.56 | 1073.67 |
| Llama3.1 | 240 | SFT-Res | Source | 76.47 | 69.84 | 3907.78 | 1198.74 | 1704.56 |
| Llama3.1 | 240 | SFT-Res | i2b2 | 75.54 | 69.57 | 2677.67 | 867.44 | 1171.33 |
| Llama3.1 | 480 | No-Public 1R | Source | 79.85 | 73.04 | 4083.89 | 1034.67 | 1528.44 |
| Llama3.1 | 480 | No-Public 1R | i2b2 | 79.24 | 72.70 | 2798.11 | 734.11 | 1050.89 |
| Llama3.1 | 480 | SFT-Cont | Source | 80.44 | 74.61 | 4167.85 | 1016.56 | 1444.48 |
| Llama3.1 | 480 | SFT-Cont | i2b2 | 79.93 | 74.27 | 2858.78 | 719.22 | 990.22 |
| Llama3.1 | 480 | SFT-Res | Source | 79.83 | 73.32 | 4097.96 | 1039.56 | 1514.37 |
| Llama3.1 | 480 | SFT-Res | i2b2 | 79.46 | 73.13 | 2814.89 | 730.33 | 1034.11 |
| Qwen3.5 | 30 | Fed-MedLoRA+ 1R | Source | 63.69 | 41.58 | 2314.30 | 1343.19 | 3298.04 |
| Qwen3.5 | 30 | Fed-MedLoRA+ 1R | i2b2 | 58.66 | 42.21 | 1624.78 | 1161.56 | 2224.22 |
| Qwen3.5 | 30 | Meta-Cont | Source | 71.82 | 59.54 | 3346.19 | 1320.52 | 2266.15 |
| Qwen3.5 | 30 | Meta-Cont | i2b2 | 69.13 | 59.10 | 2274.89 | 1022.22 | 1574.11 |
| Qwen3.5 | 30 | Meta-Res | Source | 63.37 | 45.97 | 2581.44 | 1504.85 | 3030.89 |
| Qwen3.5 | 30 | Meta-Res | i2b2 | 57.36 | 42.70 | 1643.33 | 1239.22 | 2205.67 |
| Qwen3.5 | 30 | No-Public 1R | Source | 63.85 | 42.79 | 2383.00 | 1375.93 | 3229.33 |
| Qwen3.5 | 30 | No-Public 1R | i2b2 | 58.93 | 43.09 | 1658.56 | 1173.44 | 2190.44 |
| Qwen3.5 | 30 | SFT-Cont | Source | 71.50 | 60.93 | 3422.00 | 1382.63 | 2190.33 |
| Qwen3.5 | 30 | SFT-Cont | i2b2 | 69.07 | 60.45 | 2326.67 | 1053.67 | 1522.33 |
| Qwen3.5 | 30 | SFT-Res | Source | 62.21 | 45.82 | 2570.04 | 1589.00 | 3042.30 |
| Qwen3.5 | 30 | SFT-Res | i2b2 | 56.62 | 43.93 | 1690.89 | 1320.22 | 2158.11 |
| Qwen3.5 | 60 | Fed-MedLoRA+ 1R | Source | 72.26 | 54.70 | 3068.74 | 1167.30 | 2543.59 |
| Qwen3.5 | 60 | Fed-MedLoRA+ 1R | i2b2 | 66.83 | 53.49 | 2058.89 | 1026.22 | 1790.11 |
| Qwen3.5 | 60 | Meta-Cont | Source | 76.92 | 64.71 | 3627.74 | 1089.74 | 1984.59 |

| Model | Steps/ client | Method | Domain | P (%) | R (%) | TP | FP | FN |
|---|---|---|---|---|---|---|---|---|
| Qwen3.5 | 60 | Meta-Cont | i2b2 | 75.52 | 64.07 | 2466.11 | 805.11 | 1382.89 |
| Qwen3.5 | 60 | Meta-Res | Source | 70.90 | 58.14 | 3267.41 | 1337.15 | 2344.93 |
| Qwen3.5 | 60 | Meta-Res | i2b2 | 67.03 | 55.30 | 2128.67 | 1051.22 | 1720.33 |
| Qwen3.5 | 60 | No-Public 1R | Source | 72.38 | 55.82 | 3130.63 | 1186.11 | 2481.70 |
| Qwen3.5 | 60 | No-Public 1R | i2b2 | 67.49 | 54.55 | 2099.67 | 1016.11 | 1749.33 |
| Qwen3.5 | 60 | No-Public 2R | Source | 71.81 | 52.00 | 2908.67 | 1134.70 | 2703.67 |
| Qwen3.5 | 60 | No-Public 2R | i2b2 | 66.98 | 51.87 | 1996.33 | 991.33 | 1852.67 |
| Qwen3.5 | 60 | SFT-Cont | Source | 76.31 | 65.99 | 3706.15 | 1147.22 | 1906.19 |
| Qwen3.5 | 60 | SFT-Cont | i2b2 | 75.31 | 65.22 | 2510.22 | 827.78 | 1338.78 |
| Qwen3.5 | 60 | SFT-Res | Source | 70.81 | 59.56 | 3350.33 | 1372.52 | 2262.00 |
| Qwen3.5 | 60 | SFT-Res | i2b2 | 67.19 | 57.59 | 2216.67 | 1083.78 | 1632.33 |
| Qwen3.5 | 120 | Fed-MedLoRA+ 1R | Source | 77.38 | 65.78 | 3677.59 | 1076.48 | 1934.74 |
| Qwen3.5 | 120 | Fed-MedLoRA+ 1R | i2b2 | 76.06 | 65.75 | 2530.67 | 798.33 | 1318.33 |
| Qwen3.5 | 120 | Fed-MedLoRA+ 2R | Source | 78.18 | 66.17 | 3699.63 | 1034.00 | 1912.70 |
| Qwen3.5 | 120 | Fed-MedLoRA+ 2R | i2b2 | 77.48 | 66.21 | 2548.33 | 742.33 | 1300.67 |
| Qwen3.5 | 120 | Meta-Cont | Source | 78.75 | 71.17 | 3986.81 | 1078.52 | 1625.52 |
| Qwen3.5 | 120 | Meta-Cont | i2b2 | 78.75 | 70.98 | 2732.11 | 739.89 | 1116.89 |
| Qwen3.5 | 120 | Meta-Res | Source | 76.18 | 67.11 | 3760.26 | 1175.37 | 1852.07 |
| Qwen3.5 | 120 | Meta-Res | i2b2 | 75.83 | 66.23 | 2549.11 | 815.67 | 1299.89 |
| Qwen3.5 | 120 | No-Public 1R | Source | 77.73 | 66.77 | 3734.22 | 1072.37 | 1878.11 |
| Qwen3.5 | 120 | No-Public 1R | i2b2 | 76.64 | 66.47 | 2558.56 | 782.11 | 1290.44 |
| Qwen3.5 | 120 | No-Public 2R | Source | 77.66 | 64.03 | 3578.44 | 1026.85 | 2033.89 |
| Qwen3.5 | 120 | No-Public 2R | i2b2 | 76.97 | 64.16 | 2469.67 | 738.67 | 1379.33 |
| Qwen3.5 | 120 | SFT-Cont | Source | 78.41 | 71.46 | 4006.33 | 1105.44 | 1606.00 |
| Qwen3.5 | 120 | SFT-Cont | i2b2 | 78.22 | 70.76 | 2723.56 | 762.78 | 1125.44 |
| Qwen3.5 | 120 | SFT-Cont 2R | Source | 79.08 | 69.65 | 3905.85 | 1028.74 | 1706.48 |
| Qwen3.5 | 120 | SFT-Cont 2R | i2b2 | 79.10 | 68.90 | 2652.00 | 701.11 | 1197.00 |
| Qwen3.5 | 120 | SFT-Res | Source | 76.31 | 67.25 | 3774.15 | 1165.67 | 1838.19 |
| Qwen3.5 | 120 | SFT-Res | i2b2 | 76.03 | 66.77 | 2569.89 | 811.22 | 1279.11 |
| Qwen3.5 | 240 | Fed-MedLoRA+ 1R | Source | 80.41 | 71.25 | 3988.15 | 968.89 | 1624.19 |
| Qwen3.5 | 240 | Fed-MedLoRA+ 1R | i2b2 | 79.64 | 71.05 | 2734.78 | 699.67 | 1114.22 |

| Model | Steps/ client | Method | Domain | P (%) | R (%) | TP | FP | FN |
|---|---|---|---|---|---|---|---|---|
| Qwen3.5 | 240 | Fed-MedLoRA+ 2R | Source | 81.13 | 72.10 | 4036.37 | 936.78 | 1575.96 |
| Qwen3.5 | 240 | Fed-MedLoRA+ 2R | i2b2 | 80.45 | 71.58 | 2755.00 | 669.67 | 1094.00 |
| Qwen3.5 | 240 | Meta-Cont | Source | 80.70 | 74.96 | 4199.85 | 1000.63 | 1412.48 |
| Qwen3.5 | 240 | Meta-Cont | i2b2 | 80.95 | 74.98 | 2885.89 | 680.33 | 963.11 |
| Qwen3.5 | 240 | Meta-Res | Source | 79.50 | 73.23 | 4102.85 | 1055.33 | 1509.48 |
| Qwen3.5 | 240 | Meta-Res | i2b2 | 79.14 | 72.85 | 2803.89 | 739.89 | 1045.11 |
| Qwen3.5 | 240 | No-Public 1R | Source | 80.76 | 72.27 | 4047.15 | 960.89 | 1565.19 |
| Qwen3.5 | 240 | No-Public 1R | i2b2 | 80.20 | 71.89 | 2767.22 | 683.11 | 1081.78 |
| Qwen3.5 | 240 | No-Public 2R | Source | 81.32 | 71.53 | 4004.59 | 916.15 | 1607.74 |
| Qwen3.5 | 240 | No-Public 2R | i2b2 | 80.93 | 70.87 | 2727.78 | 642.89 | 1121.22 |
| Qwen3.5 | 240 | SFT-Cont | Source | 80.82 | 74.48 | 4172.37 | 988.37 | 1439.96 |
| Qwen3.5 | 240 | SFT-Cont | i2b2 | 80.86 | 74.31 | 2860.00 | 677.67 | 989.00 |
| Qwen3.5 | 240 | SFT-Res | Source | 80.21 | 72.61 | 4072.22 | 1000.70 | 1540.11 |
| Qwen3.5 | 240 | SFT-Res | i2b2 | 79.92 | 72.24 | 2780.33 | 699.22 | 1068.67 |
| Qwen3.5 | 480 | Fed-MedLoRA+ 1R | Source | 82.39 | 73.82 | 4131.59 | 879.89 | 1480.74 |
| Qwen3.5 | 480 | Fed-MedLoRA+ 1R | i2b2 | 82.41 | 73.64 | 2834.22 | 605.00 | 1014.78 |
| Qwen3.5 | 480 | Meta-Cont | Source | 82.56 | 76.48 | 4281.81 | 903.44 | 1330.52 |
| Qwen3.5 | 480 | Meta-Cont | i2b2 | 82.95 | 76.47 | 2943.22 | 605.33 | 905.78 |
| Qwen3.5 | 480 | Meta-Res | Source | 82.28 | 76.16 | 4265.07 | 918.00 | 1347.26 |
| Qwen3.5 | 480 | Meta-Res | i2b2 | 82.66 | 75.88 | 2920.44 | 612.89 | 928.56 |
| Qwen3.5 | 480 | No-Public 1R | Source | 82.64 | 74.96 | 4195.33 | 878.70 | 1417.00 |
| Qwen3.5 | 480 | No-Public 1R | i2b2 | 82.71 | 74.65 | 2873.11 | 601.11 | 975.89 |
| Qwen3.5 | 480 | No-Public 2R | Source | 83.07 | 75.26 | 4213.59 | 855.70 | 1398.74 |
| Qwen3.5 | 480 | No-Public 2R | i2b2 | 83.09 | 74.46 | 2865.89 | 583.78 | 983.11 |
| Qwen3.5 | 480 | SFT-Cont | Source | 82.59 | 76.47 | 4282.41 | 901.00 | 1329.93 |
| Qwen3.5 | 480 | SFT-Cont | i2b2 | 82.97 | 76.43 | 2941.67 | 604.33 | 907.33 |
| Qwen3.5 | 480 | SFT-Res | Source | 82.65 | 74.96 | 4198.56 | 879.93 | 1413.78 |
| Qwen3.5 | 480 | SFT-Res | i2b2 | 82.79 | 74.83 | 2880.11 | 599.33 | 968.89 |

Precision, recall and counts are averaged across nine settings; Source additionally averages the three collections. Overall FP includes one untyped penalty per invalid output.

### Supplementary Table S13. RE precision, recall, and extraction counts

| Model | Steps/ client | Method | Domain | P (%) | R (%) | TP | FP | FN |
|---|---|---|---|---|---|---|---|---|
| Llama3.1 | 30 | No-Public 1R | Source | 64.34 | 3.80 | 212.48 | 126.56 | 5731.52 |
| Llama3.1 | 30 | No-Public 1R | i2b2 | 47.43 | 1.71 | 59.11 | 67.33 | 3389.89 |
| Llama3.1 | 30 | SFT-Cont | Source | 78.51 | 9.42 | 539.93 | 167.93 | 5404.07 |
| Llama3.1 | 30 | SFT-Cont | i2b2 | 70.83 | 6.41 | 221.00 | 94.67 | 3228.00 |
| Llama3.1 | 30 | SFT-Res | Source | 65.24 | 3.84 | 213.78 | 108.56 | 5730.22 |
| Llama3.1 | 30 | SFT-Res | i2b2 | 57.28 | 1.97 | 67.78 | 61.22 | 3381.22 |
| Llama3.1 | 60 | No-Public 1R | Source | 74.50 | 21.84 | 1284.26 | 476.89 | 4659.74 |
| Llama3.1 | 60 | No-Public 1R | i2b2 | 65.48 | 14.57 | 502.44 | 329.56 | 2946.56 |
| Llama3.1 | 60 | SFT-Cont | Source | 75.28 | 32.58 | 1918.04 | 743.33 | 4025.96 |
| Llama3.1 | 60 | SFT-Cont | i2b2 | 71.86 | 30.65 | 1057.11 | 514.33 | 2391.89 |
| Llama3.1 | 60 | SFT-Res | Source | 72.96 | 25.16 | 1482.70 | 578.33 | 4461.30 |
| Llama3.1 | 60 | SFT-Res | i2b2 | 69.90 | 21.76 | 750.67 | 372.22 | 2698.33 |
| Llama3.1 | 120 | No-Public 1R | Source | 75.72 | 39.58 | 2343.00 | 739.81 | 3601.00 |
| Llama3.1 | 120 | No-Public 1R | i2b2 | 72.24 | 37.87 | 1306.22 | 529.11 | 2142.78 |
| Llama3.1 | 120 | SFT-Cont | Source | 75.24 | 46.28 | 2740.30 | 887.41 | 3203.70 |
| Llama3.1 | 120 | SFT-Cont | i2b2 | 73.43 | 47.44 | 1636.11 | 607.56 | 1812.89 |
| Llama3.1 | 120 | SFT-Res | Source | 76.72 | 41.41 | 2448.00 | 722.19 | 3496.00 |
| Llama3.1 | 120 | SFT-Res | i2b2 | 74.70 | 41.57 | 1433.67 | 493.89 | 2015.33 |
| Llama3.1 | 240 | No-Public 1R | Source | 79.84 | 45.39 | 2689.37 | 675.93 | 3254.63 |
| Llama3.1 | 240 | No-Public 1R | i2b2 | 77.64 | 46.01 | 1586.78 | 466.00 | 1862.22 |
| Llama3.1 | 240 | SFT-Cont | Source | 80.75 | 46.98 | 2781.07 | 658.59 | 3162.93 |
| Llama3.1 | 240 | SFT-Cont | i2b2 | 78.17 | 48.46 | 1671.44 | 475.11 | 1777.56 |
| Llama3.1 | 240 | SFT-Res | Source | 79.64 | 46.20 | 2739.30 | 698.44 | 3204.70 |
| Llama3.1 | 240 | SFT-Res | i2b2 | 77.26 | 47.30 | 1631.22 | 490.89 | 1817.78 |
| Llama3.1 | 480 | No-Public 1R | Source | 80.83 | 55.67 | 3265.33 | 749.44 | 2678.67 |
| Llama3.1 | 480 | No-Public 1R | i2b2 | 76.84 | 54.89 | 1893.11 | 573.78 | 1555.89 |
| Llama3.1 | 480 | SFT-Cont | Source | 80.35 | 58.24 | 3423.81 | 799.96 | 2520.19 |
| Llama3.1 | 480 | SFT-Cont | i2b2 | 76.21 | 57.39 | 1979.44 | 620.00 | 1469.56 |
| Llama3.1 | 480 | SFT-Res | Source | 80.15 | 55.92 | 3292.41 | 788.93 | 2651.59 |
| Llama3.1 | 480 | SFT-Res | i2b2 | 76.01 | 55.23 | 1904.89 | 604.89 | 1544.11 |

| Model | Steps/ client | Method | Domain | P (%) | R (%) | TP | FP | FN |
|---|---|---|---|---|---|---|---|---|
| Qwen3.5 | 30 | Fed-MedLoRA+ 1R | Source | 71.16 | 30.72 | 1786.41 | 860.78 | 4157.59 |
| Qwen3.5 | 30 | Fed-MedLoRA+ 1R | i2b2 | 62.11 | 27.73 | 956.56 | 764.78 | 2492.44 |
| Qwen3.5 | 30 | Meta-Cont | Source | 75.73 | 41.14 | 2398.44 | 801.00 | 3545.56 |
| Qwen3.5 | 30 | Meta-Cont | i2b2 | 71.51 | 41.17 | 1419.89 | 630.56 | 2029.11 |
| Qwen3.5 | 30 | Meta-Res | Source | 74.84 | 30.25 | 1741.70 | 636.33 | 4202.30 |
| Qwen3.5 | 30 | Meta-Res | i2b2 | 67.37 | 27.66 | 954.00 | 536.00 | 2495.00 |
| Qwen3.5 | 30 | No-Public 1R | Source | 71.77 | 30.04 | 1745.00 | 830.89 | 4199.00 |
| Qwen3.5 | 30 | No-Public 1R | i2b2 | 62.55 | 26.95 | 929.44 | 738.11 | 2519.56 |
| Qwen3.5 | 30 | SFT-Cont | Source | 75.79 | 39.37 | 2280.59 | 758.11 | 3663.41 |
| Qwen3.5 | 30 | SFT-Cont | i2b2 | 73.02 | 40.98 | 1413.44 | 586.44 | 2035.56 |
| Qwen3.5 | 30 | SFT-Res | Source | 71.46 | 31.79 | 1839.48 | 832.41 | 4104.52 |
| Qwen3.5 | 30 | SFT-Res | i2b2 | 65.36 | 31.68 | 1092.56 | 711.44 | 2356.44 |
| Qwen3.5 | 60 | Fed-MedLoRA+ 1R | Source | 73.96 | 43.23 | 2533.67 | 903.74 | 3410.33 |
| Qwen3.5 | 60 | Fed-MedLoRA+ 1R | i2b2 | 71.42 | 44.89 | 1548.11 | 655.11 | 1900.89 |
| Qwen3.5 | 60 | Meta-Cont | Source | 77.21 | 47.23 | 2779.85 | 803.59 | 3164.15 |
| Qwen3.5 | 60 | Meta-Cont | i2b2 | 74.23 | 48.01 | 1655.89 | 589.78 | 1793.11 |
| Qwen3.5 | 60 | Meta-Res | Source | 77.84 | 40.87 | 2379.63 | 662.33 | 3564.37 |
| Qwen3.5 | 60 | Meta-Res | i2b2 | 74.38 | 40.69 | 1403.44 | 497.44 | 2045.56 |
| Qwen3.5 | 60 | No-Public 1R | Source | 74.22 | 43.62 | 2561.63 | 902.15 | 3382.37 |
| Qwen3.5 | 60 | No-Public 1R | i2b2 | 71.03 | 44.66 | 1540.33 | 673.33 | 1908.67 |
| Qwen3.5 | 60 | No-Public 2R | Source | 75.67 | 39.69 | 2327.70 | 745.48 | 3616.30 |
| Qwen3.5 | 60 | No-Public 2R | i2b2 | 70.97 | 39.20 | 1352.11 | 564.00 | 2096.89 |
| Qwen3.5 | 60 | SFT-Cont | Source | 75.77 | 47.59 | 2794.63 | 878.48 | 3149.37 |
| Qwen3.5 | 60 | SFT-Cont | i2b2 | 73.00 | 49.08 | 1692.89 | 642.78 | 1756.11 |
| Qwen3.5 | 60 | SFT-Res | Source | 76.00 | 42.02 | 2447.11 | 775.89 | 3496.89 |
| Qwen3.5 | 60 | SFT-Res | i2b2 | 72.25 | 44.28 | 1527.33 | 618.22 | 1921.67 |
| Qwen3.5 | 120 | Fed-MedLoRA+ 1R | Source | 77.46 | 50.73 | 3005.59 | 848.81 | 2938.41 |
| Qwen3.5 | 120 | Fed-MedLoRA+ 1R | i2b2 | 75.03 | 52.52 | 1811.56 | 611.44 | 1637.44 |
| Qwen3.5 | 120 | Meta-Cont | Source | 78.64 | 52.73 | 3125.52 | 807.67 | 2818.48 |
| Qwen3.5 | 120 | Meta-Cont | i2b2 | 74.83 | 52.14 | 1798.44 | 618.89 | 1650.56 |
| Qwen3.5 | 120 | Meta-Res | Source | 78.58 | 50.17 | 2962.85 | 776.48 | 2981.15 |

| Model | Steps/ client | Method | Domain | P (%) | R (%) | TP | FP | FN |
|---|---|---|---|---|---|---|---|---|
| Qwen3.5 | 120 | Meta-Res | i2b2 | 74.59 | 50.40 | 1738.44 | 600.67 | 1710.56 |
| Qwen3.5 | 120 | No-Public 1R | Source | 78.65 | 50.50 | 2997.52 | 789.04 | 2946.48 |
| Qwen3.5 | 120 | No-Public 1R | i2b2 | 76.21 | 51.86 | 1788.78 | 570.56 | 1660.22 |
| Qwen3.5 | 120 | No-Public 2R | Source | 79.56 | 50.53 | 3035.70 | 745.59 | 2908.30 |
| Qwen3.5 | 120 | No-Public 2R | i2b2 | 75.63 | 48.73 | 1680.67 | 551.11 | 1768.33 |
| Qwen3.5 | 120 | SFT-Cont | Source | 77.18 | 53.49 | 3164.48 | 891.96 | 2779.52 |
| Qwen3.5 | 120 | SFT-Cont | i2b2 | 73.32 | 55.12 | 1901.22 | 698.00 | 1547.78 |
| Qwen3.5 | 120 | SFT-Res | Source | 77.21 | 50.28 | 2976.44 | 856.00 | 2967.56 |
| Qwen3.5 | 120 | SFT-Res | i2b2 | 73.81 | 51.63 | 1780.56 | 645.22 | 1668.44 |
| Qwen3.5 | 240 | Fed-MedLoRA+ 1R | Source | 80.83 | 55.18 | 3284.22 | 736.74 | 2659.78 |
| Qwen3.5 | 240 | Fed-MedLoRA+ 1R | i2b2 | 76.98 | 56.01 | 1931.78 | 584.44 | 1517.22 |
| Qwen3.5 | 240 | Meta-Cont | Source | 81.10 | 57.65 | 3425.89 | 741.63 | 2518.11 |
| Qwen3.5 | 240 | Meta-Cont | i2b2 | 75.68 | 57.31 | 1976.56 | 637.89 | 1472.44 |
| Qwen3.5 | 240 | Meta-Res | Source | 81.31 | 53.82 | 3206.00 | 695.37 | 2738.00 |
| Qwen3.5 | 240 | Meta-Res | i2b2 | 76.62 | 53.70 | 1852.22 | 570.56 | 1596.78 |
| Qwen3.5 | 240 | No-Public 1R | Source | 81.78 | 55.96 | 3349.59 | 697.52 | 2594.41 |
| Qwen3.5 | 240 | No-Public 1R | i2b2 | 77.38 | 56.11 | 1935.22 | 571.00 | 1513.78 |
| Qwen3.5 | 240 | No-Public 2R | Source | 82.87 | 60.68 | 3691.74 | 697.59 | 2252.26 |
| Qwen3.5 | 240 | No-Public 2R | i2b2 | 76.14 | 57.04 | 1967.33 | 621.22 | 1481.67 |
| Qwen3.5 | 240 | SFT-Cont | Source | 81.26 | 56.75 | 3373.56 | 726.30 | 2570.44 |
| Qwen3.5 | 240 | SFT-Cont | i2b2 | 75.61 | 57.01 | 1966.11 | 635.78 | 1482.89 |
| Qwen3.5 | 240 | SFT-Res | Source | 80.82 | 54.24 | 3239.22 | 725.52 | 2704.78 |
| Qwen3.5 | 240 | SFT-Res | i2b2 | 75.81 | 54.39 | 1875.78 | 603.22 | 1573.22 |
| Qwen3.5 | 480 | Fed-MedLoRA+ 1R | Source | 82.40 | 63.21 | 3766.63 | 744.52 | 2177.37 |
| Qwen3.5 | 480 | Fed-MedLoRA+ 1R | i2b2 | 77.06 | 61.64 | 2125.89 | 636.67 | 1323.11 |
| Qwen3.5 | 480 | Meta-Cont | Source | 83.00 | 65.77 | 3918.48 | 746.11 | 2025.52 |
| Qwen3.5 | 480 | Meta-Cont | i2b2 | 76.50 | 62.14 | 2143.33 | 660.00 | 1305.67 |
| Qwen3.5 | 480 | Meta-Res | Source | 82.64 | 63.71 | 3797.85 | 740.96 | 2146.15 |
| Qwen3.5 | 480 | Meta-Res | i2b2 | 76.23 | 60.47 | 2085.44 | 655.44 | 1363.56 |
| Qwen3.5 | 480 | No-Public 1R | Source | 83.11 | 65.78 | 3948.33 | 729.59 | 1995.67 |
| Qwen3.5 | 480 | No-Public 1R | i2b2 | 76.85 | 62.64 | 2160.56 | 652.44 | 1288.44 |

| Model | Steps/ client | Method | Domain | P (%) | R (%) | TP | FP | FN |
|---|---|---|---|---|---|---|---|---|
| Qwen3.5 | 480 | No-Public 2R | Source | 84.61 | 70.03 | 4249.70 | 679.93 | 1694.30 |
| Qwen3.5 | 480 | No-Public 2R | i2b2 | 77.71 | 65.62 | 2263.11 | 650.78 | 1185.89 |
| Qwen3.5 | 480 | SFT-Cont | Source | 82.42 | 65.79 | 3917.48 | 769.59 | 2026.52 |
| Qwen3.5 | 480 | SFT-Cont | i2b2 | 75.84 | 63.74 | 2198.44 | 702.56 | 1250.56 |
| Qwen3.5 | 480 | SFT-Res | Source | 82.35 | 64.52 | 3853.26 | 761.07 | 2090.74 |
| Qwen3.5 | 480 | SFT-Res | i2b2 | 75.93 | 62.01 | 2138.78 | 681.67 | 1310.22 |

Precision, recall and counts are averaged across nine settings; Source additionally averages the three collections. Overall FP includes one untyped penalty per invalid output.

## Supplementary Table S14. NER entity F1 by training budget

| Model | Evaluation | Label | Gold | 30 steps | 60 steps | 120 steps | 240 steps | 480 steps |
|---|---|---|---|---|---|---|---|---|
| Qwen3.5 | Source | problem | 2677.0 | 59.35 → 70.00 | 69.74 → 74.92 | 76.46 → 79.14 | 80.33 → 81.49 | 82.27 → 82.90 |
| Qwen3.5 | Source | test | 1841.0 | 34.45 → 55.89 | 55.21 → 63.28 | 63.89 → 68.85 | 70.95 → 72.85 | 74.29 → 75.61 |
| Qwen3.5 | Source | drug | 685.0 | 68.72 → 75.79 | 74.14 → 78.28 | 77.52 → 78.93 | 80.23 → 81.04 | 83.94 → 84.17 |
| Qwen3.5 | Source | treatment | 409.3 | 27.61 → 43.00 | 40.63 → 48.65 | 49.03 → 54.24 | 55.83 → 57.81 | 56.74 → 58.93 |
| Qwen3.5 | i2b2 | problem | 1454.0 | 55.98 → 67.71 | 66.23 → 73.25 | 75.40 → 78.46 | 79.59 → 80.99 | 81.98 → 83.10 |
| Qwen3.5 | i2b2 | test | 1309.0 | 31.16 → 58.63 | 55.06 → 66.17 | 66.77 → 71.76 | 74.26 → 76.68 | 78.46 → 79.63 |
| Qwen3.5 | i2b2 | drug | 681.0 | 71.92 → 77.00 | 76.23 → 78.97 | 78.91 → 79.50 | 80.22 → 80.30 | 81.80 → 81.79 |
| Qwen3.5 | i2b2 | treatment | 405.0 | 27.56 → 42.90 | 39.26 → 49.89 | 51.46 → 55.62 | 58.44 → 60.95 | 59.75 → 61.49 |
| Llama3.1 | Source | problem | 2677.0 | 53.85 → 66.33 | 64.58 → 70.77 | 73.60 → 76.19 | 77.39 → 78.98 | 80.48 → 81.57 |
| Llama3.1 | Source | test | 1841.0 | 21.98 → 46.78 | 39.14 → 54.38 | 57.47 → 61.02 | 65.87 → 69.02 | 70.37 → 71.75 |
| Llama3.1 | Source | drug | 685.0 | 49.85 → 72.40 | 65.87 → 76.79 | 76.26 → 79.25 | 79.75 → 80.88 | 82.72 → 83.21 |
| Llama3.1 | Source | treatment | 409.3 | 25.37 → 37.81 | 33.71 → 40.30 | 45.94 → 49.18 | 51.74 → 54.86 | 55.87 → 56.73 |
| Llama3.1 | i2b2 | problem | 1454.0 | 46.71 → 58.44 | 57.66 → 65.20 | 69.46 → 73.20 | 74.75 → 77.00 | 79.02 → 80.34 |
| Llama3.1 | i2b2 | test | 1309.0 | 17.72 → 45.95 | 39.77 → 60.24 | 64.45 → 67.90 | 71.92 → 73.83 | 75.84 → 76.48 |
| Llama3.1 | i2b2 | drug | 681.0 | 60.90 → 73.99 | 71.01 → 76.45 | 76.96 → 77.92 | 79.10 → 79.16 | 80.65 → 81.03 |
| Llama3.1 | i2b2 | treatment | 405.0 | 25.26 → 36.61 | 31.52 → 40.18 | 46.72 → 50.88 | 52.82 → 55.63 | 57.69 → 58.17 |

Each budget cell shows SFT-Res → SFT-Cont F1, averaged across nine settings. Gold is the mean number of annotated targets. Source averages collection-level scores, rather than recomputing F1 from pooled counts. A zero Gold count indicates that the label is absent from that evaluation set.

## Supplementary Table S15. RE modifier F1 by training budget

| Model | Evaluation | Label | Gold | 30 steps | 60 steps | 120 steps | 240 steps | 480 steps |
|---|---|---|---|---|---|---|---|---|
| Qwen3.5 | Source | bodyloc | 747.3 | 3.36 → 4.70 | 5.88 → 8.79 | 12.27 → 18.35 | 20.13 → 31.46 | 54.42 → 56.46 |
| Qwen3.5 | Source | condition | 37.3 | 0.00 → 0.00 | 0.00 → 0.00 | 0.00 → 0.00 | 0.00 → 0.00 | 0.38 → 2.20 |
| Qwen3.5 | Source | course | 135.0 | 4.92 → 6.08 | 6.71 → 8.17 | 10.35 → 13.88 | 11.74 → 14.04 | 26.27 → 26.39 |
| Qwen3.5 | Source | dosage | 167.0 | 23.08 → 19.04 | 23.62 → 26.86 | 29.51 → 35.08 | 37.10 → 38.27 | 50.94 → 55.19 |
| Qwen3.5 | Source | duration | 18.7 | 19.69 → 19.67 | 22.55 → 20.22 | 25.49 → 25.56 | 27.79 → 32.94 | 34.95 → 34.47 |
| Qwen3.5 | Source | form | 302.3 | 30.96 → 34.17 | 38.72 → 41.56 | 44.55 → 50.14 | 50.87 → 53.38 | 59.69 → 62.75 |
| Qwen3.5 | Source | frequency | 283.7 | 41.90 → 56.77 | 61.37 → 63.11 | 64.95 → 66.01 | 66.59 → 67.70 | 70.05 → 70.74 |
| Qwen3.5 | Source | labvalue | 1226.0 | 50.01 → 62.62 | 69.08 → 77.61 | 79.48 → 82.03 | 81.88 → 83.34 | 85.40 → 86.11 |
| Qwen3.5 | Source | negation | 1096.7 | 71.41 → 81.65 | 82.23 → 85.59 | 86.72 → 87.86 | 88.65 → 89.42 | 90.51 → 90.79 |
| Qwen3.5 | Source | reference_range | 14.0 | 4.51 → 11.71 | 8.63 → 16.83 | 13.71 → 18.06 | 19.26 → 19.66 | 21.24 → 21.71 |
| Qwen3.5 | Source | route | 236.7 | 50.28 → 57.04 | 63.72 → 66.08 | 70.50 → 73.91 | 74.61 → 75.99 | 79.58 → 79.75 |
| Qwen3.5 | Source | severity | 73.0 | 21.30 → 23.27 | 26.93 → 37.95 | 36.30 → 45.79 | 43.93 → 47.19 | 50.34 → 52.20 |
| Qwen3.5 | Source | strength | 287.0 | 56.97 → 72.07 | 79.18 → 82.71 | 86.51 → 88.77 | 88.68 → 89.88 | 91.48 → 92.27 |
| Qwen3.5 | Source | subject | 56.3 | 0.56 → 1.47 | 1.12 → 4.30 | 5.75 → 7.17 | 5.59 → 3.86 | 12.43 → 13.20 |
| Qwen3.5 | Source | temporal | 1232.0 | 4.51 → 8.49 | 10.12 → 16.23 | 22.13 → 25.33 | 33.16 → 35.75 | 44.85 → 45.61 |
| Qwen3.5 | Source | uncertain | 31.0 | 4.00 → 7.27 | 6.10 → 11.25 | 9.71 → 11.86 | 9.89 → 13.90 | 19.01 → 21.72 |
| Qwen3.5 | i2b2 | bodyloc | 468.0 | 4.14 → 4.56 | 6.84 → 11.59 | 14.13 → 21.15 | 20.12 → 29.27 | 45.34 → 48.18 |
| Qwen3.5 | i2b2 | condition | 1.0 | 0.00 → 0.00 | 0.00 → 0.00 | 0.00 → 0.00 | 0.00 → 0.00 | 0.00 → 0.00 |
| Qwen3.5 | i2b2 | course | 93.0 | 6.06 → 8.00 | 7.61 → 11.90 | 13.89 → 17.94 | 14.65 → 21.56 | 34.89 → 38.47 |
| Qwen3.5 | i2b2 | dosage | 90.0 | 13.71 → 16.23 | 15.68 → 16.47 | 20.31 → 24.20 | 24.48 → 27.70 | 37.78 → 45.46 |
| Qwen3.5 | i2b2 | duration | 27.0 | 17.75 → 21.94 | 25.97 → 28.29 | 33.87 → 31.73 | 32.40 → 32.29 | 36.21 → 36.52 |
| Qwen3.5 | i2b2 | form | 129.0 | 36.16 → 38.91 | 43.52 → 44.15 | 45.19 → 48.90 | 49.60 → 50.29 | 55.74 → 60.00 |
| Qwen3.5 | i2b2 | frequency | 285.0 | 50.96 → 67.31 | 70.54 → 72.66 | 74.97 → 76.43 | 76.90 → 77.82 | 79.21 → 79.89 |
| Qwen3.5 | i2b2 | labvalue | 900.0 | 47.68 → 64.96 | 71.25 → 78.59 | 79.26 → 80.85 | 81.95 → 82.60 | 84.36 → 85.06 |
| Qwen3.5 | i2b2 | negation | 282.0 | 68.21 → 78.50 | 79.24 → 81.76 | 84.25 → 85.54 | 84.73 → 84.88 | 85.01 → 85.51 |
| Qwen3.5 | i2b2 | reference_range | 0.0 | 0.00 → 0.00 | 0.00 → 0.00 | 0.00 → 0.00 | 0.00 → 0.00 | 0.00 → 0.00 |
| Qwen3.5 | i2b2 | route | 245.0 | 61.16 → 73.16 | 77.50 → 79.39 | 81.59 → 85.28 | 83.92 → 87.45 | 88.69 → 90.60 |
| Qwen3.5 | i2b2 | severity | 37.0 | 22.83 → 20.99 | 27.03 → 37.71 | 40.49 → 46.49 | 47.24 → 47.52 | 56.15 → 57.47 |
| Qwen3.5 | i2b2 | strength | 261.0 | 59.96 → 72.82 | 81.87 → 82.76 | 86.66 → 88.85 | 88.09 → 89.84 | 91.67 → 91.98 |
| Qwen3.5 | i2b2 | subject | 0.0 | 0.00 → 0.00 | 0.00 → 0.00 | 0.00 → 0.00 | 0.00 → 0.00 | 0.00 → 0.00 |

| Model | Evaluation | Label | Gold | 30 steps | 60 steps | 120 steps | 240 steps | 480 steps |
|---|---|---|---|---|---|---|---|---|
| Qwen3.5 | i2b2 | temporal | 621.0 | 3.18 → 5.10 | 6.07 → 8.43 | 14.87 → 18.26 | 22.70 → 26.27 | 28.68 → 27.38 |
| Qwen3.5 | i2b2 | uncertain | 10.0 | 3.52 → 6.61 | 10.03 → 15.06 | 10.81 → 16.27 | 11.76 → 12.33 | 18.46 → 21.17 |
| Llama3.1 | Source | bodyloc | 747.3 | 0.47 → 1.66 | 7.08 → 8.07 | 8.69 → 13.30 | 11.15 → 15.41 | 39.82 → 48.89 |
| Llama3.1 | Source | condition | 37.3 | 0.00 → 0.00 | 0.00 → 0.00 | 0.00 → 0.00 | 0.28 → 0.00 | 0.00 → 0.10 |
| Llama3.1 | Source | course | 135.0 | 0.61 → 1.82 | 5.71 → 10.51 | 8.37 → 13.64 | 7.36 → 9.04 | 20.66 → 25.98 |
| Llama3.1 | Source | dosage | 167.0 | 0.40 → 0.85 | 6.05 → 8.99 | 14.57 → 14.28 | 15.13 → 14.45 | 23.70 → 24.05 |
| Llama3.1 | Source | duration | 18.7 | 2.68 → 3.50 | 11.55 → 12.67 | 22.58 → 20.10 | 23.46 → 25.04 | 27.37 → 27.07 |
| Llama3.1 | Source | form | 302.3 | 0.47 → 2.25 | 16.14 → 20.00 | 33.96 → 36.06 | 36.46 → 36.98 | 45.01 → 45.98 |
| Llama3.1 | Source | frequency | 283.7 | 2.05 → 5.86 | 21.71 → 29.90 | 48.72 → 56.33 | 60.48 → 61.91 | 66.54 → 67.42 |
| Llama3.1 | Source | labvalue | 1226.0 | 16.16 → 30.20 | 56.38 → 61.10 | 75.16 → 78.62 | 78.82 → 79.95 | 84.03 → 85.24 |
| Llama3.1 | Source | negation | 1096.7 | 11.50 → 27.12 | 57.85 → 71.26 | 81.54 → 83.57 | 84.90 → 85.34 | 89.04 → 89.37 |
| Llama3.1 | Source | reference_range | 14.0 | 1.13 → 5.68 | 15.67 → 14.20 | 16.20 → 16.71 | 17.54 → 21.79 | 25.13 → 27.05 |
| Llama3.1 | Source | route | 236.7 | 2.43 → 7.59 | 29.96 → 35.82 | 55.50 → 61.34 | 63.07 → 65.82 | 73.16 → 75.12 |
| Llama3.1 | Source | severity | 73.0 | 2.43 → 4.99 | 19.55 → 21.75 | 26.20 → 33.25 | 19.53 → 25.03 | 38.45 → 42.56 |
| Llama3.1 | Source | strength | 287.0 | 0.03 → 4.74 | 35.53 → 48.08 | 75.50 → 84.62 | 86.30 → 87.80 | 90.66 → 90.75 |
| Llama3.1 | Source | subject | 56.3 | 0.00 → 0.00 | 0.06 → 1.79 | 1.03 → 2.46 | 0.63 → 1.53 | 5.33 → 7.45 |
| Llama3.1 | Source | temporal | 1232.0 | 0.14 → 0.30 | 3.62 → 6.05 | 9.35 → 14.73 | 18.91 → 18.38 | 29.72 → 32.74 |
| Llama3.1 | Source | uncertain | 31.0 | 0.12 → 0.68 | 0.93 → 2.97 | 2.14 → 4.18 | 2.36 → 3.75 | 8.55 → 9.94 |
| Llama3.1 | i2b2 | bodyloc | 468.0 | 0.13 → 0.53 | 4.97 → 7.54 | 8.31 → 12.75 | 10.65 → 14.39 | 35.96 → 42.61 |
| Llama3.1 | i2b2 | condition | 1.0 | 0.00 → 0.00 | 0.00 → 0.00 | 0.00 → 0.00 | 0.00 → 0.00 | 0.00 → 0.00 |
| Llama3.1 | i2b2 | course | 93.0 | 0.24 → 0.95 | 6.34 → 13.12 | 8.00 → 16.09 | 8.14 → 11.95 | 24.48 → 27.99 |
| Llama3.1 | i2b2 | dosage | 90.0 | 0.00 → 1.33 | 7.81 → 15.08 | 19.96 → 31.21 | 38.46 → 35.81 | 40.30 → 44.43 |
| Llama3.1 | i2b2 | duration | 27.0 | 0.00 → 0.67 | 7.22 → 8.37 | 15.63 → 15.77 | 19.63 → 22.23 | 28.40 → 30.61 |
| Llama3.1 | i2b2 | form | 129.0 | 0.00 → 2.06 | 16.53 → 21.46 | 34.37 → 38.15 | 43.85 → 43.88 | 48.56 → 48.86 |
| Llama3.1 | i2b2 | frequency | 285.0 | 2.99 → 10.06 | 37.95 → 43.89 | 63.40 → 69.82 | 73.32 → 73.24 | 75.39 → 75.61 |
| Llama3.1 | i2b2 | labvalue | 900.0 | 10.36 → 24.51 | 48.43 → 56.04 | 71.26 → 75.27 | 76.34 → 78.27 | 80.34 → 81.69 |
| Llama3.1 | i2b2 | negation | 282.0 | 2.18 → 16.41 | 40.09 → 62.90 | 79.47 → 82.90 | 84.03 → 84.09 | 86.35 → 86.96 |
| Llama3.1 | i2b2 | reference_range | 0.0 | 0.00 → 0.00 | 0.00 → 0.00 | 0.00 → 0.00 | 0.00 → 0.00 | 0.00 → 0.00 |
| Llama3.1 | i2b2 | route | 245.0 | 3.11 → 10.14 | 46.68 → 51.38 | 74.69 → 78.93 | 79.73 → 80.54 | 85.42 → 86.10 |
| Llama3.1 | i2b2 | severity | 37.0 | 0.57 → 3.16 | 11.08 → 19.99 | 23.05 → 31.50 | 20.46 → 31.91 | 43.61 → 47.28 |
| Llama3.1 | i2b2 | strength | 261.0 | 0.00 → 6.06 | 46.07 → 59.20 | 84.69 → 88.87 | 89.27 → 89.98 | 91.41 → 91.71 |

| Model | Evaluation | Label | Gold | 30 steps | 60 steps | 120 steps | 240 steps | 480 steps |
|---|---|---|---|---|---|---|---|---|
| Llama3.1 | i2b2 | subject | 0.0 | 0.00 → 0.00 | 0.00 → 0.00 | 0.00 → 0.00 | 0.00 → 0.00 | 0.00 → 0.00 |
| Llama3.1 | i2b2 | temporal | 621.0 | 0.07 → 0.21 | 1.22 → 3.05 | 2.72 → 4.87 | 5.64 → 6.95 | 12.79 → 15.95 |
| Llama3.1 | i2b2 | uncertain | 10.0 | 0.00 → 0.00 | 0.00 → 0.00 | 0.00 → 0.00 | 0.00 → 0.00 | 1.59 → 6.88 |

Each budget cell shows SFT-Res → SFT-Cont F1, averaged across nine settings. Gold is the mean number of annotated targets. Source averages collection-level scores, rather than recomputing F1 from pooled counts. A zero Gold count indicates that the label is absent from that evaluation set.

## S6 Computational resources

Public preparation is an upstream cost, reported separately from clinical fitting and communication.

### Supplementary Table S16. Clinical training time and communication

| Model | Task | Method | Total steps/client | Rounds | Client training (min) | Total time (min) | Communication (MiB) |
|---|---|---|---|---|---|---|---|
| Qwen3.5 | NER | Fed-MedLoRA+ 2R | 120 | 2 | 46.761 | 47.613 | 180 |
| Llama3.1 | NER | No-Public 2R | 120 | 2 | 9.552 | 10.740 | 175.5 |
| Llama3.1 | NER | Fed-MedLoRA+ 2R | 120 | 2 | 9.593 | 10.065 | 175.5 |
| Qwen3.5 | NER | Fed-MedLoRA+ 2R | 240 | 2 | 93.390 | 93.968 | 180 |
| Llama3.1 | NER | No-Public 2R | 240 | 2 | 19.098 | 19.433 | 175.5 |
| Qwen3.5 | NER | SFT-Cont 2R | 120 | 2 | 46.052 | 46.504 | 180 |
| Llama3.1 | NER | SFT-Cont 2R | 120 | 2 | 8.282 | 8.719 | 175.5 |
| Qwen3.5 | NER | SFT-Cont | 30 | 1 | 11.403 | 11.686 | 90 |
| Llama3.1 | NER | SFT-Cont | 30 | 1 | 2.064 | 2.365 | 87.75 |
| Llama3.1 | NER | No-Public 2R | 60 | 2 | 4.770 | 5.101 | 175.5 |
| Qwen3.5 | NER | No-Public 1R | 30 | 1 | 11.665 | 12.155 | 90 |
| Qwen3.5 | NER | SFT-Res | 30 | 1 | 11.680 | 11.904 | 90 |
| Llama3.1 | NER | No-Public 1R | 30 | 1 | 2.411 | 2.760 | 87.75 |
| Llama3.1 | NER | SFT-Res | 30 | 1 | 2.398 | 2.612 | 87.75 |
| Qwen3.5 | NER | Fed-MedLoRA+ 1R | 30 | 1 | 11.679 | 12.066 | 90 |
| Qwen3.5 | NER | Meta-Res | 30 | 1 | 11.678 | 11.909 | 90 |
| Qwen3.5 | NER | Meta-Cont | 30 | 1 | 11.430 | 11.683 | 90 |
| Qwen3.5 | RE | No-Public 1R | 30 | 1 | 12.811 | 13.269 | 90 |
| Qwen3.5 | RE | SFT-Res | 30 | 1 | 12.763 | 13.106 | 90 |
| Qwen3.5 | RE | SFT-Cont | 30 | 1 | 12.552 | 12.863 | 90 |
| Qwen3.5 | RE | Meta-Res | 30 | 1 | 12.833 | 13.089 | 90 |

| Model | Task | Method | Total steps/client | Rounds | Client training (min) | Total time (min) | Communication (MiB) |
|---|---|---|---|---|---|---|---|
| Qwen3.5 | RE | Meta-Cont | 30 | 1 | 12.502 | 12.759 | 90 |
| Llama3.1 | RE | No-Public 1R | 30 | 1 | 2.383 | 2.630 | 87.75 |
| Llama3.1 | RE | SFT-Res | 30 | 1 | 2.370 | 2.611 | 87.75 |
| Llama3.1 | RE | SFT-Cont | 30 | 1 | 2.066 | 2.305 | 87.75 |
| Qwen3.5 | RE | Fed-MedLoRA+ 1R | 30 | 1 | 12.825 | 13.305 | 90 |
| Qwen3.5 | NER | Fed-MedLoRA+ 2R | 60 | 2 | 23.197 | 24.150 | 180 |
| Llama3.1 | NER | Fed-MedLoRA+ 2R | 60 | 2 | 4.773 | 5.270 | 175.5 |
| Llama3.1 | NER | Fed-MedLoRA+ 2R | 240 | 2 | 19.076 | 19.501 | 175.5 |

Times average nine runs, each training three clients sequentially on one H200. Client training is optimization time; total time also includes setup and server work. Total steps are split equally across rounds. Communication excludes public preparation and initial adapter distribution. Public preparation cost is reported in Table S6.

Each FP32 adapter occupies 15 MiB for Qwen3.5 or 14.625 MiB for Llama3.1. Continuation retains one adapter; residual adaptation retains two, totaling 30 MiB or 29.25 MiB. One clinical round exchanges 90 MiB for Qwen3.5 or 87.75 MiB for Llama3.1 across three clients. Initial public-adapter distribution is 45 MiB or 43.875 MiB and is accounted for separately.

## S7 Statistical robustness

Initialization-block intervals provide a sensitivity analysis of the paired bootstrap estimates.

### Supplementary Table S17. Initialization-block sensitivity of primary source-mean comparisons

| Model | Task | Steps/client | Domain | Contrast | Mean [bootstrap CI] | Block-t 95% CI | Positive blocks |
|---|---|---|---|---|---|---|---|
| Qwen3.5 | NER | 60 | Source | SFT-Cont − No-Public 1R | +7.82 [6.65, 9.04] | [5.93, 9.71] | 3/3 |
| Qwen3.5 | NER | 60 | Source | SFT-Cont − SFT-Res | +6.05 [4.26, 7.33] | [2.06, 10.03] | 3/3 |
| Llama3.1 | NER | 60 | Source | SFT-Cont − SFT-Res | +8.89 [5.78, 11.50] | [4.41, 13.37] | 3/3 |
| Llama3.1 | NER | 60 | Source | SFT-Cont − No-Public 1R | +10.34 [8.08, 12.54] | [6.03, 14.64] | 3/3 |
| Qwen3.5 | NER | 120 | Source | SFT-Cont − No-Public 1R | +3.00 [1.96, 4.18] | [1.09, 4.91] | 3/3 |
| Qwen3.5 | NER | 120 | Source | SFT-Cont − SFT-Res | +3.29 [2.49, 4.27] | [2.59, 3.99] | 3/3 |
| Llama3.1 | NER | 120 | Source | SFT-Cont − SFT-Res | +2.84 [1.86, 3.77] | [0.80, 4.89] | 3/3 |
| Llama3.1 | NER | 120 | Source | SFT-Cont − No-Public 1R | +4.00 [2.92, 5.07] | [1.43, 6.57] | 3/3 |
| Qwen3.5 | NER | 240 | Source | SFT-Cont − No-Public 1R | +1.24 [0.77, 1.78] | [−0.01, 2.50] | 3/3 |
| Qwen3.5 | NER | 240 | Source | SFT-Cont − SFT-Res | +1.31 [0.80, 1.99] | [0.11, 2.51] | 3/3 |
| Llama3.1 | NER | 240 | Source | SFT-Cont − SFT-Res | +1.93 [1.56, 2.24] | [1.62, 2.23] | 3/3 |
| Llama3.1 | NER | 240 | Source | SFT-Cont − No-Public 1R | +2.44 [1.65, 2.91] | [1.29, 3.58] | 3/3 |
| Qwen3.5 | NER | 480 | Source | SFT-Cont − No-Public 1R | +0.80 [0.31, 1.19] | [−0.27, 1.87] | 3/3 |
| Qwen3.5 | NER | 480 | Source | SFT-Cont − SFT-Res | +0.80 [0.39, 1.17] | [0.27, 1.33] | 3/3 |

| Model | Task | Steps/client | Domain | Contrast | Mean [bootstrap CI] | Block-t 95% CI | Positive blocks |
|---|---|---|---|---|---|---|---|
| Llama3.1 | NER | 480 | Source | SFT-Cont − SFT-Res | +0.99 [0.49, 1.55] | [−0.28, 2.26] | 3/3 |
| Llama3.1 | NER | 480 | Source | SFT-Cont − No-Public 1R | +1.12 [0.60, 1.61] | [0.05, 2.18] | 3/3 |
| Qwen3.5 | RE | 60 | Source | SFT-Cont − No-Public 1R | +3.83 [1.67, 8.38] | [−3.46, 11.13] | 3/3 |
| Qwen3.5 | RE | 60 | Source | SFT-Cont − SFT-Res | +4.53 [2.41, 7.59] | [0.14, 8.92] | 3/3 |
| Llama3.1 | RE | 60 | Source | SFT-Cont − SFT-Res | +7.06 [−7.04, 20.13] | [−20.91, 35.04] | 2/3 |
| Llama3.1 | RE | 60 | Source | SFT-Cont − No-Public 1R | +10.66 [4.51, 15.56] | [−0.19, 21.51] | 3/3 |
| Qwen3.5 | RE | 120 | Source | SFT-Cont − No-Public 1R | +1.69 [1.16, 2.23] | [1.07, 2.31] | 3/3 |
| Qwen3.5 | RE | 120 | Source | SFT-Cont − SFT-Res | +2.29 [1.32, 3.24] | [0.18, 4.41] | 3/3 |
| Llama3.1 | RE | 120 | Source | SFT-Cont − SFT-Res | +3.49 [1.94, 5.05] | [0.81, 6.18] | 3/3 |
| Llama3.1 | RE | 120 | Source | SFT-Cont − No-Public 1R | +5.56 [3.47, 8.32] | [−0.10, 11.23] | 3/3 |
| Qwen3.5 | RE | 240 | Source | SFT-Cont − No-Public 1R | +0.41 [−1.28, 2.03] | [−2.40, 3.23] | 2/3 |
| Qwen3.5 | RE | 240 | Source | SFT-Cont − SFT-Res | +1.90 [0.95, 2.85] | [1.02, 2.78] | 3/3 |
| Llama3.1 | RE | 240 | Source | SFT-Cont − SFT-Res | +0.97 [−0.57, 2.65] | [−1.94, 3.87] | 3/3 |
| Llama3.1 | RE | 240 | Source | SFT-Cont − No-Public 1R | +1.51 [0.25, 2.51] | [0.59, 2.44] | 3/3 |
| Qwen3.5 | RE | 480 | Source | SFT-Cont − No-Public 1R | −0.33 [−1.45, 1.02] | [−2.15, 1.50] | 1/3 |
| Qwen3.5 | RE | 480 | Source | SFT-Cont − SFT-Res | +0.79 [−0.13, 1.57] | [−0.48, 2.06] | 3/3 |
| Llama3.1 | RE | 480 | Source | SFT-Cont − SFT-Res | +1.66 [0.61, 2.39] | [0.07, 3.26] | 3/3 |
| Llama3.1 | RE | 480 | Source | SFT-Cont − No-Public 1R | +1.67 [0.62, 2.71] | [−0.42, 3.76] | 3/3 |
| Qwen3.5 | NER | 60/120 | Source | SFT-Cont (60 steps) − Fed-MedLoRA+ 2R (120 steps) | −0.90 [−4.16, 2.32] | [−8.11, 6.30] | 2/3 |
| Llama3.1 | NER | 60/120 | Source | SFT-Cont (60 steps) − No-Public 2R (120 steps) | +0.66 [−1.92, 2.98] | [−3.92, 5.25] | 2/3 |
| Qwen3.5 | NER | 60 | Source | SFT-Cont: global − own-local | +0.65 [−0.87, 1.97] | [−2.86, 4.15] | 2/3 |
| Qwen3.5 | NER | 60 | Source | No-Public 1R: global − own-local | −0.42 [−1.93, 1.08] | [−2.76, 1.92] | 1/3 |
| Qwen3.5 | NER | 60 | Source | SFT-Cont local − No-Public 1R local | +6.75 [5.16, 8.39] | [6.17, 7.34] | 3/3 |
| Qwen3.5 | NER | 60 | Source | (Global − local): SFT-Cont − No-Public | +1.07 [−0.64, 3.15] | [−1.27, 3.40] | 3/3 |
| Llama3.1 | NER | 60 | Source | SFT-Cont: global − own-local | +3.85 [0.29, 7.85] | [−0.36, 8.05] | 3/3 |
| Llama3.1 | NER | 60 | Source | No-Public 1R: global − own-local | +1.32 [−1.16, 5.26] | [−5.10, 7.74] | 2/3 |
| Llama3.1 | NER | 60 | Source | SFT-Cont local − No-Public 1R local | +7.81 [4.79, 10.58] | [5.61, 10.02] | 3/3 |
| Llama3.1 | NER | 60 | Source | (Global − local): SFT-Cont − No-Public | +2.52 [−0.01, 6.67] | [−3.40, 8.45] | 3/3 |
| Qwen3.5 | NER | 120/240 | Source | SFT-Cont (120 steps) − Fed-MedLoRA+ 2R (240 steps) | −1.62 [−2.53, −0.92] | [−2.71, −0.53] | 0/3 |
| Llama3.1 | NER | 120/240 | Source | SFT-Cont (120 steps) − No-Public 2R (240 steps) | −1.51 [−2.94, −0.10] | [−4.74, 1.71] | 0/3 |
| Llama3.1 | NER | 120/240 | Source | SFT-Cont (120 steps) − No-Public 1R (240 steps) | −1.61 [−3.01, −0.25] | [−4.23, 1.01] | 0/3 |

| Model | Task | Steps/client | Domain | Contrast | Mean [bootstrap CI] | Block-t 95% CI | Positive blocks |
|---|---|---|---|---|---|---|---|
| Qwen3.5 | NER | 120 | Source | SFT-Cont 2R − SFT-Cont | −0.67 [−1.30, −0.16] | [−1.90, 0.55] | 0/3 |
| Qwen3.5 | NER | 30/60 | Source | SFT-Cont (30 steps) − No-Public 2R (60 steps) | +5.44 [0.86, 9.66] | [−3.12, 14.01] | 3/3 |
| Llama3.1 | NER | 120 | Source | SFT-Cont 2R − SFT-Cont | −0.36 [−1.66, 0.94] | [−3.50, 2.78] | 1/3 |
| Llama3.1 | NER | 30/60 | Source | SFT-Cont (30 steps) − No-Public 2R (60 steps) | +11.02 [7.79, 14.91] | [4.92, 17.13] | 3/3 |
| Qwen3.5 | NER | 120 | Source | Agg-D − SFT-Cont | −0.16 [−0.23, −0.06] | [−0.25, −0.07] | 0/3 |
| Llama3.1 | NER | 60 | Source | Agg-D − SFT-Cont | +0.08 [−0.02, 0.19] | [−0.04, 0.20] | 3/3 |
| Qwen3.5 | NER | 30 | Source | SFT-Cont − No-Public 1R | +14.81 [11.40, 19.20] | [5.86, 23.76] | 3/3 |
| Qwen3.5 | NER | 30 | Source | SFT-Cont − SFT-Res | +13.26 [10.76, 16.07] | [7.98, 18.54] | 3/3 |
| Qwen3.5 | RE | 30 | Source | SFT-Cont − No-Public 1R | +10.61 [3.75, 19.17] | [−6.04, 27.25] | 3/3 |
| Qwen3.5 | RE | 30 | Source | SFT-Cont − SFT-Res | +8.76 [4.02, 15.19] | [−2.89, 20.42] | 3/3 |
| Qwen3.5 | NER | 30/60 | Source | SFT-Cont (30 steps) − No-Public 1R (60 steps) | +2.47 [−1.37, 4.56] | [−2.87, 7.81] | 3/3 |
| Llama3.1 | NER | 30 | Source | SFT-Cont − No-Public 1R | +18.68 [15.29, 24.19] | [9.02, 28.33] | 3/3 |
| Llama3.1 | NER | 30 | Source | SFT-Cont − SFT-Res | +16.52 [14.14, 19.35] | [10.90, 22.14] | 3/3 |
| Llama3.1 | RE | 30 | Source | SFT-Cont − No-Public 1R | +8.92 [1.04, 17.07] | [0.71, 17.13] | 3/3 |
| Llama3.1 | RE | 30 | Source | SFT-Cont − SFT-Res | +8.70 [3.32, 14.59] | [−1.54, 18.93] | 3/3 |
| Llama3.1 | NER | 30/60 | Source | SFT-Cont (30 steps) − No-Public 1R (60 steps) | +5.24 [2.31, 7.83] | [−0.53, 11.02] | 3/3 |
| Llama3.1 | NER | 60 | Source | SFT-Cont − Fed-MedLoRA+ 1R | +11.24 [8.60, 14.12] | [5.41, 17.06] | 3/3 |
| Llama3.1 | NER | 60/120 | Source | SFT-Cont (60 steps) − Fed-MedLoRA+ 2R (120 steps) | −1.39 [−4.07, 0.95] | [−6.35, 3.58] | 1/3 |
| Qwen3.5 | NER | 30/60 | Source | SFT-Cont (30 steps) − Fed-MedLoRA+ 1R (60 steps) | +3.22 [−0.33, 5.21] | [−1.84, 8.28] | 3/3 |
| Qwen3.5 | NER | 30/60 | Source | SFT-Cont (30 steps) − Fed-MedLoRA+ 2R (60 steps) | +4.58 [−0.05, 8.89] | [−3.62, 12.78] | 3/3 |
| Qwen3.5 | NER | 60/120 | Source | SFT-Cont (60 steps) − No-Public 1R (120 steps) | −1.05 [−4.32, 2.19] | [−8.56, 6.46] | 1/3 |
| Qwen3.5 | NER | 60/120 | Source | SFT-Cont (60 steps) − Fed-MedLoRA+ 1R (120 steps) | −0.33 [−3.72, 2.82] | [−8.20, 7.54] | 2/3 |
| Qwen3.5 | NER | 60/120 | Source | SFT-Cont (60 steps) − No-Public 2R (120 steps) | +0.56 [−2.14, 3.35] | [−5.32, 6.44] | 2/3 |
| Qwen3.5 | NER | 120/240 | Source | SFT-Cont (120 steps) − No-Public 1R (240 steps) | −1.55 [−2.38, −0.93] | [−2.54, −0.56] | 0/3 |
| Qwen3.5 | NER | 120/240 | Source | SFT-Cont (120 steps) − Fed-MedLoRA+ 1R (240 steps) | −0.82 [−1.77, 0.04] | [−2.02, 0.38] | 0/3 |
| Qwen3.5 | NER | 120/240 | Source | SFT-Cont (120 steps) − No-Public 2R (240 steps) | −1.39 [−2.10, −0.78] | [−2.51, −0.27] | 0/3 |
| Llama3.1 | NER | 30/60 | Source | SFT-Cont (30 steps) − Fed-MedLoRA+ 1R (60 steps) | +6.14 [3.01, 8.83] | [0.08, 12.20] | 3/3 |
| Llama3.1 | NER | 30/60 | Source | SFT-Cont (30 steps) − Fed-MedLoRA+ 2R (60 steps) | +4.76 [2.15, 7.22] | [0.00, 9.52] | 3/3 |
| Llama3.1 | NER | 60/120 | Source | SFT-Cont (60 steps) − No-Public 1R (120 steps) | −1.26 [−3.59, 0.82] | [−5.75, 3.23] | 1/3 |
| Llama3.1 | NER | 60/120 | Source | SFT-Cont (60 steps) − Fed-MedLoRA+ 1R (120 steps) | −0.17 [−2.53, 2.09] | [−4.88, 4.54] | 1/3 |
| Llama3.1 | NER | 120/240 | Source | SFT-Cont (120 steps) − Fed-MedLoRA+ 2R (240 steps) | −2.37 [−3.95, −0.92] | [−5.71, 0.96] | 0/3 |

| Model | Task | Steps/client | Domain | Contrast | Mean [bootstrap CI] | Block-t 95% CI | Positive blocks |
|---|---|---|---|---|---|---|---|
| Qwen3.5 | NER | 30 | Source | SFT-Cont − Fed-MedLoRA+ 1R | +15.75 [12.74, 19.83] | [7.18, 24.32] | 3/3 |
| Qwen3.5 | NER | 60 | Source | SFT-Cont − Fed-MedLoRA+ 1R | +8.57 [7.31, 9.80] | [6.49, 10.65] | 3/3 |
| Qwen3.5 | NER | 120 | Source | SFT-Cont − Fed-MedLoRA+ 1R | +3.72 [2.62, 4.80] | [1.42, 6.02] | 3/3 |
| Qwen3.5 | NER | 240 | Source | SFT-Cont − Fed-MedLoRA+ 1R | +1.97 [1.33, 2.80] | [0.10, 3.85] | 3/3 |
| Qwen3.5 | NER | 480 | Source | SFT-Cont − Fed-MedLoRA+ 1R | +1.54 [1.18, 1.85] | [0.80, 2.28] | 3/3 |
| Llama3.1 | NER | 30 | Source | SFT-Cont − Fed-MedLoRA+ 1R | +19.07 [15.73, 24.39] | [9.89, 28.25] | 3/3 |
| Llama3.1 | NER | 120 | Source | SFT-Cont − Fed-MedLoRA+ 1R | +5.09 [4.11, 6.33] | [3.33, 6.85] | 3/3 |
| Qwen3.5 | NER | 30 | Source | SFT-Cont − No-Public in local models | +13.23 [7.92, 16.88] | [2.08, 24.38] | 3/3 |
| Qwen3.5 | NER | 30 | Source | No-Public: global − own-local | +1.98 [−2.39, 5.92] | [−8.27, 12.23] | 2/3 |
| Qwen3.5 | NER | 30 | Source | SFT-Cont: global − own-local | +3.56 [0.78, 7.77] | [−3.34, 10.46] | 3/3 |
| Qwen3.5 | NER | 30 | Source | (Global − local): SFT-Cont − No-Public | +1.58 [−1.70, 5.27] | [−5.15, 8.30] | 2/3 |
| Qwen3.5 | NER | 120 | Source | SFT-Cont − No-Public in local models | +2.09 [1.15, 3.18] | [0.42, 3.77] | 3/3 |
| Qwen3.5 | NER | 120 | Source | No-Public: global − own-local | −0.72 [−1.37, −0.11] | [−2.14, 0.69] | 0/3 |
| Qwen3.5 | NER | 120 | Source | SFT-Cont: global − own-local | +0.18 [−0.44, 0.84] | [−1.03, 1.39] | 2/3 |
| Qwen3.5 | NER | 120 | Source | (Global − local): SFT-Cont − No-Public | +0.91 [0.28, 1.48] | [0.03, 1.78] | 3/3 |
| Qwen3.5 | NER | 120 − 30 | Source | Change in SFT-Cont aggregation gain | −3.38 [−8.09, −0.57] | [−10.68, 3.92] | 0/3 |
| Llama3.1 | NER | 30 | Source | SFT-Cont − No-Public in local models | +13.82 [10.06, 18.21] | [4.57, 23.08] | 3/3 |
| Llama3.1 | NER | 30 | Source | No-Public: global − own-local | −0.79 [−2.33, 0.70] | [−4.13, 2.56] | 1/3 |
| Llama3.1 | NER | 30 | Source | SFT-Cont: global − own-local | +4.06 [1.01, 6.41] | [−1.92, 10.05] | 3/3 |
| Llama3.1 | NER | 30 | Source | (Global − local): SFT-Cont − No-Public | +4.85 [2.39, 7.18] | [2.09, 7.61] | 3/3 |
| Llama3.1 | NER | 120 | Source | SFT-Cont − No-Public in local models | +4.80 [3.43, 6.03] | [2.56, 7.04] | 3/3 |
| Llama3.1 | NER | 120 | Source | No-Public: global − own-local | +0.13 [−1.09, 1.25] | [−2.05, 2.30] | 2/3 |
| Llama3.1 | NER | 120 | Source | SFT-Cont: global − own-local | −0.67 [−1.99, 0.46] | [−3.52, 2.19] | 1/3 |
| Llama3.1 | NER | 120 | Source | (Global − local): SFT-Cont − No-Public | −0.79 [−1.87, 0.40] | [−2.69, 1.10] | 1/3 |
| Llama3.1 | NER | 120 − 30 | Source | Change in SFT-Cont aggregation gain | −4.73 [−6.85, −1.99] | [−8.13, −1.33] | 0/3 |

The table covers matched-budget extraction gains, half-budget comparisons, local preparation, aggregation, and fixed-total round controls for mean source F1. Block-t intervals use three initialization-block mean effects and two degrees of freedom. Positive blocks counts block means above zero. Intervals are pointwise and describe training variation on fixed test collections. Test-set-specific paired bootstrap estimates are in Tables S3, S5, S7, S9 and S11.

## Supplementary references